\documentclass{article}

\usepackage{authblk}
\usepackage[utf8]{inputenc}
\usepackage{main} % ACL style file
\usepackage{microtype}
\usepackage{latexsym}

\usepackage[T1]{fontenc}

\usepackage{algorithm}
\usepackage{algorithmic}
\usepackage{graphicx}
\usepackage{xcolor}
\definecolor{deepgreen}{RGB}{34,139,34}
\usepackage{amsmath}
\usepackage{amssymb}
\usepackage{amsfonts}
\usepackage{mathtools}
\usepackage{bm}

\usepackage{graphicx}
\usepackage{subcaption}
\usepackage{float}
\usepackage{wrapfig}

\usepackage{booktabs}
\usepackage{array}
\usepackage{multirow}
\usepackage{multicol}
\usepackage{makecell}
\usepackage{longtable}
\usepackage{tabularx}
\usepackage{colortbl}
\usepackage{arydshln}

\usepackage{color}
\usepackage{xcolor}
\definecolor{mydarkblue}{rgb}{0,0.08,0.45}
\definecolor{olmopink}{HTML}{FF419C}
\definecolor{seedblue}{HTML}{2E5AA8}  % Seed color from reference paper
\definecolor{wkblue}{rgb}{0.2, 0.3, 0.6}
\definecolor{meta-color}{rgb}{0.5, 0.5, 0.5}
\definecolor{bgblue}{RGB}{245,243,253}
\definecolor{ttblue}{RGB}{91,194,224}
\definecolor{mybrown}{RGB}{128,64,0}
\definecolor{titlecolor}{HTML}{4c9cff}
\definecolor{coolblue3}{rgb}{0.91, 0.94, 0.98}
\definecolor{myblue}{rgb}{0.9, 0.1, 0.94}
\definecolor{mygreen}{rgb}{0.64, 0.56, 0.88}
\definecolor{myyellow}{rgb}{0.68, 0.6, 0.1}
\definecolor{fancygreen}{rgb}{0.33, 0.68, 0.20}
\definecolor{salmon}{rgb}{0.94, 0.52, 0.49}
\definecolor{tablegreen}{rgb}{0.82, 0.94, 0.75}
\definecolor{tableblue}{rgb}{0.81, 0.90, 0.94}
\definecolor{tablered}{rgb}{0.97, 0.85, 0.85}
\definecolor{tableorange}{rgb}{0.96, 0.85, 0.81}

\definecolor{gred}{RGB}{250, 210, 207}
\definecolor{gblue}{RGB}{210, 227, 252}
\definecolor{gyellow}{RGB}{254, 239, 195}
\definecolor{ggreen}{RGB}{206, 234, 214}
\definecolor{gorange}{RGB}{254, 223, 200}
\definecolor{gblue9}{RGB}{23, 78, 166}
\definecolor{gred9}{RGB}{165, 14, 14}
\definecolor{gyellow9}{RGB}{227, 116, 0}
\definecolor{ggreen9}{RGB}{13, 101, 45}
\definecolor{gorange9}{RGB}{176, 96, 0}

\definecolor{catblue}{HTML}{DDEEFF}   % category 行，再淡一档
\definecolor{rowblue}{HTML}{F2F8FF}   % 交替行
\usepackage{enumitem}
\usepackage{footnote}
\usepackage{lipsum}
\usepackage{setspace}

\usepackage{geometry}
\usepackage{fancyhdr}
\usepackage{lscape}
\usepackage{rotating}

\usepackage{algorithm}
\usepackage{algorithmic}

\usepackage[most]{tcolorbox}
\tcbuselibrary{skins,breakable,fitting,hooks}  % 添加tcolorbox额外库
\usepackage[framemethod=tikz]{mdframed}
\usepackage{awesomebox}
\usepackage{tcolorbox}
\usepackage{courier} % 可选：更像代码字体
\usepackage{mdframed}
\definecolor{darkblue}{RGB}{84, 112, 198}

\tcbset{
  takeawaystyle/.style={
    width=\linewidth,
    top=8pt,
    bottom=4pt,
    colback=blue!5!white,            % 主框背景色（浅蓝）
    colframe=darkblue,               % 主框边框颜色（深蓝）
    boxrule=0.5pt,                   % 边框线宽度
    colbacktitle=darkblue,           % 标题背景色（深蓝）
    fonttitle=\bfseries\color{white}, % 标题字体：加粗、白色
    enhanced,
    center,
    attach boxed title to top left={yshift=-0.1in,xshift=0.15in},
    boxed title style={
      boxrule=0pt,                   % 标题框无边框
      colframe=white,                % 标题边框颜色
      colback=darkblue,              % 标题背景色（深蓝）
    },
  }
}

\newtcolorbox{takeawaybox}[2][]{takeawaystyle,title=#2,#1}

\usepackage[numbers]{natbib}
\usepackage{url}
\usepackage[colorlinks=true,linkcolor=seedblue,citecolor=mydarkblue,filecolor=mydarkblue,urlcolor=mydarkblue]{hyperref}

\newtcolorbox{keybox}{
    colback=blue!5!white,
    colframe=blue!75!black,
    fonttitle=\bfseries,
    title=Takeaways
}

\newtcolorbox{myboxi}[1][]{
  breakable,
  title=#1,
  colback=red!5,
  colbacktitle=red!5,
  coltitle=black,
  fonttitle=\bfseries,
  bottomrule=0pt,
  toprule=0pt,
  leftrule=2pt,
  rightrule=2pt,
  titlerule=0pt,
  arc=0pt,
  outer arc=0pt,
  colframe=red,
}

\newtcolorbox{myboxnote}[1][]{
  breakable,
  title=#1,
  colback=orange!0,
  colbacktitle=orange!0,
  coltitle=black,
  fonttitle=\bfseries,
  bottomrule=0pt,
  toprule=0pt,
  leftrule=2pt,
  rightrule=2pt,
  titlerule=0pt,
  arc=0pt,
  outer arc=0pt,
  colframe=orange,
}

\newtcolorbox{myboxii}[1][]{
  breakable,
  freelance,
  title=#1,
  colback=white,
  colbacktitle=white,
  coltitle=black,
  fonttitle=\bfseries,
  bottomrule=0pt,
  boxrule=0pt,
  colframe=white,
  overlay unbroken and first={
  \draw[red!75!black,line width=3pt]
    ([xshift=5pt]frame.north west) -- 
    (frame.north west) -- 
    (frame.south west);
  \draw[red!75!black,line width=3pt]
    ([xshift=-5pt]frame.north east) -- 
    (frame.north east) -- 
    (frame.south east);
  },
  overlay unbroken app={
  \draw[red!75!black,line width=3pt,line cap=rect]
    (frame.south west) -- 
    ([xshift=5pt]frame.south west);
  \draw[red!75!black,line width=3pt,line cap=rect]
    (frame.south east) -- 
    ([xshift=-5pt]frame.south east);
  },
  overlay middle and last={
  \draw[red!75!black,line width=3pt]
    (frame.north west) -- 
    (frame.south west);
  \draw[red!75!black,line width=3pt]
    (frame.north east) -- 
    (frame.south east);
  },
  overlay last app={
  \draw[red!75!black,line width=3pt,line cap=rect]
    (frame.south west) --
    ([xshift=5pt]frame.south west);
  \draw[red!75!black,line width=3pt,line cap=rect]
    (frame.south east) --
    ([xshift=-5pt]frame.south east);
  },
}

\usepackage[most]{tcolorbox}
\tcbuselibrary{skins,breakable,listings}
\usepackage{fvextra}

\DefineVerbatimEnvironment{TextBlock}{Verbatim}{
  breaklines=true,
  breakanywhere=true,
  breaksymbolleft={},
  obeytabs=true,
  commandchars=\\\{\}
}

\usepackage{bbding}
\usepackage{imakeidx}
\makeindex
\usepackage[subfigure]{tocloft}  % For customizing table of contents
\usepackage[edges]{forest}
\usepackage[normalem]{ulem}
\usepackage{fontawesome5}
\usepackage{blindtext}
\usepackage{pgfplots}
\usepackage{pgfplotstable}

\usepackage{listings}
\usepackage{listingsutf8}
\newcommand\JSONnumbervaluestyle{\color{blue}}
\newcommand\JSONstringvaluestyle{\color{red}}

\newif\ifcolonfoundonthisline

\makeatletter
\lstdefinestyle{json}
{
  showstringspaces    = false,
  keywords            = {false,true},
  alsoletter          = 0123456789.,
  morestring          = [s]{"}{"},
  stringstyle         = \ifcolonfoundonthisline\JSONstringvaluestyle\fi,
  MoreSelectCharTable =%
    \lst@DefSaveDef{`:}\colon@json{\processColon@json},
  basicstyle          = \ttfamily,
  keywordstyle        = \ttfamily\bfseries,
}

\newcommand\processColon@json{%
  \colon@json%
  \ifnum\lst@mode=\lst@Pmode%
    \global\colonfoundonthislinetrue%
  \fi
}

\lst@AddToHook{Output}{%
  \ifcolonfoundonthisline%
    \ifnum\lst@mode=\lst@Pmode%
      \def\lst@thestyle{\JSONnumbervaluestyle}%
    \fi
  \fi
  \lsthk@DetectKeywords%
}

\lst@AddToHook{EOL}%
  {\global\colonfoundonthislinefalse}
\makeatother

\mdfdefinestyle{mystyle}{%
  rightline=true,
  innerleftmargin=10,
  innerrightmargin=10,
  outerlinewidth=3pt,
  topline=false,
  rightline=true,
  bottomline=false,
  skipabove=\topsep,
  skipbelow=\topsep
}

\usepackage{fontawesome5}
\usepackage{amsmath,amssymb}

\newcommand{\cmark}{\faCheck}
\newcommand{\xmark}{\faTimes}
\newcommand{\tmark}{\faBullseye}

\DeclareCaptionFont{black}{\color{black}}

\newenvironment{itemize*}%
 {\leftmargini=10pt\begin{itemize}%
  \setlength{\itemsep}{0pt}%
  \setlength{\parskip}{0pt}%
  }%
 {\end{itemize}}
\newenvironment{enumerate*}%
 {\begin{enumerate}%
  \setlength{\itemsep}{0pt}%
  \setlength{\parskip}{0pt}}%
 {\end{enumerate}}

\usepackage{etoolbox}
\makeatletter
\renewcommand{\@biblabel}[1]{\hfill[#1]}
\makeatother

\definecolor{myblue}{rgb}{0.9, 0.1, 0.94}
\definecolor{mygreen}{rgb}{0.64, 0.56, 0.88}
\definecolor{myyellow}{rgb}{0.68, 0.6, 0.1}
\definecolor{fancygreen}{rgb}{0.33, 0.68, 0.20}
\definecolor{salmon}{rgb}{0.94, 0.52, 0.49}
\definecolor{tablegreen}{rgb}{0.82, 0.94, 0.75}
\definecolor{tableblue}{rgb}{0.81, 0.90, 0.94}
\definecolor{tablered}{rgb}{0.97, 0.85, 0.85}
\definecolor{tableorange}{rgb}{0.96, 0.85, 0.81}
\definecolor{myteal}{rgb}{0.12, 0.56, 0.56}

\definecolor{catpurple}{HTML}{E8E0F5}
\definecolor{rowpurple}{HTML}{F4F0FB}
\definecolor{markpurple}{HTML}{8B6DB5}
\DefineVerbatimEnvironment{TextBlock}{Verbatim}{
  breaklines=true,
  breakanywhere=true,     % 必要时在任意位置断行
  breaksymbolleft={},     % 行尾不加断行符号
  obeytabs=true,
  commandchars=\\\{\}     % 可选：允许在文本中用 \{ \} 这些
}

\usepackage[most]{tcolorbox}
\tcbuselibrary{skins,breakable,listings}
\usepackage{fvextra} % 增强型 Verbatim，支持自动断行等
\usepackage{amsmath, amssymb, amsthm}

\begin{document}

%% ============================================================
%% Title and Authors
%% ============================================================

\title{\textsc{daVinci-LLM}: \\Towards the Science of Pretraining}

% \title{CTOM: Chinese Truly Open Models}

\author[1,2,3]{Yiwei Qin\textsuperscript{*}}
\author[1,2,3]{Yixiu Liu\textsuperscript{*}}
\author[1,3]{Tiantian Mi\textsuperscript{*}}
\author[1,3]{Muhang Xie\textsuperscript{*}}
\author[1,3]{Zhen Huang\textsuperscript{*}}
\author[1,2,3]{\authorcr Weiye Si}
\author[1,2,3]{Pengrui Lu}
\author[1]{Siyuan Feng}
\author[1]{Xia Wu}
\author[1]{Liming Liu}
\author[1]{\authorcr Ye Luo}
\author[1]{Jinlong Hou}
\author[1]{Qipeng Guo}
\author[1]{Yu Qiao}
\author[1,2,3]{Pengfei Liu\textsuperscript{†}}
\affil{SII \quad \textsuperscript{2}SJTU \quad \textsuperscript{3}GAIR}

\maketitle

%% ============================================================
%% Header Configuration for Title Page
%% ============================================================
\pagestyle{fancy}
\thispagestyle{fancy}
\fancyhead{}
\lhead{
  \raisebox{-0.3cm}{\includegraphics[height=0.95cm]{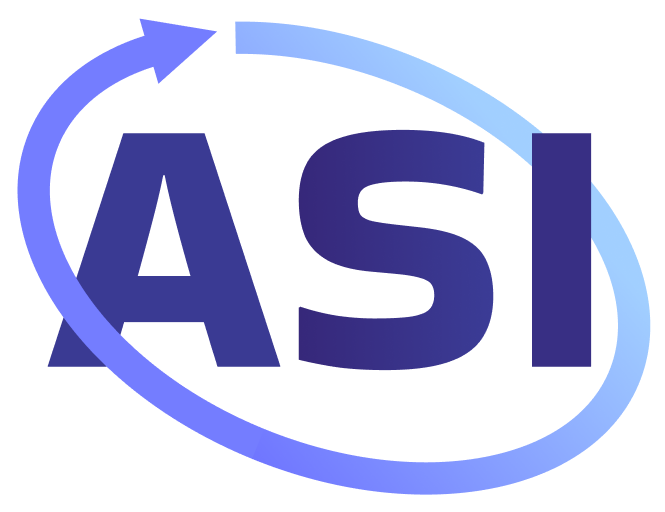}}
}
\rhead{%
  \raisebox{-0.2cm}{\includegraphics[height=0.7cm]{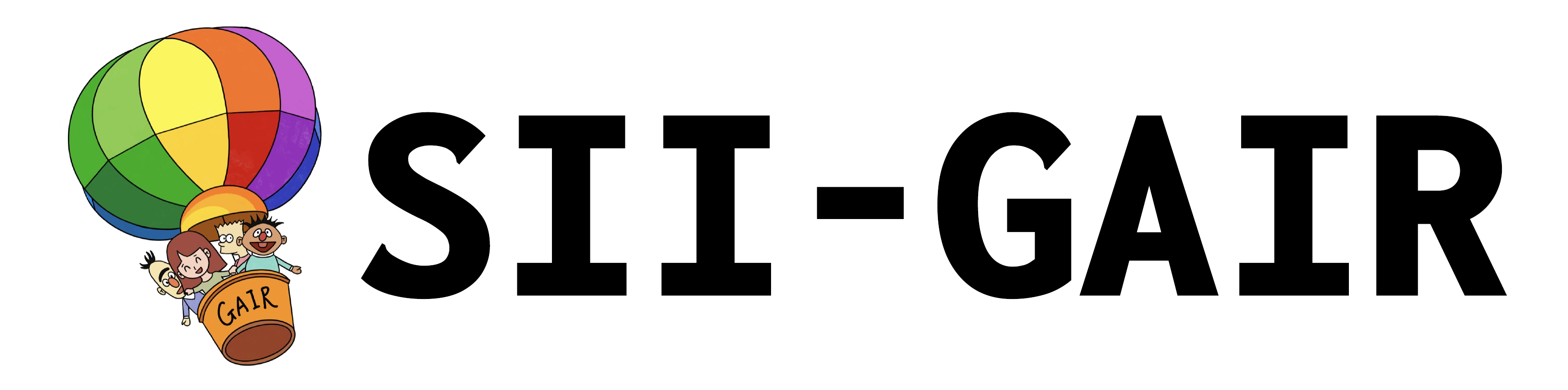}}%
}
\renewcommand{\headrulewidth}{0pt}
\setlength{\headsep}{2mm}

%% ============================================================
%% Author Footnotes
%% ============================================================
\renewcommand{\thefootnote}{}
\footnotetext{* Equal contribution.}
\footnotetext{† Corresponding author.}
\vspace{-20pt}

%% ============================================================
%% Resource Links (customize as needed)
%% ============================================================
% {\centering
% \href{https://github.com/GAIR-NLP}{\raisebox{-.15em}{\includegraphics[height=1em]{assets/sii.png}}\ SII Open Source:}
% \quad \href{https://github.com/GAIR-NLP}{\textcolor{black}\faGithub\ Code}
% \quad \href{https://huggingface.co/GAIR}{\raisebox{-.15em}{\includegraphics[height=1em]{figures/huggingface-icon.png}}\ Models}
% \quad \href{https://huggingface.co/datasets/GAIR}{\textcolor{violet}\faDatabase\ Datasets}
% \par}

%% ============================================================
%% Main Content
%% ============================================================
% !TEX root = main.tex
% maintext.tex - Main content of the paper
% This file contains all sections of the paper

{\centering
% \href{https://github.com/sii-research/GAIR}{\raisebox{-.15em}{\includegraphics[height=1em]{assets/sii.png}}\ SII Open Source:}
\quad \href{https://github.com/GAIR-NLP/daVinci-LLM}{\textcolor{black}\faGithub\ daVinci-LLM}
\quad \href{https://huggingface.co/SII-GAIR-NLP/davinci-llm-model}{\raisebox{-.15em}{\includegraphics[height=1em]{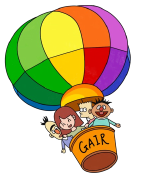}}\ daVinci-LLM Model}
\quad \href{https://huggingface.co/datasets/SII-GAIR-NLP/davinci-llm-data}{{\textcolor{violet}\faDatabase}\ daVinci-LLM Data}
\par}

\vspace{5pt}

%% ============================================================
%% Abstract
%% ============================================================
\begin{tcolorbox}[
  colback=blue!5,
  colframe=blue!5,
  boxrule=0pt,
  arc=5pt,
  left=2pt,
  right=2pt,
  top=2pt,
  bottom=2pt
]
\begin{abstract}
The foundational pretraining phase determines a model’s capability ceiling, as post-training struggles to overcome capability foundations established during pretraining, yet it remains critically under-explored. This stems from a structural paradox: organizations with computational resources operate under commercial pressures that inhibit transparent disclosure, while academic institutions possess research freedom but lack pretraining-scale computational resources. \textsc{daVinci-LLM} occupies this unexplored intersection, combining industrial-scale resources with full research freedom to advance the science of pretraining.
We adopt a fully-open paradigm that treats openness as scientific methodology, releasing complete data processing pipelines, full training processes, and systematic exploration results. Recognizing that the field lacks systematic methodology for data processing, we employ the Data Darwinism framework—a principled L0-L9 taxonomy from filtering to synthesis. We train a 3B-parameter model from random initialization across 8T tokens using a two-stage adaptive curriculum that progressively shifts from foundational capabilities to reasoning-intensive enhancement. Through 200+ controlled ablations, we establish that: processing depth systematically enhances capabilities, establishing it as a critical dimension alongside volume scaling; different domains exhibit distinct saturation dynamics, necessitating adaptive strategies from proportion adjustments to format shifts; compositional balance enables targeted intensification while preventing performance collapse; how evaluation protocol choices shape our understanding of pretraining progress. By releasing the complete exploration process, we enable the community to build upon our findings and systematic methodologies to form accumulative scientific knowledge in pretraining.

\end{abstract}
\end{tcolorbox}

%% ============================================================
%% Teaser Figure
%% ============================================================
\begin{figure*}[!ht]
  \centering
  \includegraphics[width=0.915\linewidth]{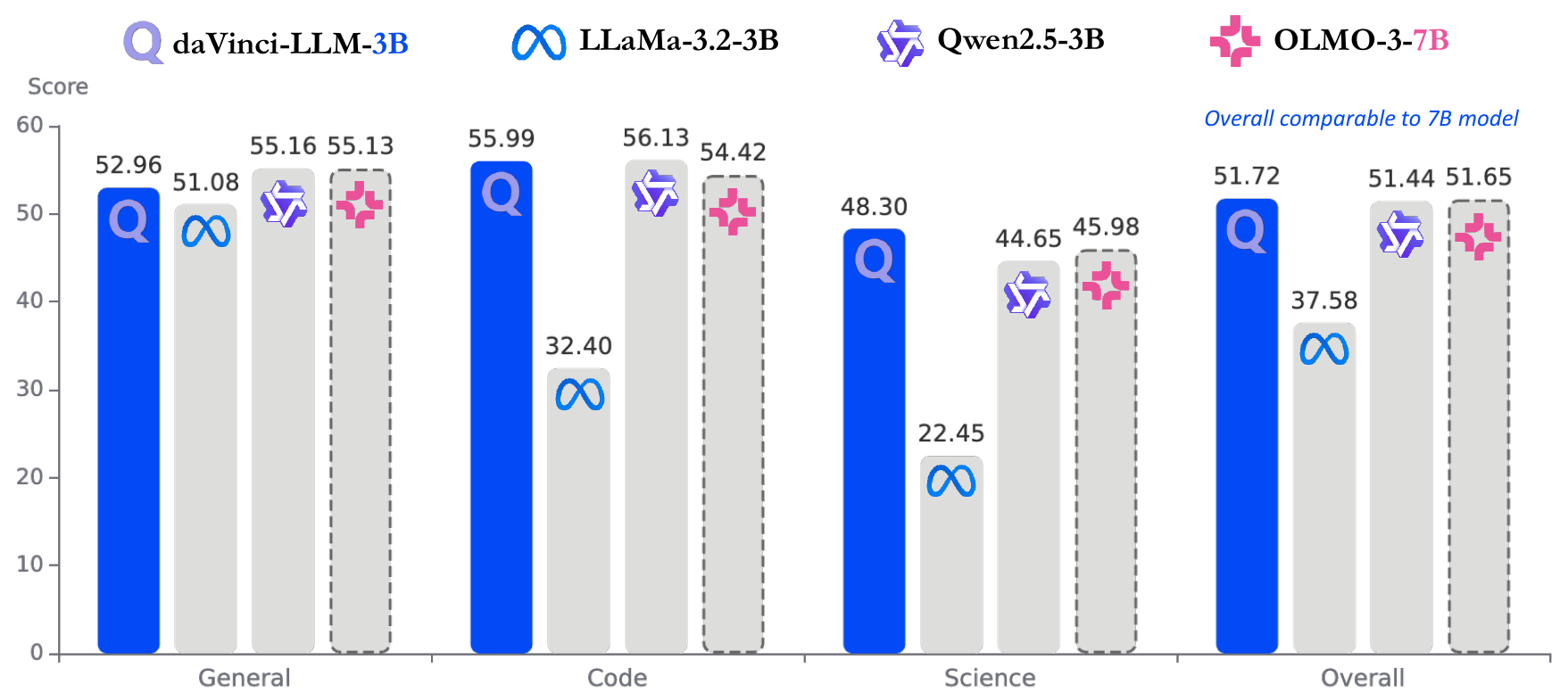}
  \caption{Performance comparison of \textsc{daVinci-LLM-3B} against baseline models with score across three capability domains, and overall score comparable to OLMo-3-7B.}
  \label{fig:teaser}
\end{figure*}

\clearpage  % 强制分页,确保目录独占一页

%% ============================================================
%% Table of Contents
%% ============================================================
\vspace*{20pt}  % 增加目录整体距离页面上边距的距离
\tableofcontents

\clearpage  % 确保正文从新页开始

\pagestyle{fancy}
\lhead{\rightmark}
\renewcommand{\headrulewidth}{0.7pt}
\setlength{\headsep}{5mm}

% Reset footnote numbering for main content
\renewcommand{\thefootnote}{\arabic{footnote}}
\setcounter{footnote}{0}

%% ============================================================
%% 1. Introduction
%% ============================================================
\section{Introduction}
\label{sec:intro}

\begin{figure}
    \centering
    \includegraphics[width=0.999\linewidth]{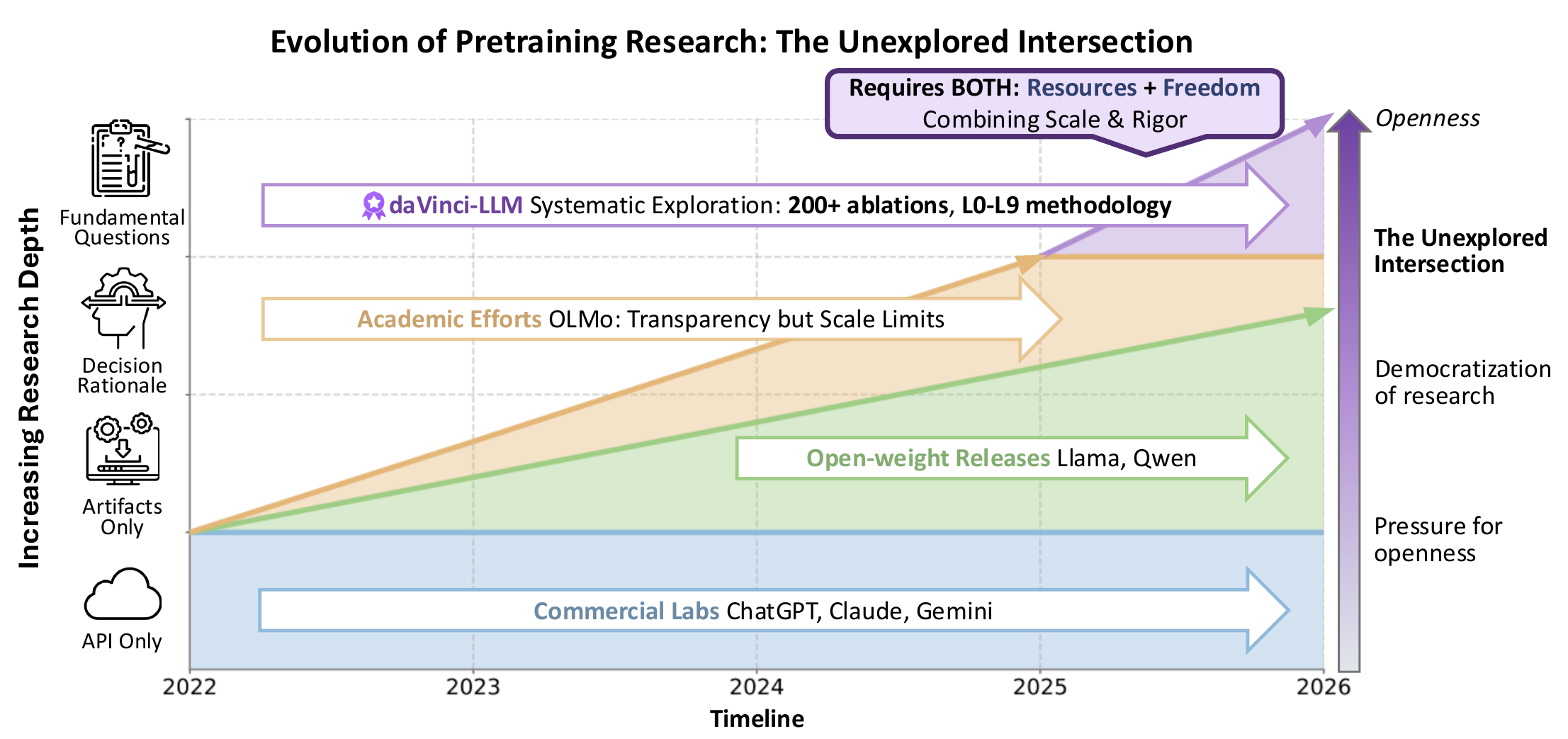}
    \caption{Evolution of pretraining research depth across institutional structures. The y-axis represents research depth from surface-level artifacts (API-only access) to fundamental scientific questions. The x-axis shows the temporal progression from 2022 to 2026. Commercial entities (blue) possess computational resources but remain constrained to API-level access due to competitive pressures. Open-weight releases (green, e.g., Llama, Qwen) provide model artifacts but withhold design rationale and negative results. Academic efforts (orange, e.g., OLMo) achieve transparency and research freedom but face severe scale limitations—making systematic exploration with 200+ configurations structurally infeasible. 
    %The top tier, requiring both computational resources for exhaustive ablations and research freedom to publish comprehensive findings, remains largely unexplored. 
    %\textsc{daVinci} (purple) occupies this intersection, enabling systematic investigation of fundamental questions: When do training dimensions saturate? Can data quality substitute for model scale? Why do balanced mixtures outperform specialized configurations?
    The top tier remains largely unexplored, as it requires the rare alignment of large-scale computational resources with the research freedom to publish comprehensive findings. 
    \textsc{daVinci-LLM} (purple) occupies this intersection, conducting the extensive ablations and systematic disclosures necessary to advance the science of pretraining. By releasing the complete decision-making logic alongside the model weights, we bridge the structural gap between industrial scale and scientific transparency.
    }
    \label{fig:placeholder}
\end{figure}

The large language model ecosystem has evolved into a stratified landscape characterized by varying levels of transparency. At the most opaque tier lie closed-source commercial models (GPT~\cite{OpenAI2025GPT5.2, achiam2023gpt}, Claude~\cite{Anthropic2026ClaudeOpus4.6}, Gemini~\cite{Google2025Gemini3, comanici2025gemini}) accessible only through APIs. The intermediate tier comprises open-weight models (LLaMA~\cite{dubey2024llama}, Qwen~\cite{qwen2,qwen2.5,qwen3, qwen3.5}, DeepSeek~\cite{guo2025deepseek}) that release checkpoints but withhold critical pretraining details—data compositions, mixture ratios, and training dynamics remain largely undisclosed. At the foundation are fully-open efforts that release complete training specifications. Despite this stratification, the field remains dominated by the first two paradigms. Yet as the ATOM (American Truly Open Models) Project~\cite{lambert2025atom} emphasizes, open language models are crucial for long-term competition by enabling the broader research community to pursue long-horizon transformative innovations rather than only immediate deployment priorities. However, research attention has concentrated disproportionately on accessible post-training techniques~\cite{peng2023instruction,ouyang2022training,jaech2024openai, guo2025deepseek, qin2024o1, huang2024o1, ye2025limo,xia2025generative,jimenez2023swe, liu2025alphago,liu2023pre,hua2025context}, while the foundational pretraining phase, which determines a model’s capability ceiling, remains critically under-explored.

This imbalance stems from structural constraints that create a fundamental paradox. Organizations with computational resources for large-scale pretraining operate under commercial pressures that favor rapid deployment over systematic exploration and inhibit transparent disclosure of training processes. Academic institutions possess research freedom but lack pretraining-scale infrastructure—even well-funded efforts like OLMo~\cite{olmo20242,olmo2025olmo} face severe scale limitations that make large-scale systematic exploration structurally infeasible, while confronting persistent challenges in sustaining both computational resources and key research personnel. The consequence is stark: precisely when emerging evidence demonstrates that pretraining choices fundamentally shape downstream capabilities~\cite{akter2025front}, the community has limited ability to systematically investigate the principles governing how models acquire and organize knowledge during pretraining. 
%Post-training techniques can refine and align model behavior, but they cannot transcend the knowledge boundaries, reasoning patterns, and compositional structures established during pretraining.
Post-training techniques can refine and align model behavior, but struggle to fundamentally overcome the capability foundations established during pretraining—research~\cite{ovadia2024fine,pletenev2025much} shows that pretraining advantages are amplified rather than compensated for in subsequent training phases.

We are positioned to address this gap by combining computational resources for billion-parameter training with the research freedom to investigate fundamental questions and publish comprehensive findings. Towards the science of pretraining, we adopt a fully-open paradigm (as shown in Table~\ref{tab:openness_comparison}) that treats openness itself as scientific methodology, releasing not only model weights but complete training trajectories, data specifications, and ablation results documenting both what works and what fails. Our work is structured around three pillars, each contributing to transparency and reproducibility.

\begin{table*}[t]
  \centering
  \small
  \caption{Transparency comparison across state-of-the-art LLMs. Unlike existing models, \textsc{daVinci} provides the complete scientific process, enabling the systematic investigation of pretraining dynamics. Symbols: \cmark~fully open, \tmark~ patial disclosure or not released, \xmark~not disclosed.}
  \label{tab:openness_comparison}
  \setlength{\tabcolsep}{6pt}
    \begin{tabular}{
    p{1.3cm} l|
    >{\centering\arraybackslash}p{1.5cm}|
    >{\centering\arraybackslash}p{1.5cm}|
    >{\centering\arraybackslash}p{1.5cm}|
    >{\columncolor{olmopink!7}\centering\arraybackslash}p{1.5cm}|
    >{\columncolor{blue!7}\centering\arraybackslash}p{1.5cm}
    }
    \toprule
    {} & {\textbf{Dimension}} & \textbf{Llama 3} & \textbf{Qwen 3} & \textbf{YuLan} & \textbf{OLMo 3} & \textbf{daVinci} \\
    \midrule
    
    % ---------------- Model Artifacts (4 rows) ----------------
    \multirow{4}{*}{\shortstack[c]{\textsc{Model}\\\textsc{Artifacts}}}
      & Model Weights & \cmark & \cmark & \cmark & \cmark & \cmark \\
      & Training Code & \xmark & \xmark & \cmark & \cmark & \cmark \\
      & Training Logs & \xmark & \xmark & \xmark & \cmark & \cmark \\
      & Intermediate Checkpoints & \xmark & \xmark & \xmark & \cmark & \cmark \\
    \midrule
    
    % ---------------- Data Transparency (4 rows) ----------------
    \multirow{4}{*}{\shortstack[c]{\textsc{Data}\\\textsc{Openness}}}
      & Data Composition & \tmark & \tmark & \cmark & \cmark & \cmark \\
      & Processing Pipeline & \tmark & \tmark & \cmark & \cmark & \cmark \\
      & Full Training Data & \xmark & \xmark & \tmark & \cmark & \cmark \\
      & \textit{Processing Methodology} & \xmark & \xmark & \xmark & \xmark & \textit{\cmark~(L0-9)} \\
    \midrule
    
    % ---------------- Scientific Process (4 rows) ----------------
    \multirow{4}{*}{\shortstack[c]{\textsc{Scientific}\\\textsc{Process}}}
      & Pretraining Ablations & \cmark & \xmark & \cmark & \cmark & \cmark \\
      & \textit{Mixture Rationale} & \xmark & \xmark & \xmark & \tmark & \textit{\cmark} \\
      & \textit{Decision Transparency} & \xmark & \xmark & \xmark & \tmark & \textit{\cmark} \\
      & \textit{Negative Results} & \xmark & \xmark & \xmark & \xmark & \textit{\cmark} \\
    \bottomrule
  \end{tabular}
\end{table*}

\paragraph{Data} (Section~\ref{sec:data}) 
Recognizing that data quality fundamentally determines outcomes yet the field lacks systematic methodology for processing decisions, we adopt the Data Darwinism framework~\cite{qin2026data}, a principled L0-L9 taxonomy organizing operations from basic filtering to content transformation to knowledge synthesis. We release our complete data processing pipeline and the processed datasets themselves. Our data pool spans general web text, code, science, and QA domains, with each source explicitly annotated by its Darwin Level to systematically organize our processing decisions and identify where further quality enhancement is feasible.

\paragraph{Training Recipe} (Section~\ref{sec:training-recipe}) 
We train a 3B-parameter model from random initialization across 8T tokens using a two-stage curriculum: Stage 1 (6T tokens) establishes broad foundational capabilities through diverse web-scale corpora with progressive data adjustment, while Stage 2 (2T tokens) shifts toward reasoning-intensive enhancement by introducing large-scale structured QA data alongside continued exposure to code and science domains. We release all intermediate checkpoints at 5k-step intervals, complete hyperparameters, training logs, and the evolution of data mixture compositions across stages, providing a complete developmental trajectory from initialization to final model.

\paragraph{Exploration} (Section~\ref{sec:exploration})  
Towards the science of pretraining, we transform design decisions into systematically verifiable research questions. Through 200+ controlled ablations, we systematically investigate key questions spanning four thematic areas about pretraining: (1) \textbf{Data Processing Depth}: How does hierarchical data processing systematically enhance model capabilities? (2) \textbf{Training Dynamics}: How should data strategies and curriculum adapt during training? (3) \textbf{Data Mixture Design}: How to balance targeted enhancement with general capability preservation? (4) \textbf{Evaluation Validity}: Which evaluation protocols reliably measure base model pretraining progress? 
%We document both successful configurations and failed experiments, providing empirical evidence for data processing, training dynamics, mixture design, and evaluation decisions.
Through these investigations, we find that processing depth systematically enhances capabilities, different domains exhibit distinct training dynamics requiring adaptive strategies, and mixture design critically determines the balance between capability enhancement and preservation. 
We document both successful configurations and failed experiments, providing empirical evidence to inform each dimension of pretraining decisions.

Our contributions span three dimensions: 
(1) \textbf{Complete research materials}: We release model weights, intermediate checkpoints, and processed datasets, enabling researchers to analyze capability emergence, reproduce training processes, and conduct extended investigations. 
(2) \textbf{Question-driven pretraining science}: By transforming key pretraining decisions into systematically verifiable research questions, we provide empirical understanding of data quality, training dynamics, and mixture strategies, offering reference points for researchers facing similar decisions.
(3) \textbf{Transferable methodological foundations}: The Data Darwinism framework, systematic exploration methods, and complete documentation of successes and failures constitute reusable research infrastructure, enabling the community to build upon documented boundary conditions and form accumulative scientific knowledge in pretraining.

To facilitate reproducibility and support the research community, we 
publicly release all datasets produced through our own processing and 
synthesis pipelines, together with the complete curation toolkit comprising 
all prompts and processing code. We further release all intermediate and 
final model checkpoints saved throughout training, training logs, 
ablation results, and our full evaluation suite. We hope these releases 
can help the community better understand, reproduce, and build upon our 
work.

%% ============================================================
%% 3. Stage~1: General Pretraining
%% ============================================================
\section{Data: What We Used and How We Processed It}
\label{sec:data}
\begin{figure*}[t]
  \centering
  \includegraphics[width=0.999\linewidth]{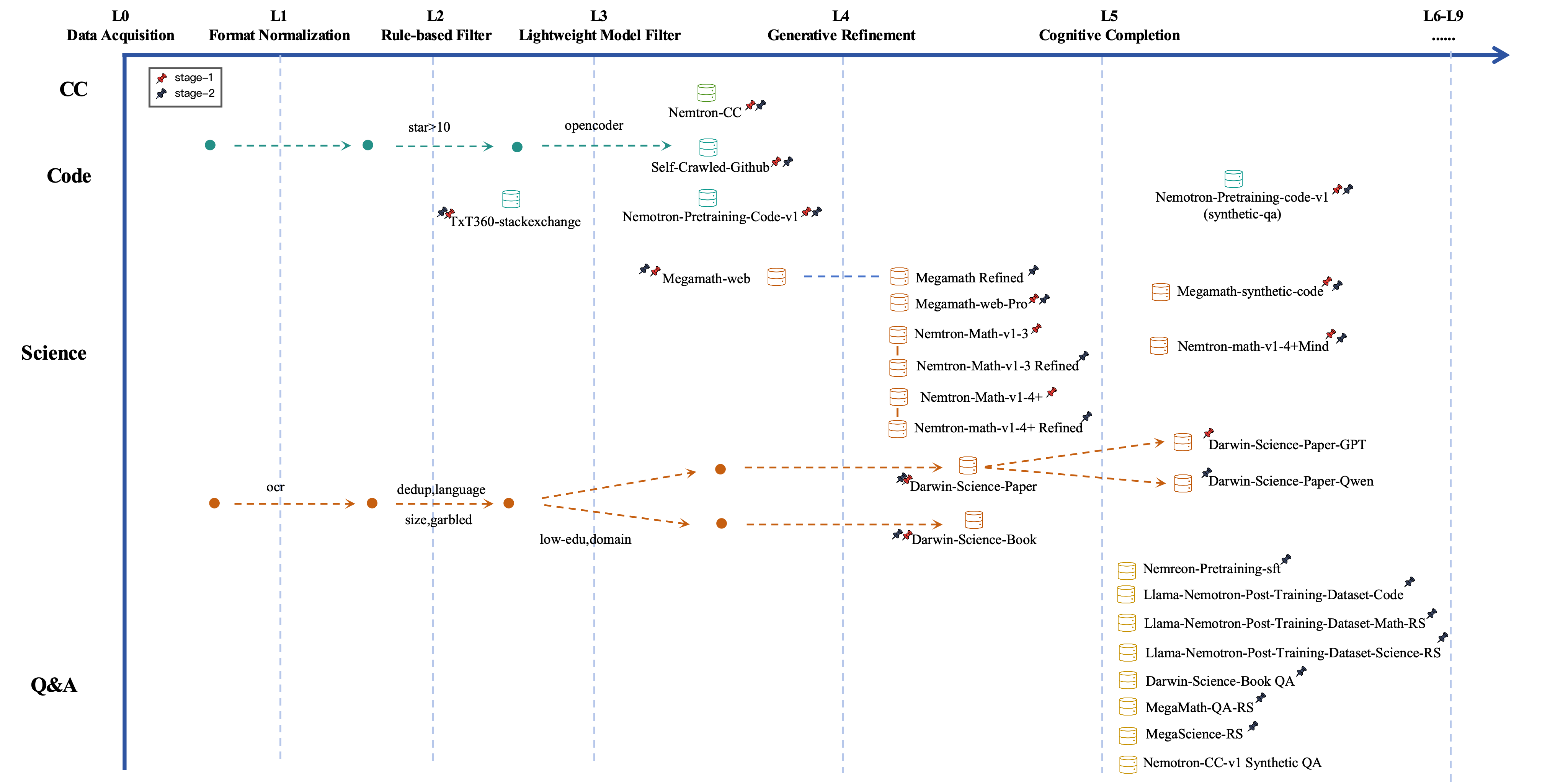}
  \caption{Mapping of our pretraining data sources onto the Data Darwinism L0--L9 
taxonomy across different training stage.}
  \label{fig:data-darwinism-level}
\end{figure*}

% The quality and composition of pretraining data are fundamental to the capabilities of the resulting model. To reason systematically about data quality across our heterogeneous corpus, we adopt the Data Darwinism framework~\cite{qin2026data}, under which every data source is annotated with its corresponding Darwin Level and further processed where meaningful quality gains can be achieved. Our pretraining corpus is organized into four categories: General, Code, Science, and QA, drawn from a combination of existing open-source datasets and data collected directly from public repositories. Table~\ref{tab:data-pools} summarizes the composition and scale of the full data pool.

Data transparency remains one of the most underserved dimensions of open pretraining research. While open-weight releases have made model checkpoints accessible, the data decisions underlying those models, including what sources were selected and how they were processed, remain largely opaque. We address this directly: beyond releasing our complete training corpus, we document the processing depth of every data source through the Data Darwinism framework (L0–L9)~\cite{qin2026data}, making our curation decisions explicit and traceable. Our pretraining corpus is organized into four categories of General, Code, Science, and QA, drawn from a combination of existing open-source datasets and data collected directly from public repositories, with each source annotated by its corresponding Darwin Level and further processed where meaningful quality gains can be achieved.

% \begin{table*}[t]
%   \centering
%   \caption{Stage~1/Stage~2 data pools (token counts) grouped by high-level categories. `QA-short'' is newly introduced in Stage~2.\yiwei{It's not clear what source pool each category contains. Consider to add some horizon lines.}\muhang{fixed}}
%   \label{tab:data-pools}
%   \begin{tabular}{l l r r}
%     \toprule
%     Category & Source pool & Stage~1 size & Stage~2 size \\
%     \midrule
%     \multirow{1}{*}{CC (web)} 
%       & Nemotron-CC-v1 & 4.28T & 4.28T \\
%     \midrule

%     \multirow{3}{*}{Code} 
%       & Self-crawled GitHub & 187B & 187B \\
%       & Nemotron-Pretraining-Code-v1 & 391B & 391B \\
%       & txt360 StackExchange & 20B & 20B \\
%     \midrule

%     \multirow{2}{*}{Math} 
%       & Megamath & 286B & 230B \\
%       & Nemotron-CC-Math & 207B & 189B \\
%     \midrule

%     \multirow{2}{*}{Science} 
%       & Books & 252B & 252B \\
%       & Papers & 505B & 655B \\
%     \midrule

%     \multirow{3}{*}{QA-short} 
%       & Nemotron-pretraining-sft-v1 & -- & 171B \\
%       & Book-derived QA & -- & 46B \\
%       & RS open-source QA & -- & 26B \\
%     \midrule

%     Total & -- & 6.128T & 6.447T \\
%     \bottomrule
%   \end{tabular}
% \end{table*}

\subsection{Data Darwinism Framework}
\label{sec:data_darwin}
The quality of pretraining data is fundamentally shaped by how it has been collected, filtered, and processed. Yet the field currently lacks a systematic framework for categorizing and comparing these operations, making it difficult to reason about quality differences across heterogeneous data sources. Data Darwinism was proposed precisely to address this gap, organizing data processing operations into a principled ten-level taxonomy (L0–L9). We adopt this framework to assess the processing depth of every data source in our training corpus—not only to understand where each dataset currently stands in the hierarchy, but also to identify whether further processing is feasible and whether the potential quality gains justify the additional investment.

Underlying this hierarchy is a coherent evolutionary logic: data processing begins with the selection and preservation of existing content, progressively moves toward active rewriting and enrichment, and ultimately reaches the capacity to synthesize entirely new content from scratch. In parallel, the agents driving these operations shift from hand-crafted deterministic rules, to lightweight classification models, to frontier large language models capable of reasoning and generation. We now describe each level in turn:

\textit{L0: Data Acquisition.} At this foundational stage, raw data is gathered from diverse sources including web crawls, PDF repositories, code platforms, and curated databases. The collected data exists in highly variable formats such as HTML, PDF, and binary files, and typically contains significant noise and duplication. The primary challenges at this level lie in achieving broad coverage, maintaining data provenance, and managing large-scale storage infrastructure.

\textit{L1: Format Normalization.} At this stage, heterogeneous raw data is converted into unified, training-ready text representations. For document-based sources, key operations include OCR processing of scanned PDFs and HTML parsing to extract clean content. No content is filtered here; the goal is uniform processability while preserving structural fidelity across sources.

\textit{L2: Rule-based Filtering.} This is the first stage of quality control, where deterministic pattern-based rules are applied to remove objectively identifiable problematic content: near-duplicates detected via MinHash LSH, excessively short or malformed text, non-target languages, and garbled text from encoding errors. The approach requires no learned models, runs efficiently on CPU infrastructure, and achieves substantial volume reduction while remaining fully interpretable.

\textit{L3: Lightweight Model Filtering.} Unlike L2, which operates on surface patterns, this stage introduces semantic-level quality assessment using pretrained lightweight classifiers. Tasks such as educational value scoring, domain identification, and document type classification enable more nuanced filtering decisions than rules alone can support. Importantly, this remains a pure selection stage: documents are retained or discarded based on predicted quality, but their content is never modified.

\textit{L4: Generative Refinement.} This stage marks a qualitative shift from selection to active, model-driven transformation. Medium-to-large generative models are deployed to purify content by removing structural noise such as navigation elements, reference lists, OCR artifacts, and formatting defects, as well as repairing fragmented text, while strictly adhering to the original content. A critical constraint is that this stage must act as a faithful refiner: no external knowledge may be introduced, and the output must remain semantically equivalent to the input.

\textit{L5: Cognitive Completion.} At this stage, frontier LLMs enrich data by making implicit reasoning explicit. Research and technical documents are typically written for expert audiences, characterized by compressed logical steps, assumed background knowledge, and implicit derivations that create a learnability gap for language models. This stage bridges that gap through reasoning reconstruction, terminological explication, and pedagogical bridging, producing content that retains full scientific fidelity while substantially lowering the cognitive barrier for model internalization.

\textit{L6–L9: Higher-Order Synthesis.} The upper levels of the hierarchy address increasingly ambitious forms of data generation. Contextual Completion (L6) expands documents by integrating external references and background knowledge to create self-contained artifacts. Environment Synthesis (L7) constructs executable environments in which data objects can be validated through actual execution. Ecosystem Synthesis (L8) builds dynamic multi-agent systems where diverse intelligent entities interact and generate emergent data through sustained collaboration. World Synthesis (L9) represents the theoretical apex of the framework, aspiring to construct comprehensive simulated worlds as a source of essentially unlimited synthetic training data. 

It is worth noting that these levels are not mutually exclusive one-time passes: any operation at a given level can be applied multiple times, with different models, prompts, or parameters, each targeting different aspects of quality to achieve progressively deeper processing. Furthermore, the ordering of operations need not strictly follow the level hierarchy---for instance, a dataset may undergo L4 Generative Refinement before being subjected to L3 model-based filtering, if such an ordering better suits the characteristics of the source data.

Throughout this work, we annotate each data source with its corresponding Darwin Level to make our curation decisions explicit and transparent. For a subset of sources, we further apply higher-level processing operations to actively improve their quality. This allows us to reason systematically about the depth of processing across our entire corpus, and to identify where additional effort is most likely to yield meaningful gains.

\subsection{Data Pool}
We curate our pretraining corpus from five major categories: CC, Code, Math, Science,
and QA. For each data source, we annotate its corresponding Darwin Level to make our
curation decisions explicit and transparent. Table~\ref{tab:data-pools} summarizes the composition and scale of the full data pool across training stages, and Figure~\ref{fig:data-darwinism-level} provides an overview of the Darwin Level assigned to each data source.

\subsubsection{General}
General web text forms the backbone of our pretraining corpus, providing broad coverage across topics, writing styles, and knowledge domains. Our general corpus is primarily drawn from Common Crawl, with quality filtering applied to balance coverage and data quality.
\paragraph{Nemotron-CC-v1.}
We adopt the non-synthetic portion of \textit{Nemotron-CC-v1}~\cite{su2024nemotron} as our web-scale general corpus. \textit{Nemotron-CC-v1} is built from 99 snapshots of Common Crawl and undergoes text extraction, English language filtering, and global deduplication, followed by an ensemble of classifiers combining educational value and informativeness signals that scores each document and groups them into five quality tiers based on downstream task performance---enabling precise control over the quality--diversity trade-off across training stages. This complete processing pipeline places \textit{Nemotron-CC-v1} at Darwin Level L3, contributing approximately 4.28T tokens to our training corpus.

\begin{table}[t]
\centering
\caption{Data pool composition and token allocation across training stages.
Shaded category rows show each category's total pool size and the tokens
allocated per stage; data source rows indicate whether the source is used
(\checkmark) or not used (--) in each stage.}
\label{tab:data-pools}
\resizebox{\textwidth}{!}{%
\begin{tabular}{
  l
  r
  >{\centering\arraybackslash}p{1.2cm}
  >{\centering\arraybackslash}p{1.6cm}
  >{\centering\arraybackslash}p{1.6cm}
  >{\centering\arraybackslash}p{1.6cm}
  >{\centering\arraybackslash}p{1.6cm}
}
\toprule
\multirow{2}{*}{\textbf{Data Source}} &
\multirow{2}{*}{\textbf{Pool Size}} &
\multirow{2}{*}{\textbf{Level}} &
\multicolumn{2}{c}{\textbf{Stage 1}} &
\multicolumn{2}{c}{\textbf{Stage 2}} \\
\cmidrule(lr){4-5}\cmidrule(lr){6-7}
& & & \textbf{Stage 1-1} & \textbf{Stage 1-2} & \textbf{Stage 2-1} & \textbf{Stage 2-2} \\

%% ── GENERAL ──────────────────────────────────────────────────────────────────
\midrule
\rowcolor{catblue}
\textbf{General} & \textbf{4.28T} & &
\makecell{\small\textbf{2.73T} \\ \small\textit{(68.2\%)}} &
\makecell{\small\textbf{1.11T} \\ \small\textit{(55.42\%)}} &
\makecell{\small\textbf{100B} \\ \small\textit{(10\%)}} &
\makecell{\small\textbf{188.4B} \\ \small\textit{(18.84\%)}} \\
\rowcolor{white}
\quad Nemotron-CC-v1              & 4.28T & L3 & \checkmark & \checkmark & \checkmark & \checkmark \\

%% ── CODE ─────────────────────────────────────────────────────────────────────
\midrule
\rowcolor{catblue}
\textbf{Code} & \textbf{598B} & &
\makecell{\small\textbf{381B} \\ \small\textit{(9.53\%)}} &
\makecell{\small\textbf{233B} \\ \small\textit{(11.66\%)}} &
\makecell{\small\textbf{300} \\ \small\textit{(30\%)}} &
\makecell{\small\textbf{26.1B} \\ \small\textit{(2.61\%)}} \\
\rowcolor{rowblue}
\quad Self-Crawled GitHub                             & 187B & L3 & \checkmark & \checkmark & \checkmark         & \checkmark         \\
\rowcolor{white}
\quad Nemotron-Pretraining-Code-v1-non-synthetic      & 220B & L3 & \checkmark & \checkmark & \checkmark & \checkmark \\
\rowcolor{rowblue}
\quad Nemotron-Pretraining-Code-v1-synthetic-code     & 171B & L5 & \checkmark & \checkmark & \checkmark & \checkmark \\
\rowcolor{white}
\quad TxT360-Stack-Exchange                           &  20B & L2 & \checkmark & \checkmark & \checkmark         & \checkmark         \\

%% ── SCIENCE ──────────────────────────────────────────────────────────────────
\midrule
\rowcolor{catblue}
\textbf{Science} & \textbf{1.94T} & &
\makecell{\small\textbf{891B} \\ \small\textit{(22.27\%)}} &
\makecell{\small\textbf{658B} \\ \small\textit{(32.92\%)}} &
\makecell{\small\textbf{300B} \\ \small\textit{(30\%)}} &
\makecell{\small\textbf{85.5B} \\ \small\textit{(8.85\%)}} \\
\rowcolor{rowblue}
\quad MegaMath-Web                                    & 231B & L3 & \checkmark & \checkmark & --         & --         \\
\rowcolor{white}
\quad MegaMath-Web-Pro                                &  13B & L4 & \checkmark & \checkmark & \checkmark & \checkmark \\
\rowcolor{rowblue}
\quad MegaMath Refined                                & 176B & L4 & -- & -- & \checkmark & \checkmark \\
\rowcolor{white}
\quad MegaMath-Synth-Code                             &   5B & L5 & \checkmark & \checkmark & \checkmark & \checkmark \\
\rowcolor{rowblue}
\quad Nemotron-CC-Math-v1-3                           &  81B & L4 & \checkmark & \checkmark & --         & --         \\
\rowcolor{white}
\quad Nemotron-CC-Math-v1-4+                          &  52B & L4 & \checkmark & \checkmark & --         & --         \\
\rowcolor{rowblue}
\quad Nemotron-CC-Math-v1-4+-MIND                     &  74B & L5 & \checkmark & \checkmark & \checkmark & \checkmark \\
\rowcolor{white}
\quad Nemotron-CC-Math-v1-3 Refined                   &  68B & L4 & -- & -- & \checkmark & \checkmark \\
\rowcolor{rowblue}
\quad Nemotron-CC-Math-v1-4+ Refined                  &  47B & L4 & -- & -- & \checkmark & \checkmark \\
\rowcolor{white}
\quad Darwin-Science-Book                             & 251B & L4 & \checkmark & \checkmark & \checkmark & \checkmark \\
\rowcolor{rowblue}
\quad Darwin-Science-Paper                            & 215B & L4 & \checkmark & \checkmark & \checkmark         & \checkmark         \\
\rowcolor{white}
\quad Darwin-Science-Paper-GPT                        & 290B & L5 & \checkmark & \checkmark & -- & -- \\
\rowcolor{rowblue}
\quad Darwin-Science-Paper-Qwen                       & 440B & L5 & -- & -- & \checkmark & \checkmark \\

%% ── QA ───────────────────────────────────────────────────────────────────────
\midrule
\rowcolor{catblue}
\textbf{QA} & \textbf{734B} & &
\makecell{\small\textbf{0} \\ \small\textit{(0\%)}} &
\makecell{\small\textbf{0} \\ \small\textit{(0\%)}} &
\makecell{\small\textbf{300B} \\ \small\textit{(30\%)}} &
\makecell{\small\textbf{700B} \\ \small\textit{(70\%)}} \\
\rowcolor{rowblue}
\quad Nemotron-CC-v1 Synthetic QA                     & 492B & L5 & -- & -- & -- & -- \\
\rowcolor{white}
\quad Nemotron-Pretraining-SFT-v1-Code                &  21B & L5 & -- & -- & \checkmark & \checkmark \\
\rowcolor{rowblue}
\quad Nemotron-Pretraining-SFT-v1-Math                & 138B & L5 & -- & -- & \checkmark & \checkmark \\
\rowcolor{white}
\quad Nemotron-Pretraining-SFT-v1-General             &  12B & L5 & -- & --& \checkmark & \checkmark \\
\rowcolor{rowblue}
\quad Llama-Nemotron-Post-Training-Dataset-Code       &   5B & L5 & -- & --& \checkmark & \checkmark \\
\rowcolor{white}
\quad Llama-Nemotron-Post-Training-Dataset-Math-RS    &  10B & L5 & -- & -- & \checkmark & \checkmark \\
\rowcolor{rowblue}
\quad Llama-Nemotron-Post-Training-Dataset-Science-RS & 0.4B & L5 & -- & -- & \checkmark & \checkmark \\
\rowcolor{white}
\quad MegaMath-QA-RS                                  &   9B & L5 & -- & -- & \checkmark & \checkmark \\
\rowcolor{rowblue}
\quad MegaScience-RS                                  &   1B & L5 & -- & -- & \checkmark & \checkmark \\
\rowcolor{white}
\quad Darwin-Science-Book QA                          &  46B & L5 & -- & -- & \checkmark & \checkmark \\

%% ── TOTAL ────────────────────────────────────────────────────────────────────
\midrule
\rowcolor{catblue}
\textbf{Total} & \textbf{$\sim$7.58T} & &
\textbf{4T} & \textbf{2T} & \textbf{1T} & \textbf{1T} \\
\bottomrule
\end{tabular}%
}
\end{table}

\subsubsection{Code}
Code data equips the model with structured, executable knowledge and exposes it to formal reasoning patterns grounded in programming languages. Our code corpus is assembled from a combination of self-crawled GitHub repositories and existing open-source code datasets, spanning both real-world source code and synthetically generated coding examples.
\paragraph{Self-Crawled GitHub.}
We directly crawled public GitHub repositories and applied a minimum threshold of 10 stars per repository as an initial rule-based quality gate, ensuring that only repositories with demonstrated community adoption are retained. The collected source files were then organized and passed through OpenCoder's filtering pipeline~\cite{huang2025opencoderopencookbooktoptier}, which removes low-quality or non-informative code files through lightweight model-based assessment, bringing the dataset to Darwin Level L3. This process yields approximately 187B tokens.

\paragraph{Nemotron-Pretraining-Code-v1.}
We incorporate two complementary subsets from \textit{Nemotron-Pretraining-Code-v1}~\cite{nvidia2025nvidianemotronnano2}, after cross-deduplication with our self-crawled GitHub collection.

\begin{itemize}
    \item \textit{Nemotron-Pretraining-Code-v1-non-synthetic.}
    This subset consists of real-world source code 
    collected from GitHub. Repositories undergo 
    license-based filtering to retain only 
    permissively licensed code, followed by both 
    exact and fuzzy deduplication to address the 
    pervasive cross-repository redundancy 
    characteristic of open-source ecosystems. The 
    OpenCoder filtering pipeline is then applied to 
    remove files that are low-quality or detrimental 
    for LLM pretraining, bringing this subset to 
    Darwin Level L3 and yielding approximately 220B 
    tokens.

    \item \textit{Nemotron-Pretraining-Code-v1-synthetic-code.}
    This subset is generated by prompting an LLM to 
    produce question-answer pairs grounded in short 
    code snippets, where the model is asked to both 
    formulate and solve coding questions across 11 
    programming languages. The resulting natural 
    language--code interleaved pairs are filtered 
    post-hoc through language-specific heuristics 
    such as Python AST parsing, reaching Darwin 
    Level L5 and yielding approximately 171B tokens.
\end{itemize}

\paragraph{TxT360-Stack-Exchange.}
We incorporate \textit{txt360-stack-exchange}~\cite{txt360data2024} as a source of technical community discourse. This dataset is compiled from the Stack Exchange network, covering 364 sub-communities spanning programming, mathematics, science, and numerous other technical domains. Raw data is extracted from archived XML dumps, where posts and comments are parsed to reconstruct the full threaded discussion hierarchy, preserving the collaborative reasoning process characteristic of community-driven knowledge building. Rule-based cleaning and format normalization are applied to ensure consistent structure, placing the dataset at Darwin Level L2 and contributing approximately 20B tokens.

\subsubsection{Science}

Scientific knowledge forms an important part of a well-rounded pretraining corpus, enriching the model with structured, knowledge-dense content spanning a broad range of disciplines. Our scientific corpus is assembled from a combination of existing open-source datasets and documents collected directly from public repositories, with additional processing applied to improve quality where appropriate.

\paragraph{MegaMath.}
We incorporate three subsets from \textit{MegaMath}~\cite{zhou2025megamath} as part of our mathematical pretraining corpus. MegaMath is curated from diverse math-focused sources via a two-stage, coarse-to-fine web extraction pipeline over 99 Common Crawl snapshots, combined with math-related code recall from Stack-V2 and LLM-based synthetic data generation.

\begin{itemize}
    \item \textit{MegaMath-Web.}
    This subset consists of mathematical content 
    extracted from Common Crawl using a two-stage 
    pipeline: an initial fast extraction pass with 
    Resiliparse followed by high-fidelity 
    re-extraction with trafilatura on 
    math-optimized HTML, with fastText-based math 
    filtering and MinHash LSH deduplication applied 
    throughout. The result is filtered through a 
    fastText classifier trained on LLM-annotated 
    math-relevance labels, reaching Darwin Level L3 
    and contributing approximately 231B tokens.

    \item \textit{MegaMath-Web-Pro.}
    This subset is a high-quality subset of 
    MegaMath-Web produced by first applying the FineMath 
    classifier with a 
    dynamic educational-value threshold (score 
    $\geq 4$ for older snapshots; score $\geq 3$ 
    for recent ones) to select high-quality 
    documents, followed by LLM-driven refinement 
    using Llama-3.3-70B-Instruct to remove noise 
    and reorganize content into a logically 
    structured, information-dense form while 
    preserving the original length. This reaches 
    Darwin Level L4 and contributes approximately 
    13B tokens.

    \item \textit{MegaMath-Synth-Code.}
    This subset consists of LLM-generated natural 
    language--code interleaved pairs targeting 
    mathematical reasoning, produced by prompting 
    LLMs to generate structured blocks of 
    mathematical text, symbolic expressions, and 
    executable Python code grounded in MegaMath-Web 
    documents, with syntax and runtime verification 
    via AST filtering and execution checks. It 
    reaches Darwin Level L5 and contributes 
    approximately 5B tokens.
\end{itemize}

\paragraph{MegaMath Refined.}
We apply L4 Generative Refinement to 
\textit{MegaMath-Web} using Qwen3-235B-A22B-Instruct~\cite{qwen3}, following 
the same prompt design as MegaMath-Web-Pro. The 
refinement strategy prompts the model to extract 
key facts and concepts, remove noisy or irrelevant 
content, and reorganize the material into a 
logically structured, information-dense form while 
preserving the original length (see Appendix~\ref{dataset construction prompt} for the full prompt). The refined version of MegaMath-Web reaches Darwin Level L4 and contributes approximately 176B tokens.

\paragraph{Nemotron-CC-Math-v1.}
We incorporate three subsets from 
\textit{Nemotron-CC-Math-v1}~\cite{mahabadi2025nemotron}, 
a high-quality mathematical corpus constructed from 98 
Common Crawl snapshots (2014--2024) spanning over 980,000 
unique domains. Its pipeline identifies math-relevant pages 
by aggregating URL lists from existing open math datasets, 
renders HTML via the Lynx text-based browser, applies a 
Phi-4-based cleanup pass to remove boilerplate and normalize 
heterogeneous mathematical representations into unified 
\LaTeX{} format, and then uses the FineMath 
classifierto score each document 
on a 1--5 scale for quality filtering, followed by fuzzy 
deduplication and benchmark decontamination.

\begin{itemize}
    \item \textit{Nemotron-CC-Math-v1-3.}
This subset retains documents with FineMath classifier 
scores of 3. The Phi-4-based cleanup pass applied during 
the original pipeline constitutes an L4 Generative 
Refinement operation, bringing this subset to Darwin 
Level L4 and contributing approximately 81B tokens.

    \item \textit{Nemotron-CC-Math-v1-4+.}
This subset retains only the highest-quality documents 
with FineMath classifier scores of 4--5. As with 
Nemotron-CC-Math-v1-3, the Phi-4-based cleanup pass 
places this subset at Darwin Level L4, contributing 
approximately 52B tokens.

    \item \textit{Nemotron-CC-Math-v1-4+-MIND.}
    This subset is obtained by applying the MIND 
    framework~\cite{akter2025mind} to 
    \textit{Nemotron-CC-Math-v1-4+}, converting each 
    mathematical document into structured multi-turn dialogues 
    via diverse conversational prompt templates (e.g., 
    Teacher-Student, Problem-Solving, Debate) that reconstruct implicit reasoning steps and lower the cognitive barrier for model internalization, while strictly preserving the original content without introducing external knowledge. It reaches Darwin Level L5 and contributes approximately 74B tokens.
    
\end{itemize}

\paragraph{Nemotron-CC-Math-v1 Refined.}
Building on the L4 Generative Refinement already applied 
during the original dataset construction, we apply a second 
round of L4 refinement to \textit{Nemotron-CC-Math-v1-3} 
and \textit{Nemotron-CC-Math-v1-4+} using the stronger 
Qwen3-235B-A22B-Instruct~\cite{qwen3}, with a more targeted 
prompt that instructs the model to extract key facts and 
concepts, remove noisy or irrelevant content, and reorganize 
the material into a logically structured, information-dense 
form while preserving the original length (see 
Appendix~\ref{dataset construction prompt}  for the full prompt). 
This reflects our broader treatment of Darwin Level operations 
as iterative rather than one-time passes.

\begin{itemize}
    \item \textit{Nemotron-CC-Math-v1-3 Refined.}
    The refined version of Nemotron-CC-Math-v1-3, reaching 
    Darwin Level L4 and contributing approximately 68B tokens.

    \item \textit{Nemotron-CC-Math-v1-4+ Refined.}
    The refined version of Nemotron-CC-Math-v1-4+, reaching 
    Darwin Level L4 and contributing approximately 47B tokens.
\end{itemize}

% \subsubsection{Science}

\paragraph{Darwin-Science.}
We construct \textit{Darwin-Science} as our primary scientific corpus, built from raw PDFs of scientific books and academic papers sourced from publicly accessible online repositories and open-source datasets including PubMed and arXiv. Both sources share a common L0--L3 processing pipeline. Raw PDFs are first converted into machine-readable text using olmOCR-7B-0225-preview~\citep{olmocr2}, a vision-language model optimized for document text extraction. The resulting text then undergoes deduplication via MinHash LSH, followed by rule-based filtering that discards files below 8kb, documents with excessive garbled characters resulting from OCR errors, and non-English content. All retained documents are subsequently annotated using EAI-Distill-0.5B~\citep{ai2025essentialwebv1024ttokens}, a lightweight classifier that performs educational value scoring and field-of-discipline classification across nine major domains; documents with no educational value are filtered out. Finally, all documents are classified into book and paper categories---using metadata where available, and Qwen2.5-7B-Instruct~\citep{qwen2.5} for ambiguous cases---as the two document types exhibit distinct learnability characteristics that warrant different downstream processing.
\begin{itemize}
\item \textit{Darwin-Science-Book}.
Scientific books are processed through L4 Generative Refinement using GPT-OSS-120B~\citep{openai2025gptoss120bgptoss20bmodel}, which removes structural noise such as table of contents entries, reference lists, headers and footers, and OCR artifacts, while repairing formatting defects such as fragmented text and damaged formulas without altering the underlying content. The detailed prompt is provided in Appendix~\ref{dataset construction prompt}. This process yields approximately 251 tokens.

\item \textit{Darwin-Science-Paper}.
The same L4 Generative Refinement pipeline is applied to academic papers using GPT-OSS-120B, yielding approximately 215 tokens. These L4-processed papers serve as the foundation for further L5 processing.

\item \textit{Darwin-Science-Paper-GPT}.
Building on the L4-processed papers, we apply L5 Cognitive Completion 
using GPT-OSS-120B to bridge the learnability 
gap inherent in expert-oriented scientific writing. The augmentation 
targets three dimensions: expanding implicit logical leaps into explicit 
step-by-step derivations (Reasoning Reconstruction), contextualizing 
domain-specific terminology within the narrative flow rather than assuming prior mastery (Terminological Explication), and grounding abstract concepts in concrete analogies and established knowledge (Pedagogical Bridging). The detailed prompt is provided in Appendix~\ref{dataset construction prompt}. This yields approximately 290B tokens.

\item \textit{Darwin-Science-Paper-Qwen}.
We also apply L5 Cognitive Completion directly to the L4-processed papers using Qwen3-235B-A22B-Instruct~\citep{qwen3}. We hypothesize that a stronger model is better equipped to reconstruct implicit reasoning and produce richer pedagogical enrichment, potentially yielding greater learnability gains. This yields approximately 440 tokens.

\end{itemize}
\subsubsection{QA}
Question-answer pairs serve as a natural complement to raw text pretraining data, and our QA data covers three broad domains: code, general, and science. Sources are drawn from a mix of existing high-quality collections, rejection-sampled open-source post-training datasets by Qwen3-32B in non-thinking mode, and QA pairs synthesized directly from scientifc documents. All subsets reach Darwin Level L5.

% Required packages:
% \usepackage{booktabs}
% \usepackage{array}
% \usepackage[table]{xcolor}
% \usepackage{tikz}

% Colour definitions — put these in your preamble:
% \definecolor{catblue}{HTML}{DDEEFF}
% \definecolor{rowblue}{HTML}{F2F8FF}
% \definecolor{markblue}{HTML}{7B9CD9}  % colour for the filled circle marker

% Helper command — put this in your preamble:
% \newcommand{\cmark}{\tikz\fill[markblue] circle (2.2pt);}

% Required packages:
% \usepackage{booktabs}
% \usepackage{array}
% \usepackage[table]{xcolor}
% \usepackage{tikz}

% Colour definitions — put these in your preamble:
% \definecolor{catpurple}{HTML}{E8E0F5}    % category/header shading
% \definecolor{rowpurple}{HTML}{F4F0FB}    % alternating row
% \definecolor{markpurple}{HTML}{8B6DB5}   % filled circle marker

% Helper command — put this in your preamble:
% \newcommand{\cmark}{\tikz\fill[markpurple] circle (2.2pt);}

\begin{table}[htbp]
\centering
\caption{Domain coverage of QA data sources. Each source is categorized
into one of three domains: General, Code, and Science
(including Mathematics).}
\label{tab:qa-domains}
\begin{tabular}{
  l
  >{\centering\arraybackslash}p{1.6cm}
  >{\centering\arraybackslash}p{1.6cm}
  >{\centering\arraybackslash}p{1.6cm}
}
\toprule
\textbf{Data Source} & \textbf{General} & \textbf{Code} & \textbf{Science} \\
\midrule
\rowcolor{rowpurple}
\quad Nemotron-CC-v1 Synthetic QA (492B)             & \cmark &        &        \\
\rowcolor{white}
\quad Nemotron-Pretraining-SFT-v1-General (12B)     & \cmark &        &        \\
\rowcolor{rowpurple}
\quad Nemotron-Pretraining-SFT-v1-Code (21B)         &        & \cmark &        \\
\rowcolor{white}
\quad Llama-Nemotron-Post-Training-Code (5B)        &        & \cmark &        \\
\rowcolor{rowpurple}
\quad Nemotron-Pretraining-SFT-v1-Math (138B)        &        &        & \cmark \\
\rowcolor{white}
\quad Llama-Nemotron-Post-Training-Math-RS (10B)    &        &        & \cmark \\
\rowcolor{rowpurple}
\quad Llama-Nemotron-Post-Training-Science-RS (0.4B)  &        &        & \cmark \\
\rowcolor{white}
\quad MegaMath-QA-RS (9B)                          &        &        & \cmark \\
\rowcolor{rowpurple}
\quad MegaScience-RS (1B)                          &        &        & \cmark \\
\rowcolor{white}
\quad Darwin-Science-Book QA (46B)                  &        &        & \cmark \\
\bottomrule
\end{tabular}
\end{table}

\paragraph{Nemotron-Pretraining-SFT-v1.}
We incorporate \textit{Nemotron-Pretraining-SFT-v1}~\cite{nvidia2025nvidianemotronnano2}as a source of short-form supervised fine-tuning style pretraining data covering code, math, and general knowledge domains. For each subset, we retain only the short chain-of-thought portion of the data, discarding verbose reasoning traces. All subsets reach Darwin Level L5.

\begin{itemize}
    \item \textit{Nemotron-Pretraining-SFT-v1-Code.}
    This subset is synthesized using the Genetic-Instruct
    framework~\cite{majumdar2025geneticinstructscalingsynthetic}, an evolutionary algorithm that begins with a small set of seed coding instructions and iteratively generates diverse, challenging instruction-code pairs through two operations: crossover, which prompts an LLM to produce new instructions from a set of seed examples, and mutation, which evolves an existing instruction into a harder or more varied one. A separate Coder-LLM generates corresponding code solutions, and a Judge-LLM filters outputs based on correctness and quality. This process yields a large-scale collection of coding problem-solution pairs spanning multiple programming languages, reaching Darwin Level L5 and contributing approximately 21B tokens.

    \item \textit{Nemotron-Pretraining-SFT-v1-Math.}
    This subset is synthesized following 
    OpenMathInstruct-2~\cite{toshniwal2024openmathinstruct2acceleratingaimath}, which uses a strong teacher model to generate multiple candidate solutions for a diverse set of seed math questions drawn from existing benchmarks such as GSM8K and MATH. High-quality solutions are selected based on answer correctness, and question diversity is explicitly optimized to maximize coverage across mathematical topics and difficulty levels. This reaches Darwin Level L5 and contributes approximately 138B tokens.

    \item \textit{Nemotron-Pretraining-SFT-v1-General.}
    This subset consists of MMLU-style question-answer pairs covering a broad range of knowledge topics across different domains and difficulty levels, synthesized by prompting an LLM to generate both questions and answers grounded in curated source documents. This reaches Darwin Level L5 and contributes approximately 12B tokens.
\end{itemize}

\paragraph{Llama-Nemotron-Post-Training-Dataset.}
We incorporate multiple subsets of \textit{Llama-Nemotron-Post-Training-Dataset}~\citep{bercovich2025llamanemotronefficientreasoningmodels}, a large-scale post-training dataset in which prompts are sourced from public corpora or synthetically generated, and responses are synthesized by a range of open-source models. We use the code subset directly without additional processing, while for the science and math subsets we use the original prompts as input and apply rejection sampling using Qwen3-32B~\cite{qwen3} in non-thinking mode, retaining only responses that pass correctness verification against ground-truth answers.

\begin{itemize}
    \item \textit{Llama-Nemotron-Post-Training-Dataset-Code.}
    % ↑ 删掉了开头重复提数据集名字和"directly without filtering"的句子
    This subset targets diverse programming tasks and problem-solving scenarios across multiple languages, with prompts sourced from public corpora or synthetically generated, and responses filtered for quality and correctness during the original dataset construction. It reaches Darwin Level L5 and contributes approximately 5B tokens.

    \item \textit{Llama-Nemotron-Post-Training-Dataset-Science-RS.}
    % ↑ 删掉了"We apply rejection sampling to the science subset of \textit{Llama-Nemotron-...}"
    This subset is obtained by applying rejection sampling to the science subset. The science subset comprises open-ended and multiple-choice questions spanning academic scientific domains including physics, biology, and chemistry. The questions are drawn from two sources: question-answer pairs extracted from StackOverflow, and synthetically generated MCQs conditioned on topic, subtopic, and difficulty level using Qwen2.5 models following the OpenMathInstruct-2 augmentation pipeline. It reaches Darwin Level L5 and contributes approximately 0.4B tokens.

    \item \textit{Llama-Nemotron-Post-Training-Dataset-Math-RS.}
    % ↑ 同上，删掉了重复的开头
 This subset is obtained by applying rejection sampling to the science subset. The subset comprises mathematical reasoning problems spanning competition-level and general math domains, with responses synthesized by frontier open-source models. It reaches Darwin Level L5 and contributes approximately 10B tokens.
\end{itemize}

\paragraph{MegaMath-QA-RS.}
We apply rejection sampling to \textit{MegaMath-QA}~\cite{zhou2025megamath},
the synthetic QA subset of MegaMath, in which question-answer pairs are
extracted from mathematical web documents and refined to make intermediate
reasoning steps explicit using an ELI5-style prompting strategy. RWe apply rejection sampling using Qwen3-32B~\cite{qwen3} in non-thinking mode, retaining only responses whose final answers are verifiably correct against ground-truth answers, yielding approximately 9B tokens at Darwin Level L5.

\paragraph{MegaScience-RS.}
We apply rejection sampling to \textit{MegaScience}~\cite{fan2025megasciencepushingfrontiersposttraining}, a post-training dataset
covering scientific reasoning across multiple STEM disciplines, with
questions drawn from textbooks and curated scientific sources. We apply rejection sampling using Qwen3-32B~\cite{qwen3} in non-thinking mode, retaining responses that meet correctness criteria, yielding approximately 1B tokens at Darwin Level L5.

\paragraph{Darwin-Science-Book QA.}
We generate knowledge-grounded QA pairs directly from \textit{Darwin-Science Book} using 
Qwen3-235B-A22B-Instruct~\cite{qwen3}. To account for the distinct knowledge structures and 
expository styles across scientific disciplines, we design domain-specific prompts for each of the book domains covered in Darwin-Science, rather than applying a single universal prompt. The model is prompted to identify key knowledge points from each source passage and formulate question-answer pairs that are strictly grounded in the original text, with the constraint that every answer must be directly verifiable against its source passage. To further enhance learnability, the model is also prompted to supply intermediate reasoning steps that bridge the question to the answer, making implicit derivations explicit. The full set of domain-specific prompts is provided in The detailed prompt is provided in Appendix~\ref{dataset construction prompt}.. This dataset reaches Darwin Level L5, contributing approximately 46B tokens.

\paragraph{Nemotron-CC-v1 Synthetic QA.}
We incorporate the Diverse QA subset of
\textit{Nemotron-CC-v1}, which is generated from
high-quality documents selected from Common Crawl. In this pipeline, an LLM
is prompted to generate question-answer pairs in multiple forms, including
yes/no questions, open-ended questions, and multiple-choice questions,that
probe factual information in the source text at different cognitive levels.
The model is required to provide clear and concise answers while preserving
concrete details such as numbers and specific facts from the original document.
This grounding in high-quality web text spanning diverse general-domain topics
makes the subset a broad-coverage complement to the domain-specific QA sources
described above. This dataset reaches Darwin Level L5 and contributes
approximately 492B tokens.

\section{Training Recipe: How We Trained It}
\label{sec:training-recipe}
In this section, we provide a comprehensive disclosure of the pretraining process for daVinci-LLM, focusing on the technical execution and methodological transparency that define our approach. Moving beyond the industry standard of releasing only final checkpoints, we adopt a fully-open paradigm that documents the model's complete developmental trajectory from random initialization, ensuring that every design choice is traceable and reproducible. The following subsections detail our evaluation protocols, the multi-stage adaptive curriculum, and the systematic data orchestration strategies that shape the model’s capabilities.

This documentation bridges the gap between opaque commercial technical reports and scale-limited academic papers, providing a transparent framework for the community to replicate our results and adapt the underlying technical workflows of pretraining.

\subsection{Model Architecture}
\label{sec:arch}

We adopt the Qwen2-based~\cite{qwen2} transformer architecture for \textsc{daVinci-LLM}, training a 3B-parameter model from random initialization using a standard decoder-only causal language modeling objective. The architecture employs several modern design choices that balance computational efficiency with representational capacity: Grouped-Query Attention (GQA)~\cite{ainslie2023gqa} with 2 key-value heads shared across 16 query heads to reduce memory bandwidth while preserving attention expressiveness; SwiGLU~\cite{shazeer2020glu} activation in the MLP layers with an expansion ratio of $\sim$5.4$\times$ (11008/2048) for enhanced nonlinearity; RMSNorm~\cite{zhang2019root} for efficient pre-normalization; and Rotary Position Embeddings (RoPE)~\cite{su2024roformer} with base frequency 10000 for length generalization. We train with a sequence length of 4096 tokens. The model utilizes a vocabulary of 151936 tokens from the Qwen2 tokenizer.

Table~\ref{tab:arch} provides the complete architectural specification. The configuration follows the Qwen2 design philosophy of prioritizing depth (36 layers) with moderate hidden dimensions (2048), a strategy that has proven effective for balancing parameter count, training throughput, and downstream task performance in the 3B scale regime.

\begin{table}[t]
  \centering
  \caption{Architectural specification of \textsc{daVinci-LLM}.}
  \label{tab:arch}
  \setlength{\tabcolsep}{16pt} % 调整列间距，默认值是6pt
  \begin{tabular}{l c l c} % 第1、3列左对齐，第2、4列居中对齐
    \toprule
    \rowcolor{blue!7} Parameters & $\sim$3.09B & MLP intermediate size & 11008 \\
    Layers) & 36 & Activation function & SwiGLU \\
    \rowcolor{blue!7} Hidden size & 2048 & Positional encoding & RoPE (base $\theta = 10000$) \\
    Attention heads & 16 & Max position embeddings & 4096 \\
    \rowcolor{blue!7} KV heads (GQA) & 2 & Normalization & RMSNorm ($\epsilon = 10^{-6}$) \\
    Head dimension & 128 & Tokenizer & Qwen2 tokenizer \\
    \rowcolor{blue!7} Vocab size & 151936 & Precision & bfloat16 \\
    \bottomrule
  \end{tabular}
\end{table}

\subsection{Evaluation Protocol}
\label{sec:eval}

\paragraph{Evaluation philosophy.} 
% \pfliu{add this paragraph, please check} \huangz{fixed}
Traditional pretraining evaluations focus solely on final performance. We adopt a more 
comprehensive approach aligned with our question-driven methodology: (1) Tracking 
capability emergence: We evaluate at 5k-step intervals to understand when different 
dimensions saturate. (2) Benchmark stability analysis: We systematically 
investigate which benchmarks remain stable indicators versus which collapse under certain 
training configurations (Section~\ref{sec:benchmark_stability}). (3) Multi-domain 
coverage: Our 19 benchmarks span general knowledge, code, and science to capture synergy 
effects and trade-offs.

We conduct comprehensive evaluation of our pretrained checkpoints across diverse benchmarks to assess general knowledge, reasoning, code generation, and mathematical problem solving capabilities. This section describes our evaluation protocol, benchmark selection, and baseline comparisons.

\paragraph{Benchmarks and metrics.}
% \yiwei{hz: Repeat the three domains for several times.}
% \yiwei{hz: Should we put how we evaluate(ppl or generative) here instead of in the appendix? It's a key decision. Also briefly mention the rationale behind why ppl or generative, and point to the evaluation section.}
% We evaluate on 19 tasks spanning three capability domains: General, Code, and Science. We report task-level metrics (accuracy for multiple-choice tasks, exact match for open-ended QA, pass@1 for code generation) and compute cluster averages for each domain. An overall average across all 19 tasks provides a single-number summary of model capability.

We evaluate on 19 tasks spanning three capability domains: General, Code, and Science. The General cluster includes MMLU~\cite{hendryckstest2021}, MMLU-Pro~\cite{wang2024mmlu}, AGIEval~\cite{zhong2023agieval}, HellaSwag~\cite{zellers2019hellaswag}, TriviaQA~\cite{joshi2017triviaqa}, RACE~\cite{zheng2024race}, WinoGrande~\cite{sakaguchi2019winogrande}, OpenBookQA~\cite{OpenBookQA2018}, and PIQA~\cite{bisk2020piqa}. The Code cluster evaluates Python code synthesis through HumanEval~\cite{chen2021codex}, EvalPlus~\cite{evalplus}, and MBPP~\cite{austin2021program}. The Science cluster includes GSM8K~\cite{cobbe2021gsm8k}, GSM-Plus~\cite{li2024gsm}, MATH~\cite{hendrycks2021measuring}, GPQA~\cite{rein2024gpqa}, SuperGPQA~\cite{du2025supergpqa}, MMLU-STEM, and MMLU-Pro-STEM. Detailed descriptions of all benchmarks are provided in Appendix~\ref{appendix:benchmarks}. We evaluate using the \texttt{lm-eval-harness} framework (EleutherAI)~\cite{eval-harness}.

Since the evaluated models are base checkpoints, we perform inference under greedy decoding to ensure consistency across experiments. We adopt two complementary evaluation strategies based on task characteristics. \textit{Perplexity-based} (PPL) evaluation is applied to tasks focused on multiple-choice selection or likelihood estimation, where the model scores each candidate answer directly: PIQA (0-shot), MMLU (5-shot), OpenBookQA (5-shot), GPQA-Main (5-shot), and MMLU-STEM (5-shot). \textit{Generative-based} evaluation is applied to tasks requiring complex reasoning chains, code generation, or Chain-of-Thought (CoT) outputs, where the model generates a free-form response: MATH (4-shot), MMLU-Pro (5-shot), SuperGPQA (5-shot), MMLU-Pro-STEM (5-shot), GSM8K (8-shot), HumanEval (0-shot), EvalPlus (0-shot), MBPP (3-shot), AGIEval (0-shot), HellaSwag (0-shot), TriviaQA (5-shot), RACE (0-shot), WinoGrande (0-shot), and GSM-Plus (5-shot).

\paragraph{Baselines.}
We compare against six open base models spanning similar parameter scales:
\begin{itemize}[leftmargin=*,noitemsep]
  \item OLMo-3 7B (7B)~\cite{olmo2025olmo}: Our primary reference baseline, representative model trained by academic institutions.
  \item OLMo-2 7B (7B)~\cite{olmo20242}: Previous generation OLMo model.
  \item Qwen-3.5-4B (4B) ~\cite{qwen3.5}: The latest model baseline from Alibaba's Qwen family.
  \item Qwen-3-4B (4B)~\cite{qwen3}: Strong baseline from Alibaba's Qwen family.
  \item Qwen-2.5-3B (3B)~\cite{qwen2.5}: Closest parameter-matched baseline (same architecture family as our model).
  \item LLaMa-3.2-3B (3B)~\cite{dubey2024llama}: Meta's LLaMa 3.2 series at matched scale.
  \item Yulan-Mini-2.4B (2.4B)~\cite{hu2024yulan}: Another model led by academic institutions.
\end{itemize}

These baselines provide both scale-matched comparisons (3--4B parameters) and capability-matched comparisons (7B models with stronger absolute performance). Our goal is to demonstrate that principled multi-stage pretraining can enable a 3B model to approach or exceed the performance of larger baselines through careful data mixture design and training stage orchestration.

\begin{table}[!htp]\centering
\caption{Training hyperparameters of different training stages. Note that the Global Batch Size refers to the total number of sequences per training step; the total number of tokens per step is calculated as $\text{Global Batch Size} \times \text{Sequence Length}$.}\label{tab: stage1_hyper_parameter}
\resizebox{\textwidth}{!}{%
\begin{tabular}{lcccc}\toprule
& \multicolumn{2}{c}{\textbf{Stage 1}} & \multicolumn{2}{c}{\textbf{Stage 2}} \\
\cmidrule(lr){2-3}\cmidrule(lr){4-5}
&Stage~1-1 &Stage~1-2 &Stage~2-1 &Stage~2-2 \\\midrule
Training Tokens &4 Trillion &2 Trillion &1 Trillion &1 Trillion \\
Global Batch Size &1024$\rightarrow$2048$\rightarrow$4096 &4096 &4096 &4096 \\
Rotatory Base &10000 &10000 &10000 &10000 \\
Sequence Length &4096 &4096 &4096 &4096 \\
Precision &bfloat16 &bfloat16 &bfloat16 &bfloat16 \\
Weight Decay &0.1 &0.1 &0.1 &0.1 \\
AdamW $\beta_1$ &0.9 &0.9 &0.9 &0.9 \\
AdamW $\beta_2$ &0.95 &0.95 &0.95 &0.95 \\\midrule
LR Strategy & \makecell{2000 warmup\\+ constant (3e-4)} & \makecell{cosine decay\\(3e-4$\rightarrow$3e-5)} & \makecell{2000 warmup\\+ constant (3e-5)} & \makecell{constant (3e-5)} \\
\bottomrule
\end{tabular}%
}
\end{table}

\subsection{Training Methodology}
\label{sec:training}
Building upon the architectural specifications and evaluation framework 
previously defined, we present the systematic pretraining methodology 
for \textsc{daVinci-LLM}. Moving beyond the view of pretraining as a 
monolithic sequence of token consumption, we adopt a \textbf{multi-stage, adaptive curriculum} that evolves in tandem with the model's maturing capabilities. Our training process is structured into two primary phases, shifting from expansive knowledge acquisition to reasoning-intensive enhancement:

\begin{itemize}[leftmargin=*,noitemsep]
    \item \textbf{Stage~1 (General Foundation Pretraining)}: Establishes 
    broad foundational capabilities through 6T tokens of diverse, web-scale corpora. This phase is executed across two substages (Stage~1-1/1-2), utilizing progressive data adjustment to calibrate the model's exposure to web text, code, and scientific content.

    \item \textbf{Stage~2 (Reasoning Capability Enhancement)}: Shifts 
    the distribution toward reasoning-dense data through an additional 
    2T tokens. By integrating structured QA, refined scientific content, and high-quality code, this stage amplifies reasoning capabilities, enabling our 3B-parameter model to match the 7B-scale OLMo-3.
\end{itemize}

The execution of these stages is not guided by heuristic conventions, 
but is principled and evidence-based, rooted in our extensive ablation 
studies (Section~\ref{sec:exploration}). Specifically, our training 
recipe is informed by three scientific pillars: (1) \textbf{Data 
Processing Depth} (Section~\ref{sec:data-processing}), utilizing the 
Data Darwinism taxonomy to systematically enhance data quality; 
(2) \textbf{Adaptive Training Dynamics} (Section~\ref{sec:training_dynamics}), 
where stage transitions are informed by differential saturation rates 
across cognitive dimensions (general knowledge plateaus early, while 
code and science sustain growth); and (3) \textbf{Mixture Optimization} 
(Section~\ref{sec:mixture-design}), balancing concentration for targeted 
enhancement with diversity for capability preservation.

% \paragraph{Training Infrastructure.}
% \yiwei{TODO: should we add this?}
% We conduct all training on ** NVIDIA ** **GB GPUs (** nodes, 8 GPUs 
% per node) using the NeMo framework~\cite{nemo}. The parallelism strategy employs data parallelism degree **, tensor parallelism degree 1, and no pipeline parallelism (DP=**, TP=1, PP=1) to maximize throughput while maintaining training stability. Stage~1 training (6T tokens) completes in approximately ** days, while Stage~2 (2T tokens) requires ** days on this configuration. Our training achieves **\% Model FLOPs Utilization (MFU), indicating efficient hardware utilization throughout both stages.

\subsubsection{Stage~1: General Foundation Pretraining}
\begin{figure}[t]
  \centering
  \includegraphics[width=0.999\linewidth]{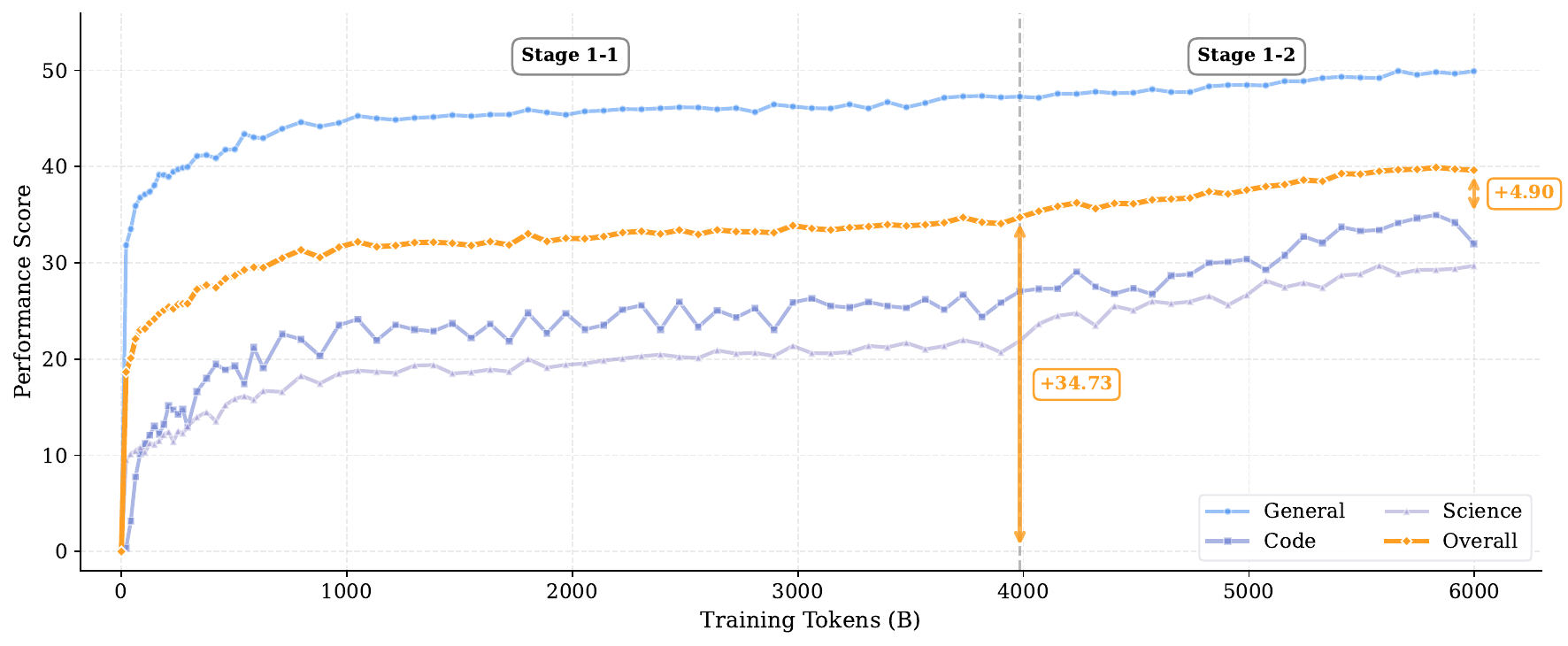}
  \caption{Progressive training results across Stage~1-1 and Stage~1-2, with checkpoints evaluated every 5000 steps. The vertical dashed line indicate the boundary between two substages.}
  \label{fig:stage1-progressive-training}
\end{figure}

\label{sec:stage1}
Stage~1 trains the model from random initialization across 6T tokens 
to establish broad foundational capabilities in natural language 
understanding, logical reasoning, and cross-domain knowledge synthesis.

\paragraph{Multi-substage Training Recipe.}
We decompose Stage 1 into two consecutive sub-phases: Stage 1-1 and Stage 1-2, consuming approximately 6 trillion tokens in total. This phased approach enables efficient scaling of batch size and fine-grained control over capability development through data mixture adjustments. The detailed hyperparameters for each substage are provided in Table~\ref{tab: stage1_hyper_parameter} and data mixtures are provided in Table~\ref{tab:data-pools}.

\begin{itemize}
    \item \textbf{Stage~1-1 (Foundation Building, 4T tokens)}: This stage prioritizes stability. We employ a progressive global batch size (GBS) scaling strategy, starting at 1,024 for 70k steps, increasing to 2,048 for 40k steps, and finally reaching 4,096. The learning rate is held constant at 3e-4 after a 2,000-step linear warmup. The data mixture is dominated by Common Crawl (68.2\%) to establish broad linguistic fluency.

    \item \textbf{Stage~1-2 (Reasoning Enhancement, 2T tokens)}: 
    Maintains GBS at 4,096 while transitioning to cosine learning rate 
    decay (3e-4$\rightarrow$3e-5). To strengthen reasoning capabilities, 
    we rebalance the mixture by reducing Common Crawl to 55.4\% and 
    increasing Code (+2.1\%) and Science (+10.6\%) 
    proportions, introducing denser logical and symbolic structures.
\end{itemize}

\paragraph{Training Stability and Convergence.} 
\begin{figure}[htbp]
    \centering
    
    \begin{subfigure}[b]{0.49\textwidth}
        \centering
        \includegraphics[width=\textwidth]{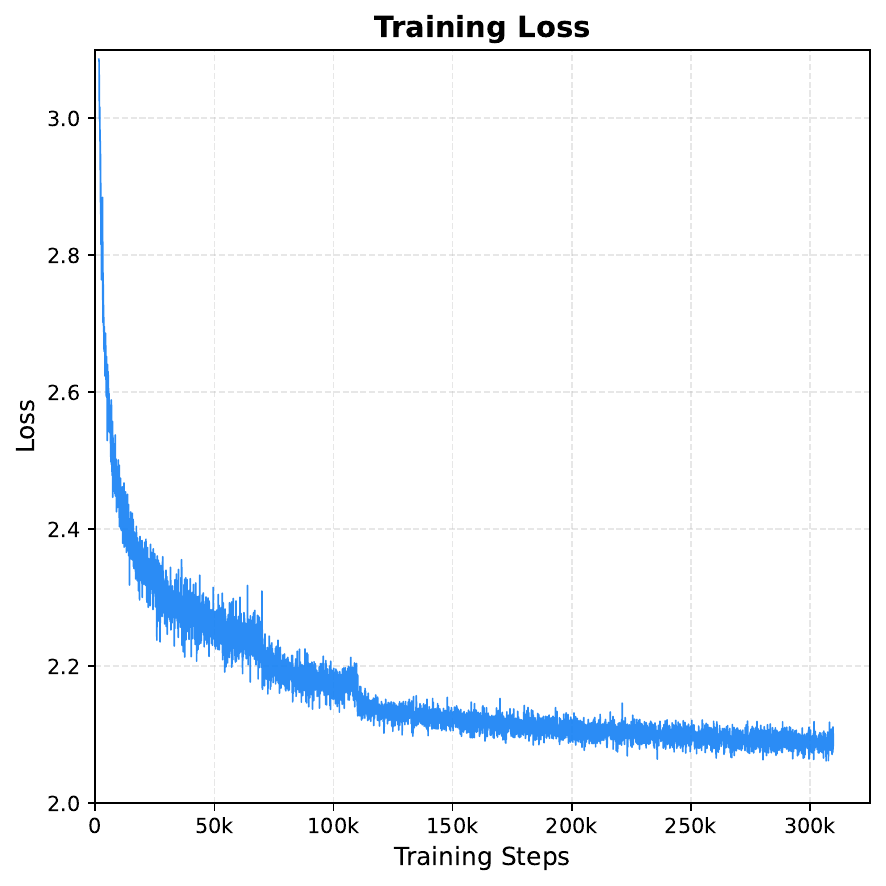}
        \caption{}
        \label{fig:stage1_loss}
    \end{subfigure}
    \hfill
    \begin{subfigure}[b]{0.49\textwidth}
        \centering
        \includegraphics[width=\textwidth]
        {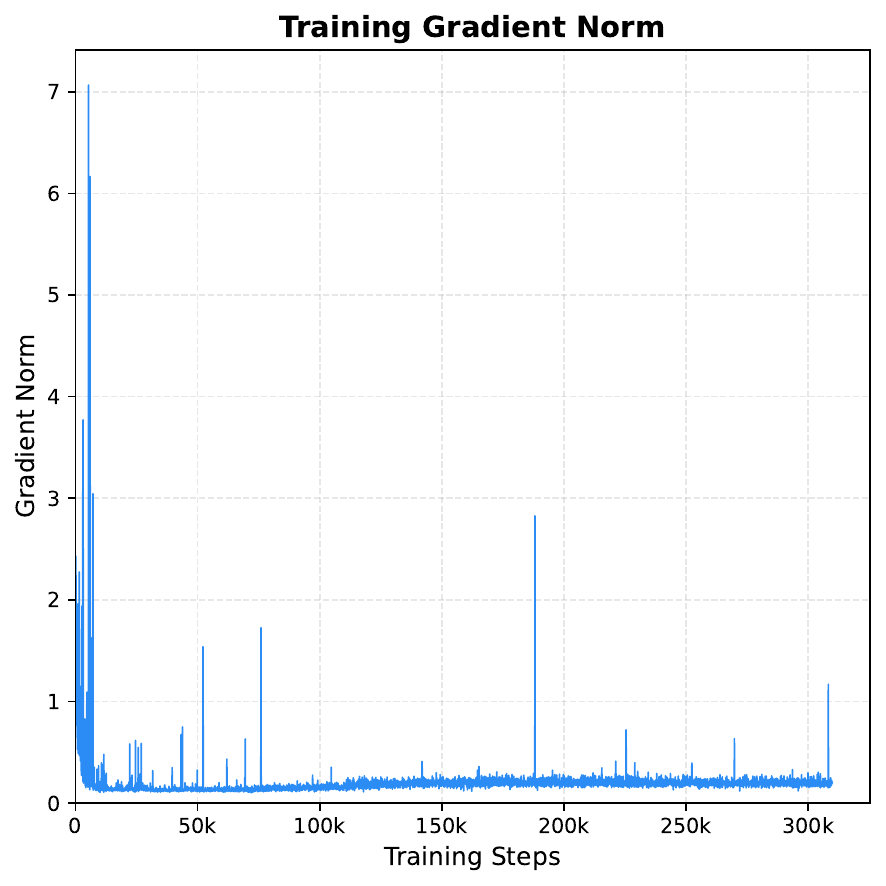}
        \caption{}
        \label{fig:stage1_grad_norm}
    \end{subfigure}
    
    \caption{Stage 1 Training dynamics for the first 300k steps: (a) training loss curve demonstrating consistent convergence, and (b) gradient norm curve tracking optimization stability across the initial training phase.}
\end{figure}

Figure~\ref{fig:stage1-progressive-training} illustrates the training dynamics for a representative portion of Stage 1, highlighting the overall optimization stability. The training loss (Figure~\ref{fig:stage1_loss}) demonstrates smooth and consistent convergence, reflecting predictable responses to adjustments in batch size and learning rate schedules. Notably, the gradient norm (Figure~\ref{fig:stage1_grad_norm}) remained stable throughout the illustrated steps, which is consistent with the behavior observed throughout the entire pretraining process. No significant gradient spikes or loss divergences were encountered during the run, enabling continuous training without the need for manual interventions or restarts.

\paragraph{Training Trajectories.}
Figure~\ref{fig:stage1-progressive-training} tracks capability evolution across 
Stage~1's 6T-token training trajectory, with checkpoints evaluated 
every 5k steps across all 19 benchmarks. General knowledge benchmarks 
plateau rapidly within the first 1T tokens, while code and science 
benchmarks sustain consistent growth throughout training, with notable 
acceleration during Stage~1-2 following the reasoning-heavy mixture 
adjustment. This differential saturation pattern,where general capabilities stabilize early while reasoning capabilities require extended training,motivated our multi-substage design with progressive data adjustment. Stage~1 concludes with an overall average of 39.58, establishing a solid foundation for Stage~2's reasoning enhancement.

\subsubsection{Stage-2: Reasoning Capability Enhancement}
\begin{figure}[t]
  \centering
  \includegraphics[width=0.999\linewidth]{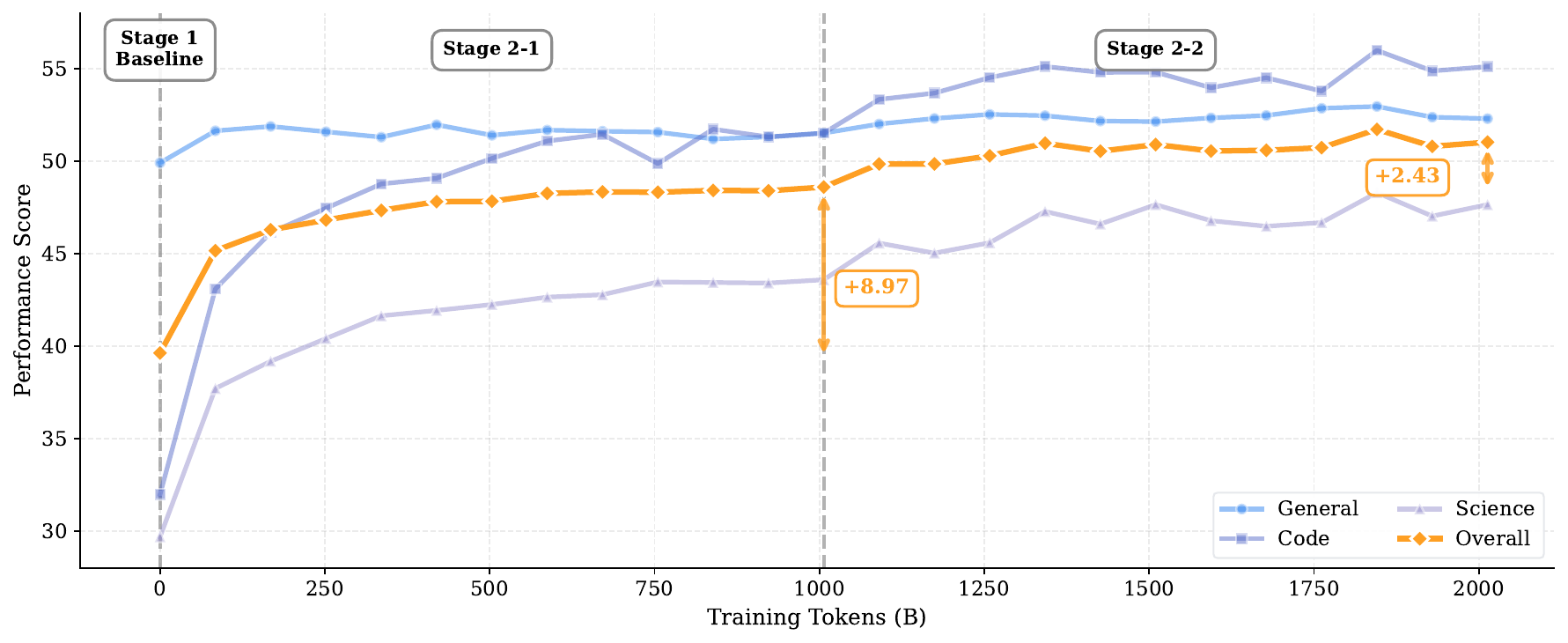}
  \caption{Progressive training results across Stage~2-1 and Stage~2-2, with checkpoints evaluated every 5000 steps. The vertical dashed line indicate the boundary between two substages.}
  \label{fig:progressive-training}
\end{figure}

Building upon the solid general language foundation established in Stage 1, Stage 2 is dedicated to transforming the model's general capabilities into high-order reasoning proficiency. Informed by the domain proportion adjustment boundary observed in Section~\ref{sec:training_dynamics}, we recognized that merely reallocating raw text proportions was no longer sufficient to sustain continuous growth in reasoning capabilities. 
Consequently, Stage 2 shifts from broad linguistic modeling to structured logic acquisition by incorporating large-scale, high-density QA data and adopting a progressive curriculum learning strategy consisting of two distinct substages totaling 2T tokens.

\paragraph{Multi-substage Training Recipe.}
The training process is governed by a strategy of consolidating the foundation through balance before targeted intensification, with detailed hyperparameters and data mixtures across substages provided in Table~\ref{tab: stage1_hyper_parameter} and Table~\ref{tab:data-pools}.
\begin{itemize}[leftmargin=*,noitemsep]
    \item \textbf{Stage 2-1}: Balanced Foundation Building (1T tokens): 
    To prevent overfitting or domain collapse when introduced to high-intensity reasoning data, we designed a balanced mixture consisting of structured QA, code data, and L4/L5-processed scientific data (30\% each). This allocation, supplemented by 10\% high-quality web text, ensures that each reasoning domain is sufficiently represented while preserving the model's general knowledge competence.
    
    \item \textbf{Stage 2-2}: QA-Intensive Enhancement (1T tokens): 
    Building on the balanced representation established in Stage 2-1, we further increased the concentration of QA data to 70\% in the final 1T tokens. This intensification strategy aims to leverage the stable representation base formed in the previous substage to amplify logical reasoning and scientific problem-solving through high-density supervisory signals.
\end{itemize}

\paragraph{Training Trajectories.}
Stage 2 trajectories demonstrate highly efficient capability gains, particularly during the transition from general foundations to specialized reasoning. As shown in Figure~\ref{fig:progressive-training}, entering the Stage 2-1 propelled the overall average from 39.58 at the Stage 1 endpoint to 48.60 within just 1T tokens. This efficiency stems from increased processing depth under the Data Darwinism framework: by utilizing L4-level Generative Refinement and L5-level Cognitive Completion, the model internalizes structured logical chains instead of redundant narratives. Consequently, our 3B model achieved an outstanding 62.8 on the MATH benchmark, drastically exceeding its Stage 1 performance of 22.0.

These dynamics also highlight a clear stage dependency in token efficiency. Transitioning to 70\% QA-intensive training in Stage 2-2 did not trigger catastrophic forgetting of general capabilities. Instead, this deliberate data mixture adjustment enabled a second acceleration in scientific reasoning, culminating in a peak overall score of 51.72. This trajectory provides empirical evidence that \textbf{in later pretraining phases, advancing data processing depth serves as a more economical scaling mechanism than simply increasing raw data volume.}

%\yiwei{Need to decide if we should put the whole evaluation trajectory in the appendix. The final result we show here is not the actual final ckpt. Instead, we cherry pick a previous ckpt.}
%Complete evaluation trajectories across all benchmarks and intermediate checkpoints for both Stage 1 and Stage 2 are provided in Appendix~\ref{appendix:trajectory}. 
A more systematic investigation of these training dynamics, including domain-specific saturation patterns, the transition from domain adjustment to structured QA, and the resulting adaptive strategies, 
is presented in Section~\ref{sec:training_dynamics} and
Section~\ref{sec:mixture-design}.
 
\subsection{Final Results}
\label{sec:final_results}

\begin{table*}[t]
  \centering
  \small
  \renewcommand{\arraystretch}{1.2}
  \caption{Comprehensive evaluation across diverse capability benchmarks. \textsc{daVinci-LLM-3B} is compared against state-of-the-art open-weight models.}
  \label{tab:main-results}
  \setlength{\tabcolsep}{3pt}
  \begin{tabular*}{\textwidth}{@{\extracolsep{\fill}} l l
  >{\columncolor{blue!7}}c >{\columncolor{olmopink!7}}c
  c c c c c c}
    \toprule
    & & \multicolumn{4}{c}{\textit{Fully-open Models}} & \multicolumn{4}{c}{\textit{Open-weight Models}} \\
    \cmidrule(lr){3-6} \cmidrule(lr){7-10}
    Domain & Benchmark & \makecell{daVinci\\3B} & \makecell{OLMO-3\\7B} & \makecell{OLMO-2\\7B} & \makecell{Yulan\\2.4B} & \makecell{LLaMa-3.2\\3B} & \makecell{Qwen-2.5\\3B} & \makecell{Qwen-3\\4B} & \makecell{Qwen-3.5\\4B} \\
    \midrule
    \multirow{9}{*}{General}
    & MMLU & 62.53 & 66.53 & 65.93 & 50.70 & 54.91 & 65.73 & 75.35 & 72.75 \\
    & MMLU-Pro & \textbf{43.50} & 35.70 & 28.40 & 23.90 & 24.50 & 39.00 & 54.10 & 53.50 \\
    & AGIEval & 26.77 & 33.75 & 31.78 & 28.22 & 22.72 & 37.15 & 45.87 & 44.89 \\
    & HellaSwag & 71.17 & 74.15 & 80.50 & 68.56 & 73.60 & 73.60 & 73.75 & 75.29 \\
    & TriviaQA & 49.90 & 55.45 & 68.01 & 27.64 & 55.22 & 51.20 & 47.44 & 49.80 \\
    & RACE & 38.56 & 40.57 & 40.96 & 35.69 & 38.95 & 38.47 & 39.62 & 39.23 \\
    & WinoGrande & 66.77 & 69.61 & 74.59 & 66.69 & 69.22 & 68.59 & 70.17 & 71.19 \\
    & OpenBookQA & 40.20 & 41.80 & 47.60 & 43.00 & 43.00 & 43.80 & 43.20 & 46.60 \\
    & PIQA & 77.26 & 78.62 & 81.07 & 76.22 & 77.58 & 78.89 & 77.86 & 78.89 \\
    & \textit{Avg General} & \textit{52.96} & \textit{55.13} & \textit{57.65} & \textit{46.74} & \textit{51.08} & \textit{55.16} & \textit{58.60} & \textit{59.13} \\
    \midrule
    \multirow{3}{*}{Code}
    & HumanEval & \textbf{61.64} & 59.05 & 16.78 & 66.77 & 33.17 & 60.17 & 65.93 & 71.46 \\
    & EvalPlus & \textbf{57.32} & 53.62 & 13.85 & 62.25 & 27.22 & 53.23 & 59.45 & 65.00 \\
    & MBPP & 49.00 & 50.60 & 23.20 & 52.00 & 36.80 & 55.00 & 67.80 & 51.40 \\
    & \textit{Avg Code} & \textit{\textbf{55.99}} & \textit{54.42} & \textit{17.94} & \textit{60.34} & \textit{32.40} & \textit{56.13} & \textit{64.39} & \textit{62.62} \\
    \midrule
    \multirow{7}{*}{Science}
    & GSM8K & 72.86 & 76.80 & 67.32 & 66.79 & 29.72 & 75.36 & 85.52 & 82.56 \\
    & GSM-Plus & 50.38 & 51.58 & 44.58 & 43.71 & 16.12 & 51.21 & 64.17 & 60.04 \\
    & MATH & \textbf{62.80} & 39.60 & 17.80 & 29.40 & 9.00 & 37.20 & 50.40 & 48.00 \\
    & GPQA-Main & 32.37 & 37.05 & 30.80 & 29.91 & 29.46 & 31.47 & 38.17 & 41.07 \\
    & SuperGPQA & 19.56 & 21.84 & 1.67 & 15.53 & 3.18 & 18.40 & 28.81 & 35.59 \\
    & MMLU-STEM & 53.41 & 60.20 & 53.63 & 44.12 & 47.64 & 61.91 & 75.36 & 72.09 \\
    & MMLU-Pro-STEM & \textbf{46.70} & 34.77 & 23.77 & 20.52 & 22.03 & 37.00 & 54.73 & 53.15 \\
    & \textit{Avg Science} & \textit{\textbf{48.30}} & \textit{45.98} & \textit{34.22} & \textit{35.71} & \textit{22.45} & \textit{44.65} & \textit{56.74} & \textit{56.07} \\
    \midrule
    \multicolumn{2}{l}{Overall Average} & \textbf{51.72} & 51.65 & 42.75 & 44.82 & 37.58 & 51.44 & 58.83 & 58.55 \\
    \bottomrule
  \end{tabular*}
\end{table*}

Table~\ref{tab:main-results} presents the comprehensive evaluation 
results of \textsc{daVinci-3B} compared to baseline models across all 
19 tasks. \textsc{daVinci-3B} achieves 51.72 overall average, matching 
OLMo-3 7B despite having less than half the parameters (3B vs. 7B), and significantly outperforming parameter-matched baselines including LLaMa-3.2-3B and Yulan-Mini-2.4B. 

Particularly notable is the model's strong reasoning performance: 
MATH exceeds the 7B-scale OLMo-3 by over 
23 points; code generation achieves 55.99 average (matching 
OLMo-3 7B's 54.42); science reasoning reaches 48.30 (exceeding 
OLMo-3 7B's 45.98). Importantly, this reasoning capability 
enhancement is achieved while maintaining general knowledge 
competence comparable to larger baselines, indicating 
no catastrophic forgetting during specialized training.

These results validate the effectiveness of our systematic, evidence-based pretraining methodology, informed by three key findings:
(1) \textbf{Data processing depth} (Section~\ref{sec:data-processing})
Advancing from L3 filtering to L4 refinement to L5 synthesis enables 
substantial capability gains in reasoning-intensive domains, with quality 
enhancement outweighing naive volume scaling; 
(2) \textbf{Adaptive training dynamics}(Section~\ref{sec:training_dynamics}) Monitoring differential saturation rates across capabilities and progressively adapting data composition (from domain adjustment to structured QA introduction) sustains growth beyond homogeneous data regime limits; 
(3) \textbf{Data Mixture optimization} (Section~\ref{sec:mixture-design}) Balancing aggressive reasoning-data concentration with capability 
preservation through adaptive composition prevents catastrophic forgetting while maximizing targeted enhancement.

Together, these findings demonstrate that systematic, question-driven 
investigation of pretraining dynamics can substantially improve base model capabilities. Our work shows that careful examination of data quality hierarchies, capability-specific saturation patterns, and mixture trade-offs enables more effective pretraining, suggesting significant headroom for base model improvement through principled, scientific exploration.

\section{Exploration: Why We Trained It This Way}
\label{sec:exploration}
While the preceding sections detailed the final training recipe of \textsc{daVinci-LLM}, this section provides the systematic investigation behind those decisions. We move beyond presenting a finalized configuration as a settled convention and instead adopt a question-driven approach to document the model's evolutionary path. By disclosing 200+ controlled ablations, we aim to elevate pretraining from an intuition-led craft to an evidence-based discipline, providing a transparent record of not only what worked, but how design choices were informed by rigorous empirical observation. This documentation serves as an empirical substrate, offering a granular view into the patterns of capability development and the strategic trade-offs encountered throughout the pretraining process.

We structure our exploration around three primary investigative themes:
\begin{enumerate}[leftmargin=*,itemsep=2pt]
    \item \textbf{Data Processing Depth} (Section~\ref{sec:data-processing}): 
    We evaluate how the hierarchical progression of data processing systematically enhances model capabilities, establishing processing depth as a pivotal dimension for exploration alongside data volume scaling.

    \item \textbf{Training Dynamics and Adaptation} (Section~\ref{sec:training_dynamics}): 
    We analyze how distinct capability saturation patterns necessitate adaptive data strategies, examining the transition from domain-proportion adjustments to structural format shifts as the effectiveness of raw text diminishes.

    \item \textbf{Intensification and Preservation} (Section~\ref{sec:mixture-design}): 
    We investigate the tension between targeted capability enhancement and the maintenance of general competence, identifying how compositional balance enables aggressive data intensification without triggering representational collapse.
\end{enumerate}

Our investigation concludes with an analysis of \textbf{Evaluation Validity} (Section~\ref{sec:benchmark_stability}), demonstrating how evaluation protocols themselves can shape our understanding of model quality. 
By providing this comprehensive disclosure, we offer the empirical evidence and decision-making logic behind our recipe, contributing to a more evidence-based and systematic understanding of pretraining science.

\subsection{Data Processing Depth: From Filtering to Synthesis}
\label{sec:data-processing}

\vspace{3pt}
\begin{takeawaybox}{\textsc{Section Takeaway}}
\textbf{Hierarchical data processing, progressing from quality filtering to content refinement to cognitive synthesis, enables systematic capability development.} Within the Darwin framework, L3 model-based filtering yields modest but consistent gains, validating progression beyond rule-based approaches. L4 generative refinement delivers substantial improvements for complex reasoning, demonstrating that content transformation outweighs volume expansion. L5 cognitive completion enables targeted capability steering through domain-aligned synthesis. \textbf{These results establish processing depth as a systematic optimization dimension, offering a principled alternative to naive scaling}: hierarchical data processing can substitute for multi-fold data volume increases.
\end{takeawaybox}

While data quality's importance for reasoning capabilities is increasingly recognized, the field lacks systematic understanding of \textit{how processing depth shapes effectiveness}. Section~\ref{sec:data_darwin} introduced the Data Darwinism L0-L9 taxonomy, characterizing operations from basic filtering to active synthesis. 
Yet a critical question remains unanswered: \textbf{\textit{Does data processing depth systematically improve reasoning capabilities, and what effectiveness patterns vary across the hierarchy?}} 
Answering this determines when practitioners should invest in advanced processing versus expanding data volume, and whether systematic quality enhancement can substitute for massive scale increases.

We conduct controlled ablations across three hierarchical levels, holding data volume and training compute constant to isolate processing depth's causal effect. L3 model-based filtering (Section~\ref{sec:code-l3-filtering}) removes low-quality content via LLM-based classifiers, preserving existing material unchanged. L4 generative refinement (Section~\ref{sec:math-l4-cleaning}) employs frontier LLMs to transform content—extracting key concepts, removing noise, reorganizing structure—while maintaining semantic integrity. L5 cognitive completion (Section~\ref{sec:l5-synthetic-qa}) actively synthesizes new reasoning chains through domain-specific QA generation, moving beyond refinement to targeted capability steering. Our experiments reveal distinct effectiveness patterns: L3 provides consistent but modest improvements across tasks, with notable gains on foundational programming (+3.40 on MBPP); L4 delivers substantial gains on complex reasoning (+7.00 on MATH), though effects remain task-specific; L5 enables domain-targeted steering, with synthetic QA exhibiting strong source-target alignment but limited cross-domain transfer. These findings establish a systematic pathway from filtering to synthesis for data-centric capability improvement.

\subsubsection{Code Data Filtering: L3 Model-Based Filtering}
\label{sec:code-l3-filtering}

\begin{table}[t]
  \centering
  \caption{Comparison of rule-based and model-based filtering approaches.}
  \label{tab:code-filtering}
  \small
  \setlength{\tabcolsep}{3pt}
  \begin{tabular}{l c c c c c c c}
    \toprule
    Configuration & HumanEval & EvalPlus & MBPP & Avg Gen. & Avg Code & Avg Sci. & Overall \\
    \midrule
    Rule-Based Filtering (L2) & 54.43 & 48.58 & 42.40 & 51.12 & 48.47 & 40.14 & 46.66 \\
    Model-Based Filtering (L3)   & \textbf{54.70} & 48.39 & \textbf{45.80} & 51.19 & \textbf{49.63} & \textbf{40.28} & \textbf{46.93} \\
    \midrule
    $\Delta$ (L3 -- L2) & +0.27 & -0.19 & \cellcolor{blue!7}\textbf{+3.40} & +0.07 & +1.16 & +0.14 & +0.27 \\
    \bottomrule
  \end{tabular}
\end{table}

To evaluate whether model-based filtering (L3) provides measurable improvements over rule-based approaches (L2), we compare their effectiveness on code generation tasks. We apply GPT-OSS-120B quality scoring following SeedCoder's assessment framework, which identifies and removes low-quality code artifacts: configuration files with extensive hard-coded data, data files dominated by constants, code with minimal logic, and auto-generated content. We train for 500B tokens, comparing L3 model-based filtering against the L2 rule-based baseline.

Table~\ref{tab:code-filtering} shows that L3 model-based filtering provides consistent but modest improvements across most metrics. The most notable gain appears on MBPP, while improvements on HumanEval and EvalPlus remain minimal. This differential pattern, where foundational programming tasks benefit more substantially than advanced algorithmic challenges, suggests that quality filtering's impact depends on task complexity. One plausible interpretation is that basic programming competence is more sensitive to training example clarity, as models acquiring fundamental patterns benefit from removing noisy artifacts like configuration boilerplate. However, the modest overall magnitude indicates that advancing from L2 to L3 yields incremental rather than transformative gains.

These results validate that model-based quality assessment (L3) provides measurable advantages over pure rule-based approaches (L2), confirming the value of advancing processing depth within the Darwin framework. However, the limited improvements, particularly on sophisticated coding benchmarks, suggest that filtering alone may be insufficient to unlock the full potential of code training data. More intensive processing interventions, such as generative refinement to improve code clarity and documentation, may be necessary to achieve breakthrough gains in code capabilities.

\subsubsection{Math Data Quality: L4 Generative Refinement}
\label{sec:math-l4-cleaning}

\begin{table}[t]
  \centering
  \caption{Comparison of baseline and generative refinement approaches.}
  \label{tab:math-cleaning}
  \small
  \setlength{\tabcolsep}{3pt}
  \begin{tabular}{l c c c c c c c}
    \toprule
    Configuration & GSM-8K & GSM-Plus & MATH & Avg Gen. & Avg Code & Avg Sci. & Overall \\
    \midrule
    Baseline & 64.06 & 40.58 & 38.00 & \textbf{51.69} & 49.80 & 40.52 & 47.27 \\
    Generative Refined (L4) & \textbf{65.43} & \textbf{42.38} & \textbf{45.00} & 51.40 & \textbf{50.15} & \textbf{42.25} & \textbf{47.83} \\
    \midrule
    $\Delta$ (L4 -- Baseline) & +1.37 & +1.80 & \cellcolor{olmopink!7}\textbf{+7.00} & -0.29 & +0.35 & +1.73 & +0.56 \\
    \bottomrule
  \end{tabular}
\end{table}

Section~\ref{sec:code-l3-filtering} indicates that L3 filtering offers marginal returns, prompting an investigation into whether active content transformation (L4) can further enhance reasoning performance. To test this, we compared L4-refined mathematical data against the original MegaMath and Nemotron-CC-Math baseline, holding data volume constant at 500B tokens. We utilize Qwen3-235B to perform generative refinement, which systematically prunes narrative noise, extracts core formulas, and reorganizes fragmented steps into logically structured proofs while strictly preserving semantic integrity.

As shown in Table~\ref{tab:math-cleaning}, L4 refinement yields substantial improvements across all mathematical benchmarks, though the gains are markedly uneven. The MATH benchmark achieved a significant +7.00 increase, while the improvement on GSM8K was relatively modest (+1.37). This divergence suggests that structural purification, specifically the transition from messy exposition to pedagogical logic, disproportionately benefits complex, multi-step reasoning. While basic word problems in GSM8K follow relatively simple and linear solution paths, the abstract problems in MATH are highly sensitive to the clarity of the underlying logical flow. By stripping away redundant exposition and reinforcing the connection between reasoning steps, L4 processing effectively reduces the difficulty for models to extract fundamental patterns from complex mathematical text, leading to a performance leap in advanced reasoning tasks.

\subsubsection{L5 Synthetic QA: Cognitive Completion}
\label{sec:l5-synthetic-qa}
\begin{figure}[t]
  \centering
  \includegraphics[width=0.999\linewidth]{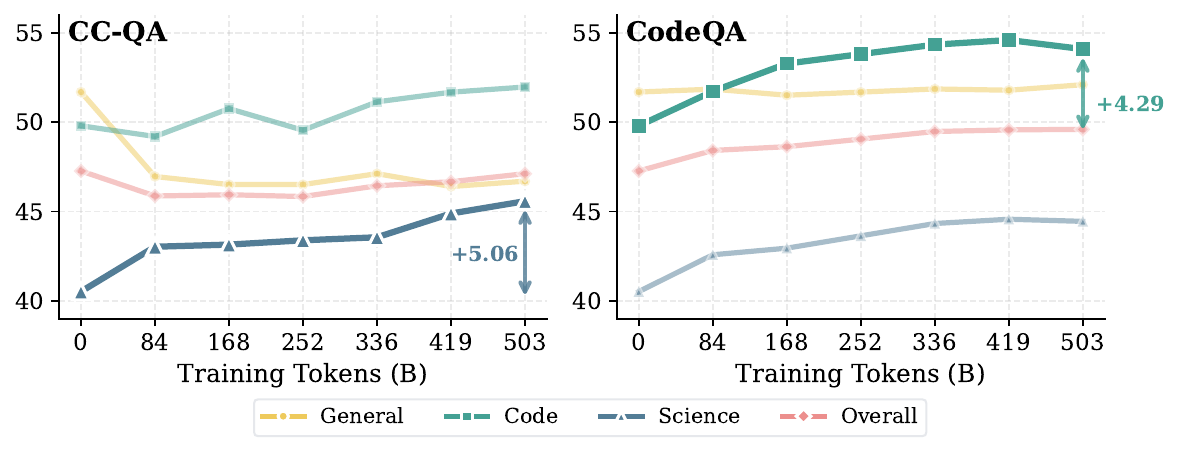}
  \caption{Effectiveness of CC-QA and CodeQA synthetic data in Stage 2-2.} 
  \label{fig:synthetic-qa}
\end{figure}

While L4 refinement reorganizes existing content, L5 cognitive completion involves the active synthesis of new reasoning chains to bridge the gap between implicit information and explicit problem-solving steps. We investigate L5 processing by incorporating synthetic QA data from two domain-specific sources: CodeQA, derived from curated repositories, and CC-QA, extracted from high-quality web text. Unlike previous levels that modify existing material, L5 synthesis allows generative models to construct structured reasoning scaffolds that may be absent in raw corpora.

The experimental results in Figure~\ref{fig:synthetic-qa} reveal a strong pattern of domain-specific steering: CodeQA substantially enhances programming benchmarks, while CC-QA primarily strengthens science reasoning. This source-target alignment suggests that the impact of synthetic reasoning is largely governed by its originating domain. For instance, code-domain QA synthesizes algorithmic structures that directly benefit programming but offer limited cross-domain transfer to unrelated areas like general knowledge or science. This limited generalization indicates that L5 synthesis acts as a high-precision tool for capability steering rather than a universal performance booster.

These findings establish that strategic synthesis can be as effective as massive data scaling. By generating data that mirrors specific reasoning structures, practitioners can deliberately guide model performance toward target domains. This positions L5 processing as a principled alternative to exhaustive data accumulation, offering a more efficient path to capability development.

%\subsection{Data Strategy Evolution Across Training Stages}
%\subsection{Training Dynamics and Adaptive Data Strategies}
\subsection{Training Dynamics: Adaptive Data Strategies}
\label{sec:training_dynamics}

\begin{takeawaybox}{\textsc{Section Takeaway}}
\textbf{Effective pretraining requires stage-specific data strategies 
guided by capability dynamics.} 
Systematic convergence tracking reveals that different capabilities exhibit vastly different saturation timescales, enabling adaptive interventions that reallocate compute toward actively learning capabilities. 
However, domain proportion adjustments encounter fundamental limitations once standard corpus formats collectively approach saturation, at which point, reallocating among these data types no longer suffices. 
Stage~1's diminishing returns from proportion adjustments motivated introducing structured question-answer data in Stage~2, which substantially outperforms continued proportion optimization. These findings establish that \textbf{no single data mixture or format suffices across extended training}: sustained capability development demands monitoring convergence patterns and adapting both domain proportions and data formats accordingly.
\end{takeawaybox}

Section~\ref{sec:data-processing} established that advancing data processing depth enables systematic quality improvements. However, even with high-quality processed data, how to compose them optimally remains a critical question. Traditional approaches apply uniform strategies with fixed data mixtures over predetermined token budgets, implicitly assuming all capability dimensions develop at similar timescales. Yet if general knowledge saturates rapidly while reasoning capabilities require sustained training, maintaining static mixtures may waste compute on converged dimensions while under-serving actively-learning ones. 

This motivates our central question: \textbf{\textit{How should data composition evolve as different capabilities mature at different rates?}} Specifically, we investigate two aspects: (1) Do different capability dimensions exhibit systematically different saturation patterns? We investigate whether tracking domain-specific convergence enables principled intervention timing and adaptive data composition adjustments. (2) What adaptation strategies work when capabilities diverge? We examine whether domain proportion adjustment suffices, or whether fundamental data format shifts become necessary as standard corpus formats collectively approach saturation.

We address these through systematic convergence tracking across Stage 1 (6T tokens) and Stage 2 (2T tokens). Section~\ref{sec:domain_proportion} examines domain-specific saturation dynamics within Stage 1, revealing differential convergence rates (general knowledge plateaus at 1T tokens; code/science sustain growth through 4T tokens) that motivate adaptive domain proportion adjustments and the boundaries of such strategies. Section~\ref{sec:introduce_qa} investigates the transition from domain adjustment to structured QA introduction in Stage 2, establishing when proportion optimization suffices versus when data format shifts become necessary.

\subsubsection{Domain Proportion Adjustment}
\label{sec:domain_proportion}

\begin{figure}[t]
  \centering
  \includegraphics[width=0.999\linewidth]{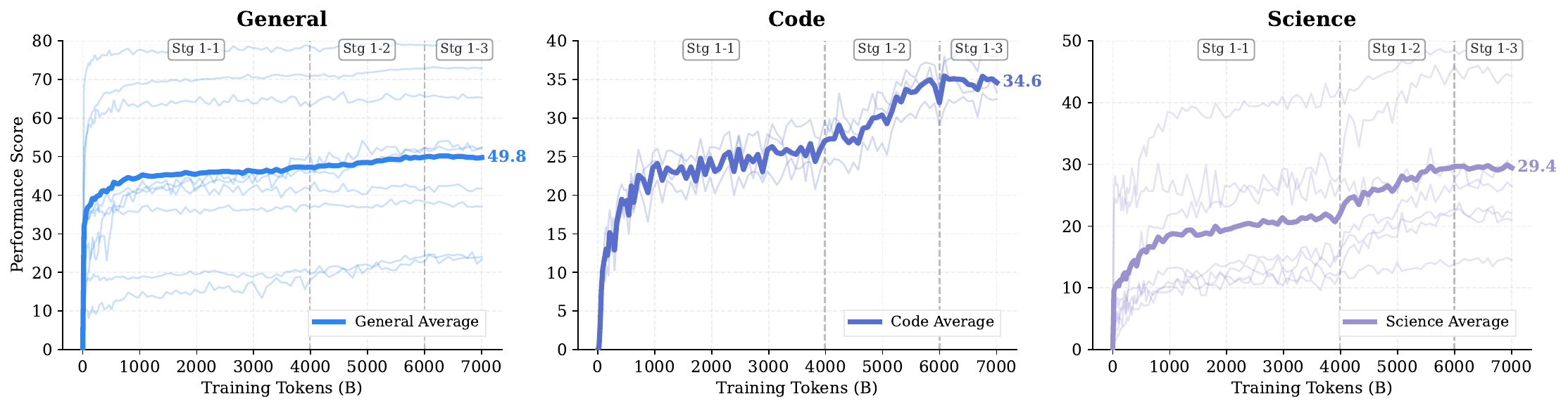}
  \caption{Stage 1 benchmark performance. Light curves show individual benchmarks; dark curves show category averages for general knowledge, code, and science. Two vertical dashed lines indicate the boundaries between substages 1-1, 1-2, and 1-3, respectively.}
  \label{fig:stage1-evaluation}
\end{figure}

Do different capability dimensions saturate at different rates, and can we use this to guide adaptive training? We investigate this through Stage~1's training trajectory (6T tokens), tracking capability-specific convergence patterns across general knowledge, code, and scientific domains. We examine whether domain proportion adjustments, reallocating compute from saturated to actively-learning capabilities, can sustain capability growth, and where such strategies encounter fundamental limitations.

\paragraph{Differential convergence motivates adaptive mixture adjustment.}
Figure~\ref{fig:stage1-evaluation} shows Stage 1's 6T-token trajectory, revealing distinct convergence patterns across data domains. General knowledge benchmarks plateau rapidly within the first 1T tokens, with minimal subsequent improvement. In contrast, reasoning-intensive benchmarks, both code and scientific domains, sustain consistent growth through the initial 4T tokens, though improvement rates begin to slow. Notably, benchmarks within each domain exhibit consistent convergence behavior: all general knowledge tasks saturate early, while all reasoning tasks maintain growth trajectories, confirming these patterns reflect domain-level characteristics rather than task-specific idiosyncrasies.

This differential saturation motivated Stage 1-2's adaptive adjustment strategy at the 4T checkpoint. With general knowledge approaching saturation while reasoning capabilities remained actively improving, we reduced generic web data proportion and increased code/science concentrations, complemented by transitioning to cosine learning rate decay. Figure~\ref{fig:stage1-evaluation} shows that this adjustment successfully amplified reasoning domain gains in Stage 1-2 (4-6T tokens): both code and science benchmarks exhibit renewed acceleration, while general knowledge maintains stable performance without degradation. These results validate adaptive mixture strategies that reallocate compute from saturated domains (general knowledge) toward capabilities exhibiting sustained learning potential (code and science reasoning).

\paragraph{Domain adjustment encounters saturation boundaries.}
Encouraged by Stage~1-2's success, we explored further domain proportion adjustments, referred to as Stage 1-3, in Stage 1's later phase to sustain reasoning capability growth. As code and science
performance began to decelerate again near 6T tokens, we attempted additional increases in code/science concentrations while further reducing generic web data.\footnote{Stage 1-3 configuration: CC 51.42\%, Code 13.64\%, Math 13.58\%, Science 21.36\%, with constant learning rate 3e-5.} However, figure~\ref{fig:stage1-evaluation} shows that these adjustments yielded only marginal improvements at Stage~1-3, with performance gains substantially smaller than those achieved during Stage~1-2's adjustment. This diminishing effectiveness reveals an inherent limitation: once standard pretraining corpora (web, code, science) collectively approaches saturation, reallocating proportions among these textual formats cannot overcome the fundamental constraint. 
These observations suggest that sustaining growth requires introducing new data formats, a transition we investigate in the following section.

\subsubsection{From Domain Adjustment to QA Introduction}
\label{sec:introduce_qa}
\begin{figure}[t]
  \centering
  \includegraphics[width=0.999\linewidth]{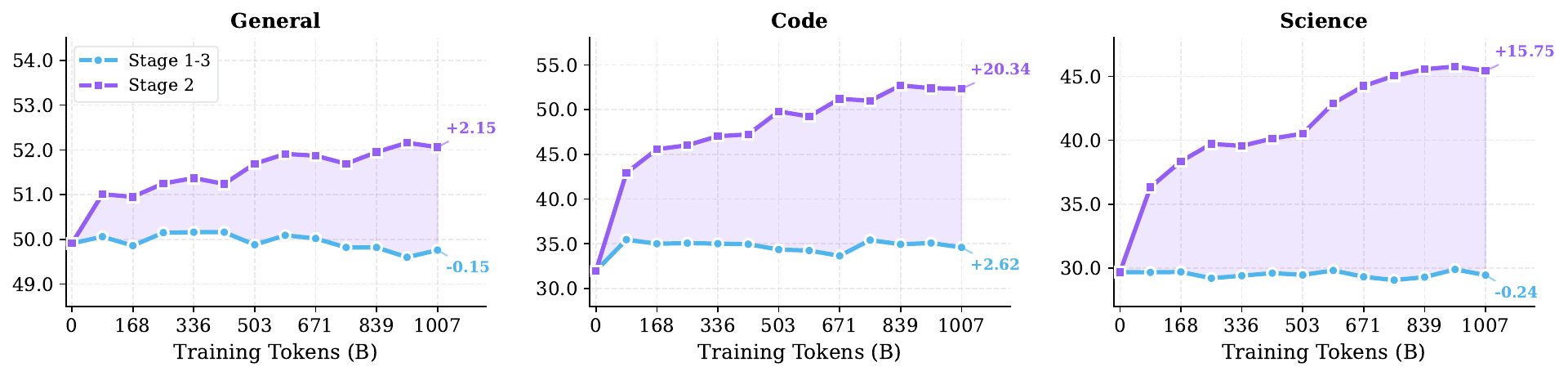}
  \caption{Performance comparison between continued data domain adjustment (stage 1-3) and QA introduction (Stage 2).}
  \label{fig:compare_stage_1_3_and_stage_2}
\end{figure}

\begin{figure}[t]
  \centering
  \includegraphics[width=0.999\linewidth]{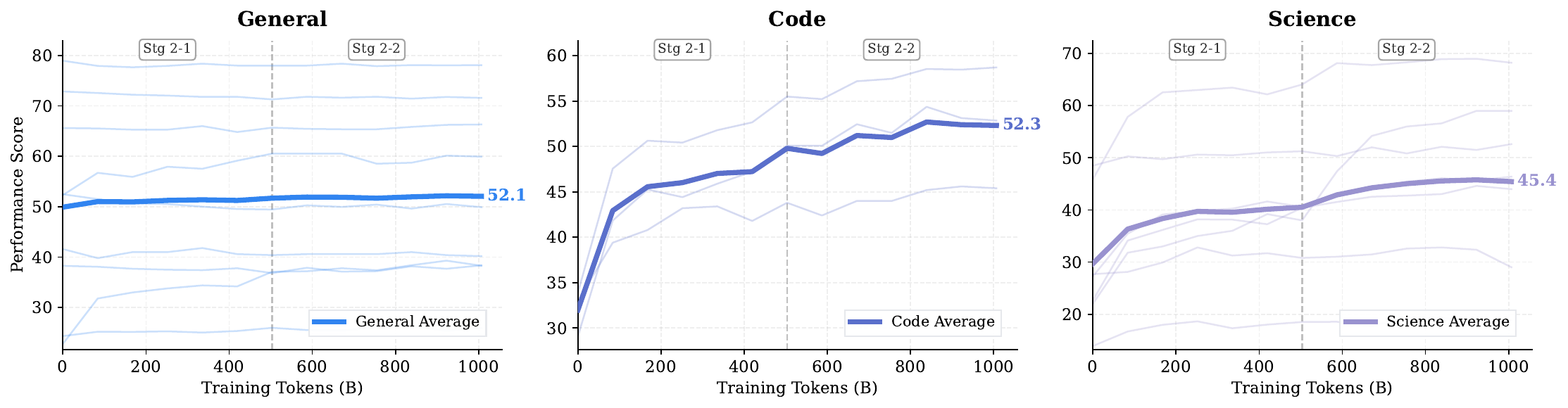}
  \caption{Stage 2 benchmark performance. Light curves show individual benchmarks; dark curves show category averages for general knowledge, code, and science. The vertical dashed lines indicate the boundaries between Substages 2-1, 2-2.}
  \label{fig:stage2-evaluation}
\end{figure}

Section~\ref{sec:domain_proportion} demonstrated that domain proportion adjustments encounter diminishing returns as standard pretraining corpora approach saturation. To overcome this limitation, we hypothesize that sustaining capability growth requires shifting from raw text to data with explicit reasoning structures. We therefore introduce structured question-answer pairs in Stage 2. 
Unlike raw text, QA data provides inherent problem-solving scaffolds: questions define the reasoning target, while answers supply direct supervision signals for multi-step inference. 
This format is uniquely suited for reasoning-intensive domains, precisely the domains that Section~\ref{sec:domain_proportion} showed require extended training investment.

\paragraph{Structured QA outperforms domain adjustment alone.}
To isolate the value of introducing QA data, we conducted a controlled comparison. Starting from the Stage 1-2 checkpoint, we examined two continuation paths: (1) Stage 1-3, which further increases code/science proportions without adding QA (as described in Section~\ref{sec:domain_proportion}), versus (2) Stage 2, which introduces 30\% QA.
Figure~\ref{fig:compare_stage_1_3_and_stage_2} presents the performance trajectories. Stage 2 substantially outperforms Stage 1-3 across reasoning domains: code and scientific benchmarks achieve significantly higher performance with QA introduction compared to mere domain adjustment. While general knowledge also exhibits a modest improvement, the gains are disproportionately concentrated in reasoning-intensive capabilities. This validates that structured data formats, rather than simple proportion adjustments, are essential to overcome the saturation of unstructured text regimes and unlock advanced cognitive potential.

\paragraph{Adaptive Adjustment within Stage 2 sustains growth.}
Having established QA's effectiveness, we examine Stage 2's internal dynamics. Figure~\ref{fig:stage2-evaluation} reveals convergence patterns consistent with Stage 1's hierarchy: 
general knowledge benchmarks exhibit near-complete saturation throughout Stage 2, showing minimal variation. 
In contrast, the code and scientific reasoning domains achieve substantial performance gains during Stage 2. 
When improvement rates for these domains begin to decelerate toward the end of Stage 2-1, we implement further mixture adjustment in Stage 2-2, increasing QA concentration to 70\%.
This successfully triggers renewed performance growth, demonstrating that adaptive mixture adjustment remains effective across training stages, with detailed mixture design explored in Section~\ref{sec:mixture-design}.

\paragraph{Implementation details beyond mixture design.}
%\huangz{move original LR-Decay and QA-masking section here as a simple paragraph linked to the appendix}
In addition to mixture adjustment, we examined two practical training decisions that could plausibly affect Stage~2 outcomes: the learning rate schedule in later substages and the masking policy for QA data. Empirically, cosine decay provides small but consistent gains over a constant learning rate, whereas masking question tokens yields only marginal improvements when the QA data is already diverse and high quality. We defer the full analysis to Appendix~\ref{sec:app-lr-masking}.

\subsection{Data Mixture Design: Balancing and Intensifying}
\label{sec:mixture-design}

\begin{takeawaybox}{\textsc{Section Takeaway}}
\textbf{Mixture optimization balances reasoning intensification with global capability preservation.} Our results show that while high reasoning data concentrations are essential, internal balance between domains is critical to prevent over-specialization sacrificing broader competence. Similarly, structured QA intensification requires navigating a stage-dependent tolerance: conservative ratios preserve stability during foundational training, while progressive concentration can be leveraged once a balanced representational base is established. These findings demonstrate that \textbf{concentration and preservation are not mutually exclusive, as effective mixture design achieves both through an adaptive composition that evolves with training progress.}
\end{takeawaybox}

Section~\ref{sec:training_dynamics} established that transitioning to structured reasoning formats is essential for surpassing the performance plateaus of unstructured text. However, the introduction of these data types raises a new challenge: the tension between \textbf{targeted intensification} and \textbf{capability preservation}. While aggressive concentration of reasoning-intensive data is necessary to drive performance, it also risks triggering catastrophic trade-offs where gains in one domain come at the expense of others. This leads to a critical research question: \textbf{\textit{How to manage the intensification of reasoning data to maximize gains without compromising the model's overall capability breadth?}}

We address this question through two ablation studies. Section~\ref{sec:code-math-balance} examines the internal composition of reasoning domains, documenting how an aggressive yet balanced allocation between code and science prevents over-specialization and yields more robust results than extreme domain-specific concentrations. Section~\ref{sec:qa-concentration} investigates the progressive intensification of structured QA data. We demonstrate that the model’s tolerance for high-concentration supervision is stage-dependent: a conservative mixture is required to maintain stability during foundational training (Stage 2-1), whereas an intensified regime can be successfully applied (Stage 2-2) once a balanced capability base has been established.

\subsubsection{Domain Balance: Code and Science Composition}
\label{sec:code-math-balance}

\begin{figure}[t]
  \centering
  \begin{subfigure}[t]{0.49\textwidth}
    \centering
    \includegraphics[width=\textwidth]{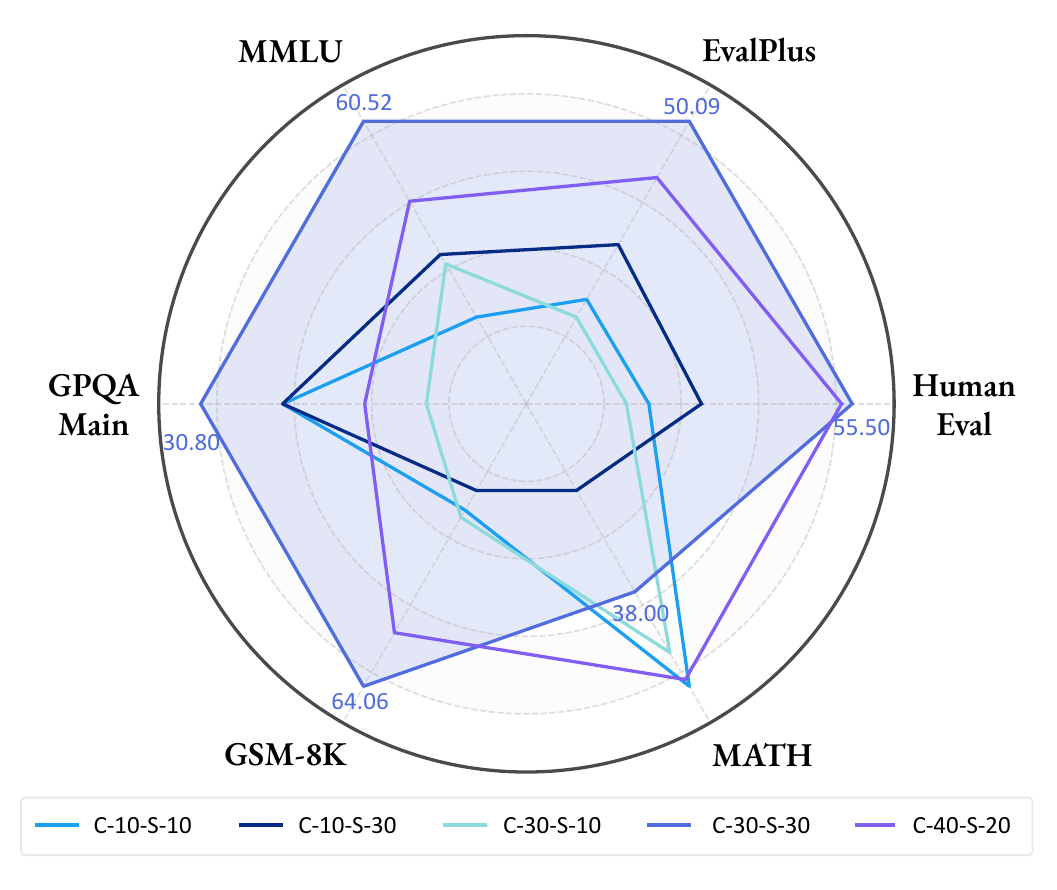}
    \caption{}
    \label{fig:code-ratio-radar}
  \end{subfigure}
  \hfill
  \begin{subfigure}[t]{0.49\textwidth}
    \centering
    \includegraphics[width=\textwidth]{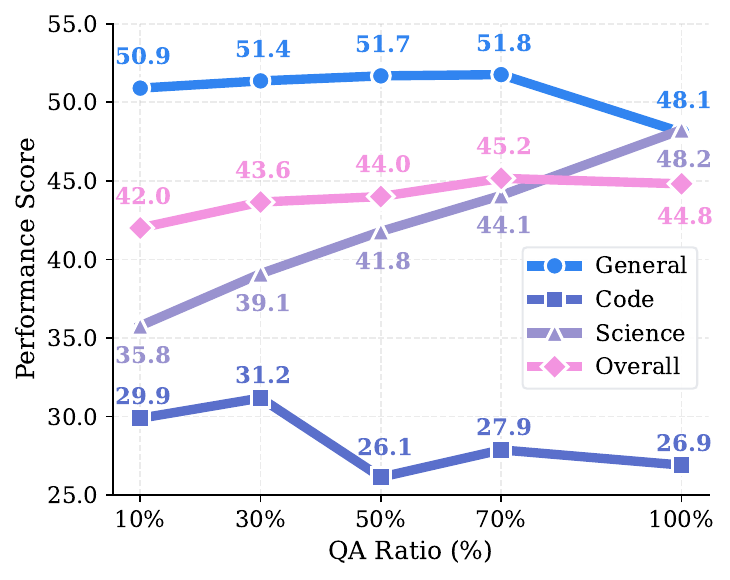}
    \caption{}
    \label{fig:data-synergy}
  \end{subfigure}
  \caption{Code and science data ratio analysis. Notation: C-X-S-Y indicates X\% code and Y\% science. (a) Performance across code and science proportions (QA fixed at 30\%, remainder allocated to CommonCrawl web data). (b) Data synergy effects across QA ratios in Stage 2-1.}
  \label{fig:code-ratio-and-synergy}
\end{figure}

Section~\ref{sec:domain_proportion} reveals that different domains exhibit distinct convergence rates during pretraining. Building on this observation, we investigate the internal composition of reasoning-intensive data (specifically code and science) when training jointly with QA. We fix QA concentration at 30\% and systematically vary code-science proportions: 10\%-10\%, 10\%-30\%, 30\%-10\%, 30\%-30\%, and 40\%-20\%, with the remainder allocated to CommonCrawl web data. All configurations are trained for 500B tokens from the Stage 1 checkpoint. We evaluate performance across general knowledge, code generation, and science reasoning, visualizing the results via radar plots (Figure~\ref{fig:code-ratio-radar}).

Two critical patterns emerge from this ablation. First, \textbf{reasoning-intensive intensification is essential for Stage 2 progress}: configurations allocating 60\% to code and science (C-30-S-30 and C-40-S-20) substantially outperform lower concentrations. This validates that transitioning from Stage 1's CC-dominated mixture to Stage 2's reasoning-heavy regime requires aggressive reallocation toward symbolic and logical domains. Second, \textbf{within high-concentration regimes, internal balance is critical for maintaining overall capability}: C-30-S-30 achieves superior overall performance compared to the code-heavy C-40-S-20. This indicates that while intensification is necessary, extreme specialization in a single domain can trigger trade-offs, whereas equal allocation between code and science maintains general competence while delivering competitive reasoning gains across all dimensions.

These findings establish the design principle for Stage 2-1: aggressive reasoning-data concentration with internal balance prevents under-resourcing of specific reasoning domains while preserving general capabilities. The balanced configuration (30\% QA, 30\% code, 30\% science, 10\% web text) forms the foundation for subsequent QA intensification strategies examined in the next section.

\subsubsection{QA Concentration: Progressive Intensification}
\label{sec:qa-concentration}

Section~\ref{sec:code-math-balance} established balanced proportions within the reasoning domains. We now investigate the concentration of structured QA itself, revealing stage-dependent optimal ratios: conservative choices in Stage 2-1 prioritize stability, while higher concentrations in Stage 2-2 leverage established foundations to safely amplify reasoning gains.

\begin{table}[t]
  \centering
  \caption{Impact of QA data ratio on model performance in Stage 2-2.}
  \label{tab:qa-ratio-ablation-stage22}
  \small
  \setlength{\tabcolsep}{3pt}
  \begin{tabular}{c c c c c c}
    \toprule
    Configuration & Training Tokens & General & Code & Science & Overall \\
    \midrule
    \texttt{QA-30\%} & 84B & 51.59 & 47.71 & 40.25 & 46.80 \\
     & 252B & 51.66 & 46.79 & 40.03 & 46.61 \\
     & 419B & 51.80 & 47.16 & 40.59 & 46.94 \\
    \midrule
    \texttt{QA-50\%} & 84B & 51.72 & 49.86 & 41.60 & 47.70 \\
     & 252B & 51.85 & 49.01 & 42.80 & 48.06 \\
     & 419B & 51.83 & 49.89 & 43.42 & 48.43 \\
    \midrule
    \texttt{QA-70\%} & 84B & 51.91 & 49.23 & 42.89 & 48.17 \\
     & 252B & 51.69 & 50.99 & 45.04 & 49.13 \\
     & 419B & \textbf{52.16} & \textbf{52.40} & \textbf{45.77} & \textbf{49.84} \\
    \bottomrule
  \end{tabular}
\end{table}

\paragraph{Stage 2-1: Conservative QA choice prioritizes capability balance.}

We conduct a systematic investigation in Stage 2-1, training configurations with 10\% to 100\% QA. As shown in Figure~\ref{fig:data-synergy}, general knowledge remains stable across moderate concentrations but degrades sharply at 100\% QA, indicating that exclusive structured exposure cannot sustain broad linguistic competence. Notably, code performance exhibits a non-monotonic behavior: improving initially but collapsing beyond the 30\% threshold. In contrast, science reasoning improves monotonically.

This divergence between science and code merits further investigation. We hypothesize that this stems from the compositional imbalance of our QA corpus: our Stage 2-1 QA pool is heavily weighted toward science (approx. 80\%) , while code-related QA is relatively scarce (approx. 26B tokens). At higher total QA concentrations, the insufficient diversity of code samples may trigger premature over-fitting or representation collapse, whereas the abundant science data sustain growth. This suggests a critical requirement for scaling QA: performance is governed not just by the total ratio, but by the absolute diversity of the supervised signals in each domain. 

In our final Stage 2-1 recipe, we select 30\% QA to maintain balanced capabilities. Although 70\% QA yields higher average peak scores, the resulting code degradation suggests that a conservative foundation-building phase is essential to prevent irreversible domain loss before further intensification.

\paragraph{Stage 2-2: Foundation enables aggressive QA intensification without collapse.}
Continuing from the Stage 2-1 checkpoint, we investigated higher QA concentrations in Stage 2-2 (30\%, 50\%, and 70\%). In this phase, performance increases monotonically across all clusters, including code. This contrasts sharply with Stage 2-1, where increasing QA beyond 30\% triggered collapse. We hypothesize that Stage 2-1’s balanced training establishes a necessary foundation that enables the model to effectively internalize high-intensity supervision without capability loss.

These findings validate a progressive mixture strategy: Stage 2-1 employs a conservative 30\% QA ratio to build stability, while Stage 2-2 intensifies to 70\% QA for targeted enhancement. This staged approach achieves what static mixtures cannot by preserving broad competence while delivering significant reasoning gains. The complete trajectory demonstrates that no single mixture suffices: sustained development requires adapting data composition across training stages.

\subsection{Evaluation Validity: PPL-based vs. Generative-based}
\label{sec:benchmark_stability}

% \huangz{Omit original 'Evaluation Stability' section as the analysis maybe a bit narrow.}

\begin{takeawaybox}{\textsc{Takeaway}}
\textbf{Evaluation design critically impacts conclusions drawn about base model pretraining:} \\
The choice between PPL-based and generative evaluation is not merely technical---these protocols probe different aspects of model capability, and models with extensive QA pretraining can exhibit ranking reversals across them. Benchmark selection and evaluation methodology must therefore be validated jointly against the intended use case.
\end{takeawaybox}

Previous sections focused on the \emph{training side}: how data mixture design, QA proportion, and data quality shape model behavior during continued pretraining. We now turn to the \emph{evaluation side}. Even when training settings are fixed, the conclusions we draw about model quality can vary substantially depending on how the model is evaluated. In particular, base models are commonly assessed under two distinct protocols: \textbf{PPL-based evaluation}, which measures whether the model assigns higher likelihood to the correct answer, and \textbf{generative evaluation}, which requires the model to actively produce an answer. These protocols are often treated as interchangeable, but they in fact probe different aspects of capability and can lead to different model rankings. Therefore, it is necessary to clarify what each evaluation setting is actually measuring and which one better matches the intended use case.

\begin{figure}[t]
  \centering
  \includegraphics[width=0.999\linewidth]{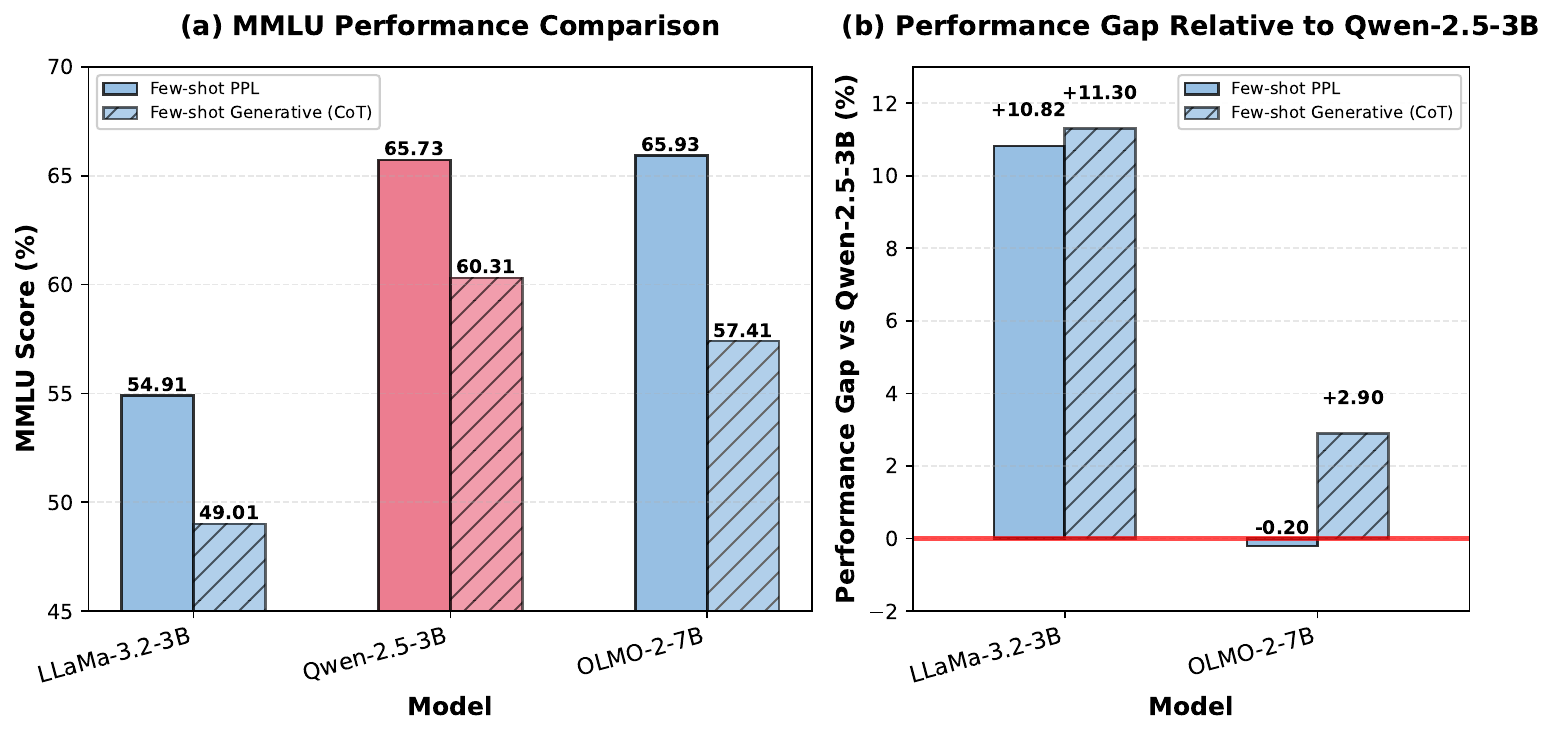}
  \caption{MMLU performance comparison across evaluation protocols. Panel (a) shows absolute scores under PPL-based and generative (CoT) evaluation. Panel (b) highlights the performance gap relative to Qwen-2.5-3B, demonstrating how models with extensive QA training exhibit amplified advantages in generative settings. The gap between OLMO-2-7B and Qwen-2.5-3B increases from -0.20\% (PPL) to +2.90\% (generative), a 3.10\% swing that reverses the ranking.}
  \label{fig:eval-protocol-comparison}
\end{figure}

Figure~\ref{fig:eval-protocol-comparison} illustrates this difference on MMLU across three representative base models. Under PPL-based evaluation, OLMO-2-7B slightly outperforms Qwen-2.5-3B; under generative evaluation, the ranking reverses in Qwen-2.5-3B's favor---a 3.10\% swing. We argue that this discrepancy arises because the two protocols place different demands on the model. \textbf{PPL-based evaluation is closer to latent knowledge access}: the model only needs to assign relatively higher probability to the correct candidate among presented options. By contrast, \textbf{generative evaluation requires the model to surface and organize knowledge into an explicit answer}, often under additional reasoning demands. In this sense, generative evaluation is more sensitive not only to what the model knows, but also to whether it has learned to operationalize that knowledge through answer production. This distinction helps explain why models with heavier QA exposure during pretraining tend to gain disproportionately under generative evaluation. QA-style pretraining does not merely inject factual content; it also trains the model in the behavioral format of mapping questions to explicit answers. As a result, two models with similar underlying knowledge may appear much closer under PPL-based evaluation, yet diverge substantially when evaluated generatively. The ranking reversal between OLMO-2-7B and Qwen-2.5-3B is a concrete example of this effect.

Framing one protocol as inherently superior would therefore be misleading. The appropriate choice depends on the deployment scenario. Applications requiring direct answer generation--such as chatbots or open-ended QA systems--are better matched by generative evaluation, whereas applications that use language models as scoring or ranking functions are better aligned with PPL-based evaluation. More broadly, when comparing base models, substantial discrepancies across protocols should not be dismissed as noise: they often indicate meaningful differences in pretraining data composition, especially QA exposure, and should be reported transparently. Evaluating base models under both protocols provides a more complete capability profile and helps avoid protocol-induced ranking artifacts.

\section{Conclusion}
We have presented \textsc{daVinci-LLM}, a systematic investigation of pretraining dynamics through controlled experimentation and comprehensive transparency. Through the Data Darwinism framework and 200+ controlled ablations, we establish empirical understanding of data processing depth, training dynamics, and mixture design, releasing complete pipelines and exploration results to the community. This work represents a step toward transparency, and transforming pretraining into rigorous scientific discipline requires community-wide open collaboration. Only when systematic exploration becomes the norm and empirical knowledge accumulates across organizations can pretraining advance from intuition-driven practice toward genuine scientific understanding.

%% ============================================================
%% Appendix
%% ============================================================
\clearpage

%% ============================================================
%% Bibliography
%% ============================================================
\bibliographystyle{unsrtnat}
\bibliography{bib}

@article{hua2025context,
  title={Context engineering 2.0: The context of context engineering},
  author={Hua, Qishuo and Ye, Lyumanshan and Fu, Dayuan and Xiao, Yang and Cai, Xiaojie and Wu, Yunze and Lin, Jifan and Wang, Junfei and Liu, Pengfei},
  journal={arXiv preprint arXiv:2510.26493},
  year={2025}
}

@article{qin2026data,
  title={Data Darwinism Part I: Unlocking the Value of Scientific Data for Pre-training},
  author={Qin, Yiwei and Huang, Zhen and Mi, Tiantian and Si, Weiye and Zhou, Chenyang and Guo, Qipeng and Feng, Siyuan and Liu, Pengfei},
  journal={arXiv preprint arXiv:2602.07824},
  year={2026}
}

@misc{openai2025gptoss120bgptoss20bmodel,
      title={gpt-oss-120b \& gpt-oss-20b Model Card}, 
      author={OpenAI},
      year={2025},
      eprint={2508.10925},
      archivePrefix={arXiv},
      primaryClass={cs.CL},
      url={https://arxiv.org/abs/2508.10925}, 
}

@misc{nvidia2025nvidianemotronnano2,
      title={NVIDIA Nemotron Nano 2: An Accurate and Efficient Hybrid Mamba-Transformer Reasoning Model}, 
      author={NVIDIA and : and Aarti Basant and Abhijit Khairnar and Abhijit Paithankar and Abhinav Khattar and Adithya Renduchintala and Aditya Malte and Akhiad Bercovich and Akshay Hazare and Alejandra Rico and Aleksander Ficek and Alex Kondratenko and Alex Shaposhnikov and Alexander Bukharin and Ali Taghibakhshi and Amelia Barton and Ameya Sunil Mahabaleshwarkar and Amy Shen and Andrew Tao and Ann Guan and Anna Shors and Anubhav Mandarwal and Arham Mehta and Arun Venkatesan and Ashton Sharabiani and Ashwath Aithal and Ashwin Poojary and Ayush Dattagupta and Balaram Buddharaju and Banghua Zhu and Barnaby Simkin and Bilal Kartal and Bita Darvish Rouhani and Bobby Chen and Boris Ginsburg and Brandon Norick and Brian Yu and Bryan Catanzaro and Charles Wang and Charlie Truong and Chetan Mungekar and Chintan Patel and Chris Alexiuk and Christian Munley and Christopher Parisien and Dan Su and Daniel Afrimi and Daniel Korzekwa and Daniel Rohrer and Daria Gitman and David Mosallanezhad and Deepak Narayanan and Dima Rekesh and Dina Yared and Dmytro Pykhtar and Dong Ahn and Duncan Riach and Eileen Long and Elliott Ning and Eric Chung and Erick Galinkin and Evelina Bakhturina and Gargi Prasad and Gerald Shen and Haifeng Qian and Haim Elisha and Harsh Sharma and Hayley Ross and Helen Ngo and Herman Sahota and Hexin Wang and Hoo Chang Shin and Hua Huang and Iain Cunningham and Igor Gitman and Ivan Moshkov and Jaehun Jung and Jan Kautz and Jane Polak Scowcroft and Jared Casper and Jian Zhang and Jiaqi Zeng and Jimmy Zhang and Jinze Xue and Jocelyn Huang and Joey Conway and John Kamalu and Jonathan Cohen and Joseph Jennings and Julien Veron Vialard and Junkeun Yi and Jupinder Parmar and Kari Briski and Katherine Cheung and Katherine Luna and Keith Wyss and Keshav Santhanam and Kezhi Kong and Krzysztof Pawelec and Kumar Anik and Kunlun Li and Kushan Ahmadian and Lawrence McAfee and Laya Sleiman and Leon Derczynski and Luis Vega and Maer Rodrigues de Melo and Makesh Narsimhan Sreedhar and Marcin Chochowski and Mark Cai and Markus Kliegl and Marta Stepniewska-Dziubinska and Matvei Novikov and Mehrzad Samadi and Meredith Price and Meriem Boubdir and Michael Boone and Michael Evans and Michal Bien and Michal Zawalski and Miguel Martinez and Mike Chrzanowski and Mohammad Shoeybi and Mostofa Patwary and Namit Dhameja and Nave Assaf and Negar Habibi and Nidhi Bhatia and Nikki Pope and Nima Tajbakhsh and Nirmal Kumar Juluru and Oleg Rybakov and Oleksii Hrinchuk and Oleksii Kuchaiev and Oluwatobi Olabiyi and Pablo Ribalta and Padmavathy Subramanian and Parth Chadha and Pavlo Molchanov and Peter Dykas and Peter Jin and Piotr Bialecki and Piotr Januszewski and Pradeep Thalasta and Prashant Gaikwad and Prasoon Varshney and Pritam Gundecha and Przemek Tredak and Rabeeh Karimi Mahabadi and Rajen Patel and Ran El-Yaniv and Ranjit Rajan and Ria Cheruvu and Rima Shahbazyan and Ritika Borkar and Ritu Gala and Roger Waleffe and Ruoxi Zhang and Russell J. Hewett and Ryan Prenger and Sahil Jain and Samuel Kriman and Sanjeev Satheesh and Saori Kaji and Sarah Yurick and Saurav Muralidharan and Sean Narenthiran and Seonmyeong Bak and Sepehr Sameni and Seungju Han and Shanmugam Ramasamy and Shaona Ghosh and Sharath Turuvekere Sreenivas and Shelby Thomas and Shizhe Diao and Shreya Gopal and Shrimai Prabhumoye and Shubham Toshniwal and Shuoyang Ding and Siddharth Singh and Siddhartha Jain and Somshubra Majumdar and Soumye Singhal and Stefania Alborghetti and Syeda Nahida Akter and Terry Kong and Tim Moon and Tomasz Hliwiak and Tomer Asida and Tony Wang and Tugrul Konuk and Twinkle Vashishth and Tyler Poon and Udi Karpas and Vahid Noroozi and Venkat Srinivasan and Vijay Korthikanti and Vikram Fugro and Vineeth Kalluru and Vitaly Kurin and Vitaly Lavrukhin and Wasi Uddin Ahmad and Wei Du and Wonmin Byeon and Ximing Lu and Xin Dong and Yashaswi Karnati and Yejin Choi and Yian Zhang and Ying Lin and Yonggan Fu and Yoshi Suhara and Zhen Dong and Zhiyu Li and Zhongbo Zhu and Zijia Chen},
      year={2025},
      eprint={2508.14444},
      archivePrefix={arXiv},
      primaryClass={cs.CL},
      url={https://arxiv.org/abs/2508.14444}, 
}

@misc{toshniwal2024openmathinstruct2acceleratingaimath,
      title={OpenMathInstruct-2: Accelerating AI for Math with Massive Open-Source Instruction Data}, 
      author={Shubham Toshniwal and Wei Du and Ivan Moshkov and Branislav Kisacanin and Alexan Ayrapetyan and Igor Gitman},
      year={2024},
      eprint={2410.01560},
      archivePrefix={arXiv},
      primaryClass={cs.CL},
      url={https://arxiv.org/abs/2410.01560}, 
}

@misc{majumdar2025geneticinstructscalingsynthetic,
      title={Genetic Instruct: Scaling up Synthetic Generation of Coding Instructions for Large Language Models}, 
      author={Somshubra Majumdar and Vahid Noroozi and Mehrzad Samadi and Sean Narenthiran and Aleksander Ficek and Wasi Uddin Ahmad and Jocelyn Huang and Jagadeesh Balam and Boris Ginsburg},
      year={2025},
      eprint={2407.21077},
      archivePrefix={arXiv},
      primaryClass={cs.CL},
      url={https://arxiv.org/abs/2407.21077}, 
}

@misc{olmocr2,
      title={olmOCR 2: Unit Test Rewards for Document OCR}, 
      author={Jake Poznanski and Luca Soldaini and Kyle Lo},
      year={2025},
      eprint={2510.19817},
      archivePrefix={arXiv},
      primaryClass={cs.CV},
      url={https://arxiv.org/abs/2510.19817}, 
}

@inproceedings{
akter2025mind,
title={{MIND}: Math Informed syNthetic Dialogues for Pretraining {LLM}s},
author={Syeda Nahida Akter and Shrimai Prabhumoye and John Kamalu and Sanjeev Satheesh and Eric Nyberg and Mostofa Patwary and Mohammad Shoeybi and Bryan Catanzaro},
booktitle={The Thirteenth International Conference on Learning Representations},
year={2025},
url={https://openreview.net/forum?id=TuOTSAiHDn}
}

@article{hendryckstest2021,
  title={Measuring Massive Multitask Language Understanding},
  author={Dan Hendrycks and Collin Burns and Steven Basart and Andy Zou and Mantas Mazeika and Dawn Song and Jacob Steinhardt},
  journal={Proceedings of the International Conference on Learning Representations (ICLR)},
  year={2021}
}

@article{wang2024mmlu,
  title={Mmlu-pro: A more robust and challenging multi-task language understanding benchmark},
  author={Wang, Yubo and Ma, Xueguang and Zhang, Ge and Ni, Yuansheng and Chandra, Abhranil and Guo, Shiguang and Ren, Weiming and Arulraj, Aaran and He, Xuan and Jiang, Ziyan and others},
  journal={arXiv preprint arXiv:2406.01574},
  year={2024}
}

@article{su2024nemotron,
  title={Nemotron-CC: Transforming Common Crawl into a refined long-horizon pretraining dataset},
  author={Su, Dan and Kong, Kezhi and Lin, Ying and Jennings, Joseph and Norick, Brandon and Kliegl, Markus and Patwary, Mostofa and Shoeybi, Mohammad and Catanzaro, Bryan},
  journal={arXiv preprint arXiv:2412.02595},
  year={2024}
}

@article{mahabadi2025nemotron,
  title={Nemotron-cc-math: A 133 billion-token-scale high quality math pretraining dataset},
  author={Mahabadi, Rabeeh Karimi and Satheesh, Sanjeev and Prabhumoye, Shrimai and Patwary, Mostofa and Shoeybi, Mohammad and Catanzaro, Bryan},
  journal={arXiv preprint arXiv:2508.15096},
  year={2025}
}

@misc{bercovich2025llamanemotronefficientreasoningmodels,
      title={Llama-Nemotron: Efficient Reasoning Models}, 
      author={Akhiad Bercovich and Itay Levy and Izik Golan and Mohammad Dabbah and Ran El-Yaniv and Omri Puny and Ido Galil and Zach Moshe and Tomer Ronen and Najeeb Nabwani and Ido Shahaf and Oren Tropp and Ehud Karpas and Ran Zilberstein and Jiaqi Zeng and Soumye Singhal and Alexander Bukharin and Yian Zhang and Tugrul Konuk and Gerald Shen and Ameya Sunil Mahabaleshwarkar and Bilal Kartal and Yoshi Suhara and Olivier Delalleau and Zijia Chen and Zhilin Wang and David Mosallanezhad and Adi Renduchintala and Haifeng Qian and Dima Rekesh and Fei Jia and Somshubra Majumdar and Vahid Noroozi and Wasi Uddin Ahmad and Sean Narenthiran and Aleksander Ficek and Mehrzad Samadi and Jocelyn Huang and Siddhartha Jain and Igor Gitman and Ivan Moshkov and Wei Du and Shubham Toshniwal and George Armstrong and Branislav Kisacanin and Matvei Novikov and Daria Gitman and Evelina Bakhturina and Jane Polak Scowcroft and John Kamalu and Dan Su and Kezhi Kong and Markus Kliegl and Rabeeh Karimi and Ying Lin and Sanjeev Satheesh and Jupinder Parmar and Pritam Gundecha and Brandon Norick and Joseph Jennings and Shrimai Prabhumoye and Syeda Nahida Akter and Mostofa Patwary and Abhinav Khattar and Deepak Narayanan and Roger Waleffe and Jimmy Zhang and Bor-Yiing Su and Guyue Huang and Terry Kong and Parth Chadha and Sahil Jain and Christine Harvey and Elad Segal and Jining Huang and Sergey Kashirsky and Robert McQueen and Izzy Putterman and George Lam and Arun Venkatesan and Sherry Wu and Vinh Nguyen and Manoj Kilaru and Andrew Wang and Anna Warno and Abhilash Somasamudramath and Sandip Bhaskar and Maka Dong and Nave Assaf and Shahar Mor and Omer Ullman Argov and Scot Junkin and Oleksandr Romanenko and Pedro Larroy and Monika Katariya and Marco Rovinelli and Viji Balas and Nicholas Edelman and Anahita Bhiwandiwalla and Muthu Subramaniam and Smita Ithape and Karthik Ramamoorthy and Yuting Wu and Suguna Varshini Velury and Omri Almog and Joyjit Daw and Denys Fridman and Erick Galinkin and Michael Evans and Katherine Luna and Leon Derczynski and Nikki Pope and Eileen Long and Seth Schneider and Guillermo Siman and Tomasz Grzegorzek and Pablo Ribalta and Monika Katariya and Joey Conway and Trisha Saar and Ann Guan and Krzysztof Pawelec and Shyamala Prayaga and Oleksii Kuchaiev and Boris Ginsburg and Oluwatobi Olabiyi and Kari Briski and Jonathan Cohen and Bryan Catanzaro and Jonah Alben and Yonatan Geifman and Eric Chung and Chris Alexiuk},
      year={2025},
      eprint={2505.00949},
      archivePrefix={arXiv},
      primaryClass={cs.CL},
      url={https://arxiv.org/abs/2505.00949}, 
}

@misc{huang2025opencoderopencookbooktoptier,
      title={OpenCoder: The Open Cookbook for Top-Tier Code Large Language Models}, 
      author={Siming Huang and Tianhao Cheng and J. K. Liu and Jiaran Hao and Liuyihan Song and Yang Xu and J. Yang and Jiaheng Liu and Chenchen Zhang and Linzheng Chai and Ruifeng Yuan and Zhaoxiang Zhang and Jie Fu and Qian Liu and Ge Zhang and Zili Wang and Yuan Qi and Yinghui Xu and Wei Chu},
      year={2025},
      eprint={2411.04905},
      archivePrefix={arXiv},
      primaryClass={cs.CL},
      url={https://arxiv.org/abs/2411.04905}, 
}

@misc{txt360data2024,
      title={TxT360: A Top-Quality LLM Pre-training Dataset Requires the Perfect Blend}, 
      author={Liping Tang and Nikhil Ranjan and Omkar Pangarkar and Xuezhi Liang and Zhen Wang and Li An and Bhaskar Rao and Linghao Jin and Huijuan Wang and Zhoujun Cheng and Suqi Sun and Cun Mu and Victor Miller and Xuezhe Ma and Yue Peng and Zhengzhong Liu and Eric P. Xing},
      year={2024}
}

@misc{zhong2023agieval,
      title={AGIEval: A Human-Centric Benchmark for Evaluating Foundation Models}, 
      author={Wanjun Zhong and Ruixiang Cui and Yiduo Guo and Yaobo Liang and Shuai Lu and Yanlin Wang and Amin Saied and Weizhu Chen and Nan Duan},
      year={2023},
      eprint={2304.06364},
      archivePrefix={arXiv},
      primaryClass={cs.CL}
}

@inproceedings{zellers2019hellaswag,
    title={HellaSwag: Can a Machine Really Finish Your Sentence?},
    author={Zellers, Rowan and Holtzman, Ari and Bisk, Yonatan and Farhadi, Ali and Choi, Yejin},
    booktitle ={Proceedings of the 57th Annual Meeting of the Association for Computational Linguistics},
    year={2019}
}

@article{joshi2017triviaqa,
  title={Triviaqa: A large scale distantly supervised challenge dataset for reading comprehension},
  author={Joshi, Mandar and Choi, Eunsol and Weld, Daniel S and Zettlemoyer, Luke},
  journal={arXiv preprint arXiv:1705.03551},
  year={2017}
}

@misc{zheng2024race,
      title={Beyond Correctness: Benchmarking Multi-dimensional Code Generation for Large Language Models}, 
      author={Jiasheng Zheng and Boxi Cao and Zhengzhao Ma and Ruotong Pan and Hongyu Lin and Yaojie Lu and Xianpei Han and Le Sun},
      year={2024},
      eprint={2407.11470},
      archivePrefix={arXiv},
      primaryClass={cs.SE},
      url={https://arxiv.org/abs/2407.11470}, 
}

@article{sakaguchi2019winogrande,
    title={WinoGrande: An Adversarial Winograd Schema Challenge at Scale},
    author={Sakaguchi, Keisuke and Bras, Ronan Le and Bhagavatula, Chandra and Choi, Yejin},
    journal={arXiv preprint arXiv:1907.10641},
    year={2019}
}

@inproceedings{OpenBookQA2018,
    title={Can a Suit of Armor Conduct Electricity? A New Dataset for Open Book Question Answering},
    author={Todor Mihaylov and Peter Clark and Tushar Khot and Ashish Sabharwal},
    booktitle={EMNLP},
    year={2018}
}

@inproceedings{bisk2020piqa,
  title={Piqa: Reasoning about physical commonsense in natural language},
  author={Bisk, Yonatan and Zellers, Rowan and Gao, Jianfeng and Choi, Yejin and others},
  booktitle={Proceedings of the AAAI conference on artificial intelligence},
  volume={34.05},
  pages={7432--7439},
  year={2020}
}

@misc{chen2021codex,
  title={Evaluating Large Language Models Trained on Code},
  author={Mark Chen and Jerry Tworek and Heewoo Jun and Qiming Yuan and Henrique Ponde de Oliveira Pinto and Jared Kaplan and Harri Edwards and Yuri Burda and Nicholas Joseph and Greg Brockman and Alex Ray and Raul Puri and Gretchen Krueger and Michael Petrov and Heidy Khlaaf and Girish Sastry and Pamela Mishkin and Brooke Chan and Scott Gray and Nick Ryder and Mikhail Pavlov and Alethea Power and Lukasz Kaiser and Mohammad Bavarian and Clemens Winter and Philippe Tillet and Felipe Petroski Such and Dave Cummings and Matthias Plappert and Fotios Chantzis and Elizabeth Barnes and Ariel Herbert-Voss and William Hebgen Guss and Alex Nichol and Alex Paino and Nikolas Tezak and Jie Tang and Igor Babuschkin and Suchir Balaji and Shantanu Jain and William Saunders and Christopher Hesse and Andrew N. Carr and Jan Leike and Josh Achiam and Vedant Misra and Evan Morikawa and Alec Radford and Matthew Knight and Miles Brundage and Mira Murati and Katie Mayer and Peter Welinder and Bob McGrew and Dario Amodei and Sam McCandlish and Ilya Sutskever and Wojciech Zaremba},
  year={2021},
  eprint={2107.03374},
  archivePrefix={arXiv},
  primaryClass={cs.LG}
}

@inproceedings{evalplus,
  title = {Is Your Code Generated by Chat{GPT} Really Correct? Rigorous Evaluation of Large Language Models for Code Generation},
  author = {Liu, Jiawei and Xia, Chunqiu Steven and Wang, Yuyao and Zhang, Lingming},
  booktitle = {Thirty-seventh Conference on Neural Information Processing Systems},
  year = {2023},
  url = {https://openreview.net/forum?id=1qvx610Cu7},
}

@article{austin2021program,
  title={Program synthesis with large language models},
  author={Austin, Jacob and Odena, Augustus and Nye, Maxwell and Bosma, Maarten and Michalewski, Henryk and Dohan, David and Jiang, Ellen and Cai, Carrie and Terry, Michael and Le, Quoc and others},
  journal={arXiv preprint arXiv:2108.07732},
  year={2021}
}

@article{cobbe2021gsm8k,
  title={Training Verifiers to Solve Math Word Problems},
  author={Cobbe, Karl and Kosaraju, Vineet and Bavarian, Mohammad and Chen, Mark and Jun, Heewoo and Kaiser, Lukasz and Plappert, Matthias and Tworek, Jerry and Hilton, Jacob and Nakano, Reiichiro and Hesse, Christopher and Schulman, John},
  journal={arXiv preprint arXiv:2110.14168},
  year={2021}
}

@article{li2024gsm,
  title={Gsm-plus: A comprehensive benchmark for evaluating the robustness of llms as mathematical problem solvers},
  author={Li, Qintong and Cui, Leyang and Zhao, Xueliang and Kong, Lingpeng and Bi, Wei},
  journal={arXiv preprint arXiv:2402.19255},
  year={2024}
}

@article{hendrycks2021measuring,
  title={Measuring mathematical problem solving with the math dataset},
  author={Hendrycks, Dan and Burns, Collin and Kadavath, Saurav and Arora, Akul and Basart, Steven and Tang, Eric and Song, Dawn and Steinhardt, Jacob},
  journal={arXiv preprint arXiv:2103.03874},
  year={2021}
}

@inproceedings{rein2024gpqa,
  title={Gpqa: A graduate-level google-proof q\&a benchmark},
  author={Rein, David and Hou, Betty Li and Stickland, Asa Cooper and Petty, Jackson and Pang, Richard Yuanzhe and Dirani, Julien and Michael, Julian and Bowman, Samuel R},
  booktitle={First Conference on Language Modeling},
  year={2024}
}

@article{du2025supergpqa,
  title={Supergpqa: Scaling llm evaluation across 285 graduate disciplines},
  author={Du, Xinrun and Yao, Yifan and Ma, Kaijing and Wang, Bingli and Zheng, Tianyu and Zhu, King and Liu, Minghao and Liang, Yiming and Jin, Xiaolong and Wei, Zhenlin and others},
  journal={arXiv preprint arXiv:2502.14739},
  year={2025}
}

@article{zhou2025megamath,
  title={Megamath: Pushing the limits of open math corpora},
  author={Zhou, Fan and Wang, Zengzhi and Ranjan, Nikhil and Cheng, Zhoujun and Tang, Liping and He, Guowei and Liu, Zhengzhong and Xing, Eric P},
  journal={arXiv preprint arXiv:2504.02807},
  year={2025}
}

@misc{OpenAI2025GPT5.2,
  author       = {{OpenAI}},
  title        = {Introducing GPT-5.2},
  year         = {2025},
  month        = dec,
  day          = {11},
  howpublished = {\url{https://openai.com/zh-Hans-CN/index/introducing-gpt-5-2/}},
  note         = {Accessed: 2026-02-07}
}

@article{achiam2023gpt,
  title={Gpt-4 technical report},
  author={Achiam, Josh and Adler, Steven and Agarwal, Sandhini and Ahmad, Lama and Akkaya, Ilge and Aleman, Florencia Leoni and Almeida, Diogo and Altenschmidt, Janko and Altman, Sam and Anadkat, Shyamal and others},
  journal={arXiv preprint arXiv:2303.08774},
  year={2023}
}

@misc{Anthropic2026ClaudeOpus4.6,
  author       = {{Anthropic}},
  title        = {Introducing Claude Opus 4.6},
  year         = {2026},
  month        = feb,
  day          = {5},
  howpublished = {\url{https://www.anthropic.com/news/claude-opus-4-6}},
  note         = {Accessed: 2026-02-07}
}

@misc{Google2025Gemini3,
  author       = {{Google}},
  title        = {A new era of intelligence with Gemini 3},
  year         = {2025},
  month        = nov,
  day          = {18},
  howpublished = {\url{https://blog.google/products-and-platforms/products/gemini/gemini-3/}},
  note         = {Accessed: 2026-02-07}
}

@article{comanici2025gemini,
  title={Gemini 2.5: Pushing the frontier with advanced reasoning, multimodality, long context, and next generation agentic capabilities},
  author={Comanici, Gheorghe and Bieber, Eric and Schaekermann, Mike and Pasupat, Ice and Sachdeva, Noveen and Dhillon, Inderjit and Blistein, Marcel and Ram, Ori and Zhang, Dan and Rosen, Evan and others},
  journal={arXiv preprint arXiv:2507.06261},
  year={2025}
}

@article{dubey2024llama,
  title={The llama 3 herd of models},
  author={Dubey, Abhimanyu and Jauhri, Abhinav and Pandey, Abhinav and Kadian, Abhishek and Al-Dahle, Ahmad and Letman, Aiesha and Mathur, Akhil and Schelten, Alan and Yang, Amy and Fan, Angela and others},
  journal={arXiv e-prints},
  pages={arXiv--2407},
  year={2024}
}

@misc{fan2025megasciencepushingfrontiersposttraining,
      title={MegaScience: Pushing the Frontiers of Post-Training Datasets for Science Reasoning}, 
      author={Run-Ze Fan and Zengzhi Wang and Pengfei Liu},
      year={2025},
      eprint={2507.16812},
      archivePrefix={arXiv},
      primaryClass={cs.CL},
      url={https://arxiv.org/abs/2507.16812}, 
}

@misc{ai2025essentialwebv1024ttokens,
      title={Essential-Web v1.0: 24T tokens of organized web data}, 
      author={Essential AI and : and Andrew Hojel and Michael Pust and Tim Romanski and Yash Vanjani and Ritvik Kapila and Mohit Parmar and Adarsh Chaluvaraju and Alok Tripathy and Anil Thomas and Ashish Tanwer and Darsh J Shah and Ishaan Shah and Karl Stratos and Khoi Nguyen and Kurt Smith and Michael Callahan and Peter Rushton and Philip Monk and Platon Mazarakis and Saad Jamal and Saurabh Srivastava and Somanshu Singla and Ashish Vaswani},
      year={2025},
      eprint={2506.14111},
      archivePrefix={arXiv},
      primaryClass={cs.CL},
      url={https://arxiv.org/abs/2506.14111}, 
}

@article{guo2025deepseek,
  title={Deepseek-r1: Incentivizing reasoning capability in llms via reinforcement learning},
  author={Guo, Daya and Yang, Dejian and Zhang, Haowei and Song, Junxiao and Zhang, Ruoyu and Xu, Runxin and Zhu, Qihao and Ma, Shirong and Wang, Peiyi and Bi, Xiao and others},
  journal={arXiv preprint arXiv:2501.12948},
  year={2025}
}

@article{olmo20242,
  title={2 OLMo 2 Furious},
  author={OLMo, Team and Walsh, Pete and Soldaini, Luca and Groeneveld, Dirk and Lo, Kyle and Arora, Shane and Bhagia, Akshita and Gu, Yuling and Huang, Shengyi and Jordan, Matt and others},
  journal={arXiv preprint arXiv:2501.00656},
  year={2024}
}

@article{olmo2025olmo,
  title={Olmo 3},
  author={Olmo, Team and Ettinger, Allyson and Bertsch, Amanda and Kuehl, Bailey and Graham, David and Heineman, David and Groeneveld, Dirk and Brahman, Faeze and Timbers, Finbarr and Ivison, Hamish and others},
  journal={arXiv preprint arXiv:2512.13961},
  year={2025}
}

@misc{qwen3.5,
    title  = {{Qwen3.5}: Towards Native Multimodal Agents},
    author = {{Qwen Team}},
    month  = {February},
    year   = {2026},
    url    = {https://qwen.ai/blog?id=qwen3.5}
}

@article{qwen3,
    title={Qwen3 Technical Report}, 
    author={An Yang and Anfeng Li and Baosong Yang and Beichen Zhang and Binyuan Hui and Bo Zheng and Bowen Yu and Chang Gao and Chengen Huang and Chenxu Lv and Chujie Zheng and Dayiheng Liu and Fan Zhou and Fei Huang and Feng Hu and Hao Ge and Haoran Wei and Huan Lin and Jialong Tang and Jian Yang and Jianhong Tu and Jianwei Zhang and Jianxin Yang and Jiaxi Yang and Jing Zhou and Jingren Zhou and Junyang Lin and Kai Dang and Keqin Bao and Kexin Yang and Le Yu and Lianghao Deng and Mei Li and Mingfeng Xue and Mingze Li and Pei Zhang and Peng Wang and Qin Zhu and Rui Men and Ruize Gao and Shixuan Liu and Shuang Luo and Tianhao Li and Tianyi Tang and Wenbiao Yin and Xingzhang Ren and Xinyu Wang and Xinyu Zhang and Xuancheng Ren and Yang Fan and Yang Su and Yichang Zhang and Yinger Zhang and Yu Wan and Yuqiong Liu and Zekun Wang and Zeyu Cui and Zhenru Zhang and Zhipeng Zhou and Zihan Qiu},
    journal = {arXiv preprint arXiv:2505.09388},
    year={2025}
}

@article{qwen2.5,
    title   = {Qwen2.5 Technical Report}, 
    author  = {An Yang and Baosong Yang and Beichen Zhang and Binyuan Hui and Bo Zheng and Bowen Yu and Chengyuan Li and Dayiheng Liu and Fei Huang and Haoran Wei and Huan Lin and Jian Yang and Jianhong Tu and Jianwei Zhang and Jianxin Yang and Jiaxi Yang and Jingren Zhou and Junyang Lin and Kai Dang and Keming Lu and Keqin Bao and Kexin Yang and Le Yu and Mei Li and Mingfeng Xue and Pei Zhang and Qin Zhu and Rui Men and Runji Lin and Tianhao Li and Tingyu Xia and Xingzhang Ren and Xuancheng Ren and Yang Fan and Yang Su and Yichang Zhang and Yu Wan and Yuqiong Liu and Zeyu Cui and Zhenru Zhang and Zihan Qiu},
    journal = {arXiv preprint arXiv:2412.15115},
    year    = {2024}
}

@article{qwen2,
    title   = {Qwen2 Technical Report}, 
    author  = {An Yang and Baosong Yang and Binyuan Hui and Bo Zheng and Bowen Yu and Chang Zhou and Chengpeng Li and Chengyuan Li and Dayiheng Liu and Fei Huang and Guanting Dong and Haoran Wei and Huan Lin and Jialong Tang and Jialin Wang and Jian Yang and Jianhong Tu and Jianwei Zhang and Jianxin Ma and Jin Xu and Jingren Zhou and Jinze Bai and Jinzheng He and Junyang Lin and Kai Dang and Keming Lu and Keqin Chen and Kexin Yang and Mei Li and Mingfeng Xue and Na Ni and Pei Zhang and Peng Wang and Ru Peng and Rui Men and Ruize Gao and Runji Lin and Shijie Wang and Shuai Bai and Sinan Tan and Tianhang Zhu and Tianhao Li and Tianyu Liu and Wenbin Ge and Xiaodong Deng and Xiaohuan Zhou and Xingzhang Ren and Xinyu Zhang and Xipin Wei and Xuancheng Ren and Yang Fan and Yang Yao and Yichang Zhang and Yu Wan and Yunfei Chu and Yuqiong Liu and Zeyu Cui and Zhenru Zhang and Zhihao Fan},
    journal = {arXiv preprint arXiv:2407.10671},
    year    = {2024}
}

@article{hu2024yulan,
  title={Yulan-mini: An open data-efficient language model},
  author={Hu, Yiwen and Song, Huatong and Deng, Jia and Wang, Jiapeng and Chen, Jie and Zhou, Kun and Zhu, Yutao and Jiang, Jinhao and Dong, Zican and Zhao, Wayne Xin and others},
  journal={arXiv preprint arXiv:2412.17743},
  year={2024}
}

@article{ainslie2023gqa,
  title={Gqa: Training generalized multi-query transformer models from multi-head checkpoints},
  author={Ainslie, Joshua and Lee-Thorp, James and De Jong, Michiel and Zemlyanskiy, Yury and Lebr{\'o}n, Federico and Sanghai, Sumit},
  journal={arXiv preprint arXiv:2305.13245},
  year={2023}
}

@article{shazeer2020glu,
  title={Glu variants improve transformer},
  author={Shazeer, Noam},
  journal={arXiv preprint arXiv:2002.05202},
  year={2020}
}

@article{zhang2019root,
  title={Root mean square layer normalization},
  author={Zhang, Biao and Sennrich, Rico},
  journal={Advances in neural information processing systems},
  volume={32},
  year={2019}
}

@article{su2024roformer,
  title={Roformer: Enhanced transformer with rotary position embedding},
  author={Su, Jianlin and Ahmed, Murtadha and Lu, Yu and Pan, Shengfeng and Bo, Wen and Liu, Yunfeng},
  journal={Neurocomputing},
  volume={568},
  pages={127063},
  year={2024},
  publisher={Elsevier}
}

@misc{eval-harness,
  author       = {Gao, Leo and Tow, Jonathan and Abbasi, Baber and Biderman, Stella and Black, Sid and DiPofi, Anthony and Foster, Charles and Golding, Laurence and Hsu, Jeffrey and Le Noac'h, Alain and Li, Haonan and McDonell, Kyle and Muennighoff, Niklas and Ociepa, Chris and Phang, Jason and Reynolds, Laria and Schoelkopf, Hailey and Skowron, Aviya and Sutawika, Lintang and Tang, Eric and Thite, Anish and Wang, Ben and Wang, Kevin and Zou, Andy},
  title        = {The Language Model Evaluation Harness},
  month        = 07,
  year         = 2024,
  publisher    = {Zenodo},
  version      = {v0.4.3},
  doi          = {10.5281/zenodo.12608602},
  url          = {https://zenodo.org/records/12608602}
}

@article{peng2023instruction,
  title={Instruction tuning with gpt-4},
  author={Peng, Baolin and Li, Chunyuan and He, Pengcheng and Galley, Michel and Gao, Jianfeng},
  journal={arXiv preprint arXiv:2304.03277},
  year={2023}
}

@article{ouyang2022training,
  title={Training language models to follow instructions with human feedback},
  author={Ouyang, Long and Wu, Jeffrey and Jiang, Xu and Almeida, Diogo and Wainwright, Carroll and Mishkin, Pamela and Zhang, Chong and Agarwal, Sandhini and Slama, Katarina and Ray, Alex and others},
  journal={Advances in neural information processing systems},
  volume={35},
  pages={27730--27744},
  year={2022}
}

@article{jimenez2023swe,
  title={Swe-bench: Can language models resolve real-world github issues?},
  author={Jimenez, Carlos E and Yang, John and Wettig, Alexander and Yao, Shunyu and Pei, Kexin and Press, Ofir and Narasimhan, Karthik},
  journal={arXiv preprint arXiv:2310.06770},
  year={2023}
}

@article{liu2023pre,
  title={Pre-train, prompt, and predict: A systematic survey of prompting methods in natural language processing},
  author={Liu, Pengfei and Yuan, Weizhe and Fu, Jinlan and Jiang, Zhengbao and Hayashi, Hiroaki and Neubig, Graham},
  journal={ACM computing surveys},
  volume={55},
  number={9},
  pages={1--35},
  year={2023},
  publisher={ACM New York, NY}
}

@article{liu2025alphago,
  title={Alphago moment for model architecture discovery},
  author={Liu, Yixiu and Nan, Yang and Xu, Weixian and Hu, Xiangkun and Ye, Lyumanshan and Qin, Zhen and Liu, Pengfei},
  journal={arXiv preprint arXiv:2507.18074},
  year={2025}
}

@article{akter2025front,
  title={Front-loading reasoning: The synergy between pretraining and post-training data},
  author={Akter, Syeda Nahida and Prabhumoye, Shrimai and Nyberg, Eric and Patwary, Mostofa and Shoeybi, Mohammad and Choi, Yejin and Catanzaro, Bryan},
  journal={arXiv preprint arXiv:2510.03264},
  year={2025}
}

@inproceedings{ovadia2024fine,
  title={Fine-tuning or retrieval? comparing knowledge injection in llms},
  author={Ovadia, Oded and Brief, Menachem and Mishaeli, Moshik and Elisha, Oren},
  booktitle={Proceedings of the 2024 conference on empirical methods in natural language processing},
  pages={237--250},
  year={2024}
}

@article{pletenev2025much,
  title={How Much Knowledge Can You Pack into a LoRA Adapter without Harming LLM?},
  author={Pletenev, Sergey and Marina, Maria and Moskovskiy, Daniil and Konovalov, Vasily and Braslavski, Pavel and Panchenko, Alexander and Salnikov, Mikhail},
  journal={arXiv preprint arXiv:2502.14502},
  year={2025}
}

@article{jaech2024openai,
  title={Openai o1 system card},
  author={Jaech, Aaron and Kalai, Adam and Lerer, Adam and Richardson, Adam and El-Kishky, Ahmed and Low, Aiden and Helyar, Alec and Madry, Aleksander and Beutel, Alex and Carney, Alex and others},
  journal={arXiv preprint arXiv:2412.16720},
  year={2024}
}

@article{qin2024o1,
  title={O1 Replication Journey: A Strategic Progress Report--Part 1},
  author={Qin, Yiwei and Li, Xuefeng and Zou, Haoyang and Liu, Yixiu and Xia, Shijie and Huang, Zhen and Ye, Yixin and Yuan, Weizhe and Liu, Hector and Li, Yuanzhi and others},
  journal={arXiv preprint arXiv:2410.18982},
  year={2024}
}

@article{huang2024o1,
  title={O1 Replication Journey--Part 2: Surpassing O1-preview through Simple Distillation, Big Progress or Bitter Lesson?},
  author={Huang, Zhen and Zou, Haoyang and Li, Xuefeng and Liu, Yixiu and Zheng, Yuxiang and Chern, Ethan and Xia, Shijie and Qin, Yiwei and Yuan, Weizhe and Liu, Pengfei},
  journal={arXiv preprint arXiv:2411.16489},
  year={2024}
}

@article{ye2025limo,
  title={Limo: Less is more for reasoning},
  author={Ye, Yixin and Huang, Zhen and Xiao, Yang and Chern, Ethan and Xia, Shijie and Liu, Pengfei},
  journal={arXiv preprint arXiv:2502.03387},
  year={2025}
}

@article{xia2025generative,
  title={Generative ai act ii: Test time scaling drives cognition engineering},
  author={Xia, Shijie and Qin, Yiwei and Li, Xuefeng and Ma, Yan and Fan, Run-Ze and Chern, Steffi and Zou, Haoyang and Zhou, Fan and Hu, Xiangkun and Jin, Jiahe and others},
  journal={arXiv preprint arXiv:2504.13828},
  year={2025}
}

@misc{lambert2025atom,
  title={The ATOM Project},
  author={Lambert, Nathan},
  year={2025},
  month={August},
  day={4},
  url={https://atomproject.ai}
}
\clearpage
\appendix

\section{Evaluation Details}
\label{appendix:benchmarks}
\subsection{Benchmark Descriptions}

This appendix provides detailed descriptions of all 19 benchmarks used in our evaluation, organized by capability domain.

\subsubsection{General Knowledge and Reasoning Benchmarks}

\paragraph{MMLU (Massive Multitask Language Understanding).}
MMLU is a comprehensive multiple-choice benchmark covering 57 subjects spanning STEM, humanities, social sciences, and other areas. The dataset tests a model's broad knowledge across diverse domains including elementary mathematics, US history, computer science, law, and more. Questions range from elementary to professional level, providing a thorough assessment of general knowledge and reasoning capabilities.

\paragraph{MMLU-Pro.}
MMLU-Pro is an enhanced variant of MMLU designed to address the plateau in model performance on the original benchmark. It extends MMLU by: (1) integrating more challenging, reasoning-focused questions; (2) expanding the choice set from four to ten options; and (3) eliminating trivial and noisy questions. MMLU-Pro causes a significant 16-33\% accuracy drop compared to MMLU while demonstrating greater stability under varying prompts (2\% sensitivity vs. 4-5\% in MMLU). The benchmark shows that models utilizing Chain of Thought (CoT) reasoning achieve better performance compared to direct answering.

\paragraph{AGIEval (EN).}
AGIEval is a human-centric benchmark specifically designed to evaluate foundation models' general abilities in tasks pertinent to human cognition and problem-solving. It is derived from 20 official, public, and high-standard admission and qualification exams, including Chinese College Entrance Exam (Gaokao), American SAT, law school admission tests, math competitions, lawyer qualification tests, and national civil service exams. We evaluate on the English subset (AGIEval-EN) which includes AQUA-RAT, Gaokao English, LogiQA-EN, LSAT (Analytical Reasoning, Logical Reasoning, Reading Comprehension), SAT (English, Math), and MATH problems.

\paragraph{HellaSwag.}
HellaSwag is a commonsense natural language inference dataset designed to be trivial for humans (>95\% accuracy) but challenging for state-of-the-art models. Given an event description (e.g., ``A woman sits at a piano''), the task is to select the most likely followup from multiple options. The dataset was constructed via Adversarial Filtering (AF), where discriminators iteratively select adversarial machine-generated wrong answers. The key insight is scaling up the length and complexity of examples to a ``Goldilocks'' zone where generated text is ridiculous to humans yet often misclassified by models.

\paragraph{TriviaQA.}
TriviaQA is a large-scale reading comprehension dataset containing over 650K question-answer-evidence triples. It includes 95K question-answer pairs authored by trivia enthusiasts and independently gathered evidence documents (six per question on average) that provide high-quality distant supervision for answering the questions. The dataset tests both reading comprehension and the ability to locate relevant information across multiple documents.

\paragraph{RACE (ReAding Comprehension from Examinations).}
RACE is a large-scale reading comprehension dataset with more than 28,000 passages and nearly 100,000 questions. The dataset is collected from English examinations in China designed for middle school and high school students. Questions require understanding passage content and reasoning about implicit information, making it a challenging test of reading comprehension capabilities.

\paragraph{WinoGrande.}
WinoGrande is a collection of 44K problems inspired by the Winograd Schema Challenge, designed to test commonsense reasoning. Formulated as a fill-in-a-blank task with binary options, the goal is to choose the correct option for a given sentence requiring commonsense reasoning. The dataset was adjusted to improve scale and robustness against dataset-specific bias.

\paragraph{OpenBookQA.}
OpenBookQA is a question-answering dataset modeled after open book exams for assessing human understanding of science. It consists of 5,957 multiple-choice elementary-level science questions (4,957 train, 500 dev, 500 test) that probe understanding of 1,326 core science facts and their application to novel situations. Answering requires both the provided ``book'' of facts and additional broad common knowledge not contained in the book.

\paragraph{PIQA (Physical Interaction: Question Answering).}
PIQA is a physical commonsense reasoning benchmark designed to investigate the physical knowledge of language models. The dataset tests reasoning about physical interactions and commonsense understanding of how the physical world works. Questions are designed to assess to what extent models are actually learning about the physical world versus merely pattern matching.

\subsubsection{Code Generation Benchmarks}

\paragraph{HumanEval.}
HumanEval is a code generation benchmark introduced to measure functional correctness for synthesizing Python programs from docstrings. It contains 164 handwritten programming problems with function signatures, docstrings, reference implementations, and multiple unit tests. The benchmark evaluates pass@k metrics, measuring the probability that at least one of k generated samples passes all unit tests. HumanEval has become a standard benchmark for evaluating code generation capabilities of language models.

\paragraph{EvalPlus.}
EvalPlus extends HumanEval with more comprehensive test suites, providing stricter evaluation than the original benchmark. It augments HumanEval problems with additional test cases that better cover edge cases and corner conditions, making it more difficult to achieve high scores through superficial pattern matching.

\paragraph{MBPP (Mostly Basic Programming Problems).}
MBPP is a benchmark for program synthesis containing 974 short Python programming tasks designed to be solvable by entry-level programmers. Each problem consists of a natural language description, a reference solution, and three automated test cases. The dataset is designed to measure the ability of models to synthesize short Python programs from natural language descriptions, testing basic programming competence.

\subsubsection{Mathematics and STEM Reasoning Benchmarks}

\paragraph{GSM8K (Grade School Math 8K).}
GSM8K is a dataset of 8.5K high-quality, linguistically diverse grade school math word problems requiring multi-step mathematical reasoning. Despite the conceptual simplicity of this problem distribution, even large transformer models struggle to achieve high test performance. The dataset tests whether models can perform the sequential reasoning steps needed to solve elementary mathematics problems. Each question has a natural language solution that demonstrates the reasoning steps.

\paragraph{GSM-Plus.}
GSM-Plus is a comprehensive benchmark for evaluating the robustness of LLMs as mathematical problem solvers. It extends GSM8K with various mathematical perturbations to test whether models truly understand and apply mathematical knowledge or merely rely on shortcuts. When math questions are slightly changed (new statements added or question targets altered), LLMs can make mistakes even on problems they solved in the original GSM8K, revealing brittleness in their mathematical reasoning.

\paragraph{MATH.}
MATH is a dataset of 12,500 challenging competition mathematics problems spanning algebra, counting and probability, geometry, intermediate algebra, number theory, prealgebra, and precalculus. Each problem has a full step-by-step solution demonstrating the derivation of the answer. The competition-level difficulty makes MATH one of the most challenging mathematical reasoning benchmarks, requiring sophisticated problem-solving strategies beyond pattern matching.

\paragraph{GPQA (Graduate-Level Google-Proof Q\&A).}
GPQA is a challenging dataset of 448 multiple-choice questions written by domain experts in biology, physics, and chemistry. The questions are designed to be ``Google-proof'': experts with PhDs reach 65\% accuracy (74\% when discounting clear mistakes), while highly skilled non-expert validators only reach 3  4\% accuracy despite spending over 30 minutes with unrestricted web access. State-of-the-art AI systems also struggle, with GPT-4-based baselines achieving around 39\% accuracy. GPQA-Main refers to the main variant of this benchmark.

\paragraph{SuperGPQA.}
SuperGPQA is an extended variant of GPQA with additional challenging questions in biology, physics, and chemistry. It maintains the graduate-level difficulty and Google-proof property of the original GPQA while expanding coverage of advanced scientific topics.

\paragraph{MMLU-STEM.}
This is the STEM-focused subset of MMLU, including subjects such as abstract algebra, astronomy, college biology, college chemistry, college computer science, college mathematics, college physics, computer security, elementary mathematics, high school biology, high school chemistry, high school computer science, high school mathematics, high school physics, high school statistics, machine learning, and electrical engineering.

\paragraph{MMLU-Pro-STEM.}
This is the STEM subset of MMLU-Pro, maintaining the increased difficulty and expanded choice set (10 options) of MMLU-Pro while focusing specifically on STEM disciplines. It provides an even more challenging test of scientific and mathematical knowledge than MMLU-STEM.

\section{Training Implementation Decisions: LR Decay and QA Masking}
\label{sec:app-lr-masking}

Beyond data mixture design, two practical implementation choices can meaningfully influence training outcomes yet are rarely studied in pretraining contexts: \textbf{learning rate decay scheduling} and \textbf{QA question masking policy}. LR decay schedules determine how aggressively the model consolidates knowledge in later training steps, while masking policy governs whether the model is trained to predict question tokens or only answer tokens. Although both choices may seem like minor engineering details, their interaction with the data format and training stage can lead to non-trivial performance differences. We systematically investigate both decisions to provide principled guidance for Stage~2 configurations.

\paragraph{Learning rate decay.}
We compare constant LR against cosine decay (decreasing from 3e-5 to 3e-6) for the 70\% QA Stage~2-2 configuration (Table~\ref{tab:lr-decay}). Cosine decay yields consistent improvements in general knowledge and code generation, while science reasoning remains largely unchanged. The benefits align with a ``capability building then refinement'' pattern: aggressive learning in Stage~2-1 establishes broad capabilities, and a gradual LR reduction in Stage~2-2 consolidates and stabilizes them. The overall improvement is modest but reliable, making cosine decay a low-cost enhancement worth applying in multi-substage configurations.

\begin{table}[t]
  \centering
  \caption{Impact of learning rate decay strategy on Stage~2-2 performance. Constant LR maintains 3e-5 throughout, while cosine decay gradually reduces LR from 3e-5 to 3e-6.}
  \label{tab:lr-decay}
  \small
  \setlength{\tabcolsep}{3pt}
  \begin{tabular}{l c c c c}
    \toprule
    Configuration & Avg General & Avg Code & Avg Science & Overall Avg \\
    \midrule
    Constant LR & 52.06 & 52.32 & 45.44 & 49.66 \\
    Cosine decay & \textbf{52.49} & \textbf{53.31} & \textbf{45.39} & \textbf{50.00} \\
    \midrule
    $\Delta$ (decay - constant) & +0.43 & +0.99 & -0.05 & +0.34 \\
    \bottomrule
  \end{tabular}
\end{table}

\begin{figure}[t]
  \centering
  \includegraphics[width=0.999\linewidth]{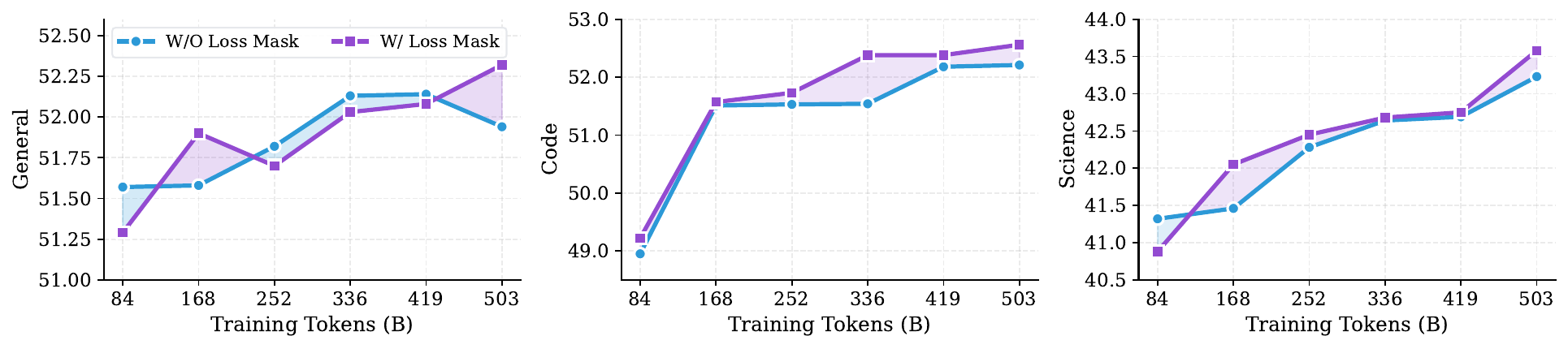}
  \caption{Impact of QA masking policy on three capability dimensions across training steps in Stage~2-2. The comparison between masked and unmasked configurations shows marginal improvements, with consistent but modest gains across all capability clusters.}
  \label{fig:qa-masking}
\end{figure}

\paragraph{QA data masking policy.}
\label{sec:qa-masking}
We compare masking the question portion of QA pairs (only answer tokens contribute to loss, following SFT convention) against treating QA as continuous text (all tokens contribute). Using the 70\% QA Stage~2-2 configuration with high-quality, diverse question data, masking achieves 49.14 overall at 30k steps versus 48.77 without masking (+0.37). This marginal gain is consistent across general (+0.38), code (+0.35), and science (+0.35) domains, and training dynamics remain stable under both strategies. In contrast to SFT---where masking is essential to prevent shortcut learning---pretraining's broader knowledge acquisition objective means that models benefit from learning \emph{both} question understanding and answer generation when questions are sufficiently diverse. Practically, masking incurs non-trivial engineering overhead (additional mask storage or runtime computation) for a modest return; the key takeaway is that \textbf{question quality and diversity matter more than masking strategy}. Figure~\ref{fig:qa-masking} illustrates the comparison across training checkpoints.

\section{Prompts for Dataset Construction}
\label{dataset construction prompt}
\begin{tcolorbox}[
colback=blue!5!white,
colframe=blue!30!gray,
  boxrule=2pt,
  fontupper=\small,
  arc=8pt,
  title={\textbf{Rewriting Prompt for Constructing MegaMath-Web-Pro}},
  fonttitle=\normalsize\bfseries,
  enhanced jigsaw,
  breakable
]
\begin{TextBlock}
Task:
- Carefully analyze the provided text to extract key facts, concrete details,
important numbers, and core concepts.
- Remove any irrelevant or noisy information, and reorganize the content into a
logically structured, information-dense, and concise version that is easy to
learn from. Output only the refined text.
- Strive to maintain the original length as much as possible (avoid excessive
shortening).
Text:
<EXAMPLE>
Just output the refined text, no other text.
\end{TextBlock}
\end{tcolorbox}

%[完整的cleaning prompt]
\begin{tcolorbox}[
colback=blue!5!white,
colframe=blue!30!gray,
  fontupper=\small,
  boxrule=2pt,
  arc=8pt,
  title={\textbf{Darwin-Science L4 Processing Prompt}},
  fonttitle=\normalsize\bfseries,
  enhanced jigsaw,
  breakable
]
\begin{TextBlock}
You are an expert document cleaner specialized in identifying and removing unwanted content and correcting OCR errors from various document (mainly academic) chunks.

## Objective:
Clean and standardize OCR text by identifying and removing redundant, erroneous, or unwanted content and correcting obvious OCR errors according to the rules below. Your task is to identify and delete unnecessary content completely, fix technical errors, while preserving all academic value.

## Deletion and Correction Rules:
### Document Structural Deletion
* Remove **table of contents and navigation structures**: Multiple consecutive chapter/section titles listed together without accompanying text content
  - **Preserve content section headings in main text**: such as chapter headings, section titles followed by explanatory text or academic material
* Remove **reference lists completely**: numbered entries with author names, publication titles, and years (e.g., "1. Smith, J. (2020). Title. Journal, 15(3), 123-145.") **[Delete entire list regardless of format]**
* Remove **front matter and back matter**: such as prefaces, acknowledgments, copyright statements, indexes, and other standard book structural elements
  - **Preserve sections with academic value**: such as abstracts, introductions, conclusions that present research background or methodology
* Remove **publication and metadata information**: such as ISBN, publisher information, revision history, version numbers, institutional affiliations, author affiliations, addresses, contact information
* Remove **page headers, page footers, and page numbers**

### Academic Content Deletion
* Remove **pure indexing appendices**: such as glossaries, symbol tables, abbreviation lists, indexes, notations and other purely referential lookup content (entries that only provide definitions without explanations, e.g., "a - alpha coefficient")
  - **Preserve**: appendices with learning value (e.g. mathematical derivations, proofs, technical explanations)
  - **Preserve**: explanatory content that directly supports main text elements (e.g. abbreviation/parameter explanations after tables/formulas/diagrams)
* Remove **image files and placeholders**: such as `<img>` tags, image file paths, image URLs, markdown syntax and image placeholders (e.g. `[Image]`, `[Picture not available]`)
  - **Preserve**: figure/table titles, descriptive text (including content within markdown image formats: ![description](path) → description)
  - **Preserve**: in-text references (e.g., "as shown in Figure 1")

### Invalid and Redundant Content Deletion
* Remove **OCR processing artifacts**: such as garbled text, encoding artifacts, duplicate characters, malformed special characters, OCR messages (`[OCR error]`), file paths, timestamps, version numbers, revision history
* Remove **garbage content**: such as junk information, advertising content, placeholders (e.g. [Insert citation here])
* Remove **duplicate content**: identical paragraphs or sections mainly caused by OCR errors
  - **Exception**: Carefully apply to technical formulas, equations, or specialized notation that may contain subtle but meaningful differences
  - **Exception**: Apply contextual analysis - preserve identical content that serves different semantic purposes or artistic purposes (e.g., poetic refrains, literary repetition)
* Remove **content and navigation markers**: [content missing], [page break], (Continued), and similar placeholder markers
* Remove **URLs and links**: all web addresses, hyperlinks, and link information

### OCR Error Correction
* **Fix text fragmentation**: repair split words, broken sentences, erroneous line breaks and paragraph divisions, missing spaces and punctuation
* **Fix fragmented structured content**: Repair OCR-damaged structured content (e.g. tables, diagrams, formulas) appearing as consecutive lines of isolated words, single characters, or short phrases
  - **Pattern**: Consecutive lines (5+) with 1-3 words/characters each
  - **Action**: Preserve content while indicating structural damage; delete if unrepairable
* **Standardize whitespace and formatting**: clean excessive whitespace, compress blank lines, standardize spacing and indentation
* **Fix character and encoding errors**: correct obvious character errors, spelling issues, and Unicode anomalies
* **Standardize punctuation**: unify quotation marks, dashes, hyphens, and other punctuation
* **Complete truncated words**: only fix obviously incomplete words from clear OCR errors, avoid modifying content at chunk edges
* **Standardize academic formatting**: remove excessive LaTeX commands and unify notation format

## Content Protection Rules:
### Always Preserve Academic and Educational Content
* Preserve **Technical and specialized content**: such as formulas, equations, proofs, symbols, chemical structures, biological sequences and their original format
  - **Preserve exact content**: do not alter variables, coefficients, structures, sequences, or any technical details
* Preserve **In-text references and citations**: such as (Smith, 2020), [15], "see Chapter 2", equation (5), "Figure 2.5", (pp. 3-7)
* Preserve **Table structures**: preserve academic table content, formatting and structural markers (e.g. "|", HTML tags)
  - **Exception**: Does not apply to navigation tables (table of contents, indexes, glossaries) which should be removed
* Preserve **Code blocks and programming examples**: preserve code block markers (```language, ```, etc.) and internal code syntax and structure
* Preserve **Educational content**: such as exercises, questions, answers, solutions, case studies, instructions, user guides
* Preserve **Explanatory content**: such as NOTE boxes, WARNING boxes, tips, author comments, supplementary information, academic footnotes
* Preserve **Chunk boundary content**: incomplete sentences and words at chunk edges due to text segmentation
* Preserve **Literary and humanities content**: including poetry, fiction, drama, creative writing, literary analysis, philosophical texts, and other humanities scholarship with educational value

## Instructions:
- Carefully identify all content matching the deletion rules
- Remove completely any content that should be deleted
- Preserve all valuable academic content by applying protection rules and retaining content that doesn't match deletion rules
- Apply OCR error corrections to fix obvious technical problems
- Ensure text flows naturally after corrections and deletions
- If the entire chunk should be deleted, leave the output tags completely empty
- **Important**: The content inside the <CLEANED_TEXT> tags must be exactly the text after deletion, with no explanations, comments, or additional text inside the tags

## Input:
OCR document chunk:
[CHUNK]

## Output Format:
<CLEANED_TEXT>
[Place the cleaned content here, or leave completely empty if everything should be deleted]
</CLEANED_TEXT>
\end{TextBlock}
\end{tcolorbox}

\begin{tcolorbox}[
colback=blue!5!white,
colframe=blue!30!gray,
  fontupper=\small,
  boxrule=2pt,
  arc=8pt,
  title={\textbf{Darwin-Science L5 Processing Prompt}},
  fonttitle=\normalsize\bfseries,
  enhanced jigsaw,
  breakable
]
\begin{TextBlock}
You are a master science communicator and pedagogical expert. Your mission is to transform the following dense, expert-level text chunk into vibrant, crystal-clear educational material. Imagine you are creating a definitive learning resource for a bright but novice audience. Your goal is not merely to simplify, but to deeply elucidate, making the complex intuitive and the implicit explicit.

Your transformation will be governed by two sets of principles: the Core Mandate (what you must actively do) and the Unbreakable Rules (what you must never violate).

### The Unbreakable Rules: Fidelity and Integrity
This principle is of paramount importance and must be followed without exception to ensure the output is valid.

* **(a) Scientific and Factual Correctness:** Maintain absolute rigor. All data, formulas, definitions, theories, experimental results, and logical arguments must be preserved without altering their meaning or context. Your additions must clarify, not contradict.

* **(b) Structural Integrity:** Preserve the original structure flawlessly. Keep ALL section headers (`##`, `###`), figure/table labels, equation numbers, etc., exactly as they appear, especially at the beginning and end of the chunk.

* **(c) Contextual Limitation and Termination:** You are processing a partial *chunk* of a document. You lack the full context. Therefore, you must work **strictly** within the provided text. Do not invent definitions or reference goals from outside the chunk. **This strict adherence means your output must terminate exactly where the provided chunk terminates.** If the chunk ends abruptly (e.g., at a new section header, in the middle of a sentence, or with a label), your output **must be cut off at that exact same point.** This is the single most critical rule for preventing hallucination and ensuring continuity.

### The Core Mandate: Deep Pedagogical Transformation
This is your primary objective. Be bold and proactive in adding educational value. Your goal is to weave a rich tapestry of understanding.

* **(a) Deconstruct and Narrate the 'Why':** This is your primary mode of explanation. Actively expand on logical leaps. When the text says "it follows that," "clearly," or "trivially," you must step in and meticulously detail the intermediate steps. More importantly, you must articulate the expert's internal monologue. When faced with an equation, a problem, or a logical step, explain the strategy. Ask and answer questions like: "Okay, what's our goal here?" "What's the first thing I should look for when I see an equation like this?" "We're going to use technique X, and here's why it's the right tool for this specific job." Your mission is to reveal the problem-solving journey, making every single connection transparent.

* **(b) From Jargon to Insight:** When you encounter a crucial technical term, or a significant variable within a formula, you must deliberately pause the narrative to explain it. Don't just provide a dry definition. Elucidate its importance: What role does this term or variable play? Why does it matter? Crucially, you must then use simpler language, vivid analogies, or concrete examples to build a strong and intuitive mental model for the reader before you continue with the main explanation. This ensures no reader is left behind due to unfamiliar notation or terminology.

* **(c) Invent Vivid Analogies and Concrete Examples:** Go beyond the text. Where a concept is abstract, create a simple, concrete example to illustrate it. Invent memorable analogies that connect the new information to a learner's existing knowledge (e.g., electron shells as floors in a hotel).

* **(d) Create Contextual Bridges:** Weave a narrative thread by connecting the current idea to the broader field of knowledge. Hint at future applications or link back to more foundational concepts. For instance: "This principle of [X] is a cornerstone of the field and will be essential for understanding [Y] later on."

* **(e) Think Like a Learner:** Proactively identify points of potential confusion. What questions would a curious student ask here? Answer them before they are asked. A great teacher warns students about common mistakes. Where applicable, insert brief, helpful asides that feel like a mentor's margin notes.

* **(f) Prioritize Narrative Flow and Clean Formatting:** When you encounter messy or noisy original LaTeX formatting, convert it to a clean and pristine style (especially for formulas and tables). Above all, strive for a smooth, cohesive, and engaging narrative. Your writing should feel like a continuous, guided tour through the material, not a collection of disconnected facts and callouts. To that end, you must avoid the overuse of overly-structured, point-by-point expressions. Let the main text flow logically and tell a story, adopting the persona of an extremely patient and encouraging teacher.

---
Summary of Principles Above: To sum it up, you must strictly respect the accuracy and structure of the original chunk while doing everything possible to make the rewritten text easier to learn and to lower the reader's cognitive load. Consequently, the rewritten text will typically be more detailed and thus longer than the original.
---

**Crucial Output Instructions:**
1.  **Self-Contained Output:** The refined text must stand on its own. Avoid any meta-commentary or phrases that refer to the original text, such as "the original paper," "the original context," "the original chunk", etc. The goal is to create a seamless, self-contained educational text, not a commentary on another document. 
2.  **Strict Termination:** You **MUST** terminate your output at the **EXACT** same point the provided chunk terminates. Do not write a single character past the end of the original chunk. In particular, if a chunk ends with the start of a new section, subsection, step (e.g., it starts with a heading) or cuts off in the middle of a proof/solution, you must NOT invent or continue writing ANY content that would follow.**

---
*You must output ONLY the refined chunk itself, without any introductory or concluding remarks.*

**Original text:**
\{chunk\}

**Refined text:**
\end{TextBlock}
\end{tcolorbox}

\begin{tcolorbox}[
colback=blue!5!white,
colframe=blue!30!gray,
  boxrule=2pt,
  fontupper=\small,
  arc=8pt,
  title={\textbf{QA Extraction Prompt for Darwin-Science-Book-QA(Biology)}},
  fonttitle=\normalsize\bfseries,
  enhanced jigsaw,
  breakable
]
\begin{TextBlock}
Below is a document extract from biological sciences domain (biology, ecology, medicine, genetics, physiology, etc.).

# Extraction Task
Extract complete, independently solvable biology Q&A pairs following these strict guidelines:

## CRITICAL RULES (Must Follow):
1. **Answer-First Principle**: Only extract a question if its complete answer exists in the document
2. **Zero References Requirement**: 
   - Questions: No references to "the document", "the text", "Figure 1", "Table 2", "above", "below", "the experiment above","as discussed in the extract"etc.
   - Answers: No references to "as in document", "as shown in text", "see the document", "Table 2", etc.
   - Replace ALL references with actual content from the document
3. **Complete Independence**: 
   - Each question must be self-contained with ALL necessary information, as others will not have access to the original text.
   - If a question is incomplete, incorporate missing context to make it fully solvable
   - Multiple questions must be mutually independent (no "as in first Question", "using previous result")
4. **Answer Must Exist**: All answers must be directly found or clearly derivable from the given text - do not extract if uncertain
5. **Value-Based Selection**: Focus on extracting the MOST IMPORTANT and EDUCATIONAL content
6. **No Data-Dependent Conclusions**: Do NOT extract questions whose answers rely heavily on complex data tables or extensive numerical data that cannot be reasonably included in the Q&A format


## For Questions/Problems:
- **Extract explicit question-answer pairs** if they exist (ensure they meet zero references and independence requirements)
- **For Mechanisms/Processes**: Create questions about how biological processes work (e.g., "How does X process occur?", "Describe the mechanism of...")
- **For Structure-Function**: Create questions linking structure to function (e.g., "What is the structure and function of X?")
- **For Phenomena and Causes**: Create questions exploring why biological phenomena occur (e.g., "Why does X happen?", "What causes X?")
- **For Comparisons**: Create questions comparing biological entities (e.g., "What are the differences between X and Y?", "How does X differ from Y?")
- **For Diseases**: Create questions about etiology, symptoms, mechanisms, or treatments (when information is complete)
- **For Experiments**: Extract questions about experimental design, findings, or significance (only if self-contained without heavy data table dependence)
- **For Evolution/Ecology**: Create questions about evolutionary explanations, adaptations, or ecological relationships
- **For Concepts/Definitions**: Ask about them flexibly, but focus on core concepts and ensure complete answers exist
- **Critical**: Each question must be complete and self-contained, as others will not have access to the original text.
- **Do NOT extract**:
  - Questions that require additional data to be provided in order to be answered
  - Questions whose answers are uncertain or incomplete
- If no valid Q&A content can be extracted following above rules, return exactly: [NO QA]

## For Answers:
- **Process**: First locate the complete answer in the document, then organize it clearly
- If the answer has explanation, reorganize it into a clear, well-structured format
- If the answer lacks explanation, add necessary biological reasoning or context
- For final answers that need exact matching (multiple-choice, calculations, fill-in-the-blank, true/false), use $\\boxed{{answer}}$ notation

## Output Format (STRICT):
Use this exact format for parsing:

---BEGIN QA---
Question: <First complete question>
Answer: <First complete answer>
---END QA---

---BEGIN QA---
Question: <Second complete question>
Answer: <Second complete answer>
---END QA---

If no valid Q&A can be extracted, return exactly: [NO QA]

---
Extract:
{chunk}

Now process the extract and return the result in the specified format.

\end{TextBlock}
\end{tcolorbox}

\begin{tcolorbox}[
colback=blue!5!white,
colframe=blue!30!gray,
  boxrule=2pt,
  arc=8pt,
fontupper=\small,
  title={\textbf{QA Extraction Prompt for Darwin-Science-Book-QA(Chemistry})},
  fonttitle=\normalsize\bfseries,
  enhanced jigsaw,
  breakable
]
\begin{TextBlock}

Below is a document extract from chemistry domain.

# Extraction Task
Extract complete, independently solvable chemistry Q&A pairs following these strict guidelines:

## CRITICAL RULES (Must Follow):
1. **Answer-First Principle**: Only extract a question if its complete answer exists in the document
2. **Zero References Requirement**: 
   - Questions: No references to "the document", "the text", "Figure 1", "Scheme 45", "compound 201", "above", "below", etc.
   - Answers: No references to "as in document", "as shown in text", "see Scheme", "compound 201", etc.
   - Replace ALL references with actual content (e.g., replace "compound 201" with its chemical name)
3. **Complete Independence**: 
   - Each question must be self-contained with ALL necessary information, as others will not have access to the original text.
   - If a question is incomplete, incorporate missing context to make it fully solvable
   - For multi-step reactions, include all necessary steps and conditions
   - Multiple questions must be mutually independent (no "as in first Question", "using previous result")
4. **Answer Must Exist**: All answers must be directly found or clearly derivable from the given text - do not extract if uncertain
5. **No Scheme/Figure-Dependent Content**: Do NOT extract questions whose answers rely heavily on reaction schemes, structural diagrams, or figures that cannot be described adequately in text

## For Questions/Problems:
- Extract explicit chemistry questions ONLY if their complete answers are present
- **Prioritize substantive content** over trivial details: focus on chemical properties, principles, mechanisms, experimental methods, calculations, and applications rather than isolated nomenclature or data-dependent conclusions
- **For Chemical Properties**: Create questions about physical/chemical properties, reactivity, or characteristics when well-described
- **For Chemical Principles/Theories**: ask to explain the principle or test understanding of key concepts (e.g., "Explain Le Chatelier's principle", "What happens to equilibrium when...")
- **For Chemical Equations**: ask about equation balancing, reaction types, or explain the reaction principles (include complete conditions)
- **For Reaction Mechanisms**: Create questions about how reactions proceed ONLY if describable in text without complex schemes
- **For Chemical Calculations**: Extract calculation problems with ALL given values, units, and formulas only if the answer exits in the text
- **For Experiments**: create questions about experimental methods, observations, or interpret/explain phenomena (include complete procedural steps)
- **Critical**: Each question must be complete and self-contained
- **Do NOT extract**:
  - Questions depending on reaction schemes or structural diagrams not describable in text
  - Questions requiring extensive analytical data tables or spectroscopic data
  - Questions that require additional data to be provided in order to be answered
  - Questions whose answers are uncertain or incomplete
  - Questions whose answers can't be found in the text
- If no valid Q&A content can be extracted following above rules, return exactly: [NO QA]

## For Answers:
- **Process**: First locate the complete answer in the document, then organize it clearly
- For numerical answers: include units and appropriate significant figures
- If the answer has explanation, reorganize it into a clear, well-structured format
- If the answer lacks explanation, add necessary context or reasoning
- For final answers that need exact matching (multiple-choice, calculations, fill-in-the-blank, true/false), use $\\boxed{{answer}}$ notation

## Output Format (STRICT):
Use this exact format for parsing:

---BEGIN QA---
Question: <First complete question>
Answer: <First complete answer>
---END QA---

---BEGIN QA---
Question: <Second complete question>
Answer: <Second complete answer>
---END QA---

If no valid Q&A can be extracted, return exactly: [NO QA]

---
Extract:
{chunk}

Now process the extract and return the result in the specified format.

\end{TextBlock}
\end{tcolorbox}

\begin{tcolorbox}[
colback=blue!5!white,
colframe=blue!30!gray,
  boxrule=2pt,
    fontupper=\small,
  arc=8pt,
  title={\textbf{QA Extraction Prompt for Darwin-Science-Book-QA(Computer Science)}},
  fonttitle=\normalsize\bfseries,
  enhanced jigsaw,
  breakable
]
\begin{TextBlock}
Below is a document extract from computer science domain (programming, systems, algorithms, software engineering, etc.).

# Extraction Task
Extract complete, independently solvable CS Q&A pairs following these strict guidelines:

## CRITICAL RULES (Must Follow):
1. **Answer-First Principle**: Only extract a question if its complete answer exists in the document
2. **Zero References Requirement**: 
   - Questions: No references to "the document", "the text", "Figure 1", "Section 2.1", "above", "below", "the code above","in Algorithm 4.2" etc.
   - Answers: No references to "as in document", "as shown in text", "see the document", "Section 2.1", etc.
   - Replace ALL references with actual content from the document
3. **Complete Independence**: 
   - Each question must be self-contained with ALL necessary information, as others will not have access to the original text.
   - If a question is incomplete, incorporate missing context to make it fully solvable
   - For code-related questions, include necessary code context within the question
   - Multiple questions must be mutually independent (no "as in first Question", "using previous result")
4. **Answer Must Exist**: All answers must be directly found or clearly derivable from the given text - do not extract if uncertain
5. **Value-Based Selection**: Focus on extracting the MOST IMPORTANT and EDUCATIONAL content


## For Questions/Problems:
- Extract explicit questions ONLY if their complete answers are present
- **For Step-by-Step Procedures**: Extract as "How to..." questions ONLY if ALL steps are present and self-contained
- **For System Design/Architecture**: Create questions about design decisions, trade-offs, or evolution (e.g., "Why did BGS migrate from Mimer to Oracle?")
- **For Algorithms/Code**: Questions can focus on functionality, complexity, implementation details, or usage, if including code, ensure sufficient context is provided in the question
- **For Comparisons**: Create questions like "What are the differences between X and Y?" or "When to use X vs Y?"
- **For Performance**: Extract questions about optimization techniques, performance metrics, or bottlenecks
- **For Concepts/Definitions**: Ask about them flexibly, but ensure complete answer exists and focus on core concepts
- **Formulate questions naturally and appropriately** - use varied question formats based on content type, but make sure the answer can be found in the document
- If no valid Q&A content can be extracted following above rules, return exactly: [NO QA]

## For Answers:
- **Process**: First locate the complete answer in the document, then organize it clearly
- If the answer has explanation, reorganize it into a clear, well-structured format
- If the answer lacks explanation, add necessary context or reasoning
- For final answers that need exact matching (multiple-choice, calculations, fill-in-the-blank, true/false), use $\\boxed{{answer}}$ notation

## Output Format (STRICT):
Use this exact format for parsing:

---BEGIN QA---
Question: <First complete question>
Answer: <First complete answer>
---END QA---

---BEGIN QA---
Question: <Second complete question>
Answer: <Second complete answer>
---END QA---

If no valid Q&A can be extracted, return exactly: [NO QA]

---
Extract:
{chunk}

Now process the extract and return the result in the specified format.
\end{TextBlock}
\end{tcolorbox}

\begin{tcolorbox}[
colback=blue!5!white,
colframe=blue!30!gray,
  boxrule=2pt,
    fontupper=\small,
  arc=8pt,
  title={\textbf{QA Extraction Prompt for Darwin-Science-Book-QA(Engineer)}},
  fonttitle=\normalsize\bfseries,
  enhanced jigsaw,
  breakable
]
\begin{TextBlock}
Below is a book document extract.

# Extraction Task
Extract complete, independently solvable engineering questions and answers from the document while following these guidelines:

## For Questions:
- Extract any explicit engineering questions with their associated answers
- For implicit engineering principles, design methods, procedures, or technical specifications presented as statements, convert them to well-formed questions ONLY if they can stand alone
- Ensure each extracted question contains ALL necessary information to be solved independently without requiring additional context
- Include any relevant diagrams, circuit schematics, system designs, specifications, or technical data mentioned (describe them if not visible)
- Extract multiple questions separately if they exist
- **If a question lacks necessary context (specifications, constraints, parameters, standards, initial conditions), incorporate the missing information from the document to make it self-contained**
- If no engineering content can be meaningfully extracted as a question, return `[NO QA]`

## For Answers:
- Include the answer provided in the extract
- Answers should capture the essential explanation of the engineering principle, design approach, or solution methodology
- If the source material contains design procedures, calculation steps, or implementation methods, include these in the answer
- For engineering problems, the answer should explain the approach, assumptions, calculations, and practical considerations as presented in the text
- If the answer already has explanation, reorganize the solution into a clear and well-structured format for better readability and understanding
- If the answer lacks explanation, add necessary intermediate steps, reasoning, and justifications as a teacher would
- For final answers that need exact matching (calculations, design values, multiple-choice, true/false), use $\\boxed{{}}$ notation
- Include units, dimensions, and tolerances where applicable

## Requirements:
- The question should include all necessary information (specifications, constraints, parameters, standards)
- The answer should be practical, accurate, and well-explained
- Both question and answer should stand alone (no references to documents or original materials)
- Preserve technical terminology, symbols, and notation accurately
- Maintain engineering rigor and attention to practical feasibility
- Include safety considerations and industry standards where relevant

## Format:
Format each question-answer pair as:
Question: [Complete engineering question with all context needed to understand]
Answer: [Corresponding answer from the text]

The extract is as follows:
{chunk}

Now process the extract and return the result.
\end{TextBlock}
\end{tcolorbox}

\begin{tcolorbox}[
colback=blue!5!white,
colframe=blue!30!gray,
  boxrule=2pt,
    fontupper=\small,
  arc=8pt,
  title={\textbf{QA Extraction Prompt for Darwin-Science-Book-QA(Humansocial)}},
  fonttitle=\normalsize\bfseries,
  enhanced jigsaw,
  breakable
]
\begin{TextBlock}
Below is a document extract from humanities and social sciences domain.

# Extraction Task
Extract complete, independently solvable humanities/social sciences Q&A pairs following these strict guidelines:

## CRITICAL RULES (Must Follow):
1. **Answer-First Principle**: Only extract a question if its complete answer exists in the document
2. **Zero References Requirement**: 
   - Questions: No references to "the document", "the text", "the author", "Chapter 1", "above", "below", etc.
   - Answers: No references to "as in document", "as shown in text", "the author states", etc.
   - Replace ALL references with actual content from the document
3. **Complete Independence**: 
   - Each question must be self-contained with ALL necessary information
   - If a question is incomplete, incorporate missing context to make it fully solvable
   - Multiple questions must be mutually independent (no "as in first Question", "using previous result")
4. **Answer Must Exist**: All answers must be directly found or clearly derivable from the given text - do not extract if uncertain
5. **Value-Based Selection**: Focus on extracting the MOST IMPORTANT and EDUCATIONAL content with analytical substance
6. **Knowledge-Based Content Only**: Only extract from texts that convey factual information, scholarly theories, analytical arguments, or objective descriptions. Do NOT extract from:
   - Narrative fiction, creative writing, or purely storytelling passages (novels, literary narratives)
   - Religious proselytizing, propaganda, or ideological indoctrination
   - Advertising, marketing, or promotional content
   - Pure emotional expressions, personal rants, or subjective opinions without analytical substance
   - Inflammatory or extremist rhetoric

## For Questions/Problems:
- **Extract explicit question-answer pairs** if they exist (ensure zero references and independence requirements)
- **Create questions about**:
  - Historical events: causes, consequences, significance, key figures
  - Theories and concepts: definitions, frameworks, applications, explanations
  - Arguments and viewpoints: main claims, evidence, reasoning, limitations
  - Comparisons: differences and similarities between theories/events/policies/concepts
  - Impacts and significance: effects, influence, historical/social importance
- **Formulate questions naturally** - use varied formats based on content type
- **Critical**: Each question must be self-contained with necessary context and clearly distinguish facts from interpretations
- **Do NOT extract**:
  - Questions from narrative fiction, religious proselytizing, propaganda, advertising, or purely emotional content
  - Rhetorical questions without answers in the text
  - Open-ended questions without clear answers
  - Questions requiring extensive background not provided
  - Trivial details without analytical significance
- If no valid Q&A content can be extracted following above rules, return exactly: [NO QA]

## For Answers:
- **Process**: First locate the complete answer in the document, then organize it clearly
- If the answer has explanation, reorganize it into a clear, well-structured format
- If the answer lacks explanation, add necessary contextual reasoning while staying true to the text
- For final answers that need exact matching (multiple-choice, calculations, fill-in-the-blank, true/false), use $\\boxed{{}}$ notation

## Output Format (STRICT):
Use this exact format for parsing:

---BEGIN QA---
Question: <First complete question>
Answer: <First complete answer>
---END QA---

---BEGIN QA---
Question: <Second complete question>
Answer: <Second complete answer>
---END QA---

If no valid Q&A can be extracted, return exactly: [NO QA]

---
Extract:
{chunk}

Now process the extract and return the result in the specified format.
\end{TextBlock}
\end{tcolorbox}

\begin{tcolorbox}[
colback=blue!5!white,
colframe=blue!30!gray,
  boxrule=2pt,
    fontupper=\small,
  arc=8pt,
  title={\textbf{QA Extraction Prompt for Darwin-Science-Book-QA(Math)}},
  fonttitle=\normalsize\bfseries,
  enhanced jigsaw,
  breakable
]
\begin{TextBlock}
# Extraction Task
Extract complete, independently solvable mathematical Q&A pairs following these strict guidelines:

## CRITICAL RULES (Must Follow):
1. **Answer-First Principle**: Only extract a question if its complete answer exists in the document
2. **Zero References Requirement**: 
   - Questions: No references to "the document", "the text", "Figure 1", "Theorem 1.1", "above", "below", etc.
   - Answers: No references to "as in document", "as shown in text", "see the document","by Theorem 1.1" etc.
   - Replace ALL references with actual content from the document
3. **Complete Independence**: 
   - Each question must be self-contained with ALL necessary information
   - If a question is incomplete, incorporate missing context to make it fully solvable
   - Multiple questions must be mutually independent (no "as in first Question ", "using previous result")
4. **Answer Must Exist**: All answers must be directly found or clearly derivable from the given text - do not extract if uncertain

## For Questions/Problems:
- Extract explicit mathematical questions ONLY if their complete answers are present
- For **Theorem/Proposition/Corollary + Proof/DEMONSTRATION** pairs: Convert into "Prove that [full theorem statement]" format
- For **Standalone Theorems** (without proof): Create questions about the theorem content (e.g., if theorem states "(x-y)(x+y)=x²-y²", ask "(x-y)(x+y)=?" or ask to state/explain the theorem) ONLY if the answer content is in the text
- For **Concepts/Definitions**: You may ask about them, but ensure the complete answer can be found in the document
- **If you see only solution steps without a question, DO NOT create a question**
- If no valid Q&A content can be extracted following above rules, return exactly: [NO QA]

## For Answers:
- **Process**: First locate the complete answer in the document, then enhance the reasoning steps
- Include the provided solution, proof, or demonstration when available
- **Replace all references**: Do NOT use "by Theorem 1.2", "using Lemma 3", "from Definition 2.1", etc. Instead, state the actual theorem/lemma/definition content or rephrase the reasoning without references
- For theorems/propositions, the answer should contain the complete proof
- If the answer has explanation, reorganize it into a clear, well-structured format
- If the answer lacks explanation, add necessary intermediate reasoning as a teacher would
- For final answers that need exact matching (multiple-choice, calculations, fill-in-the-blank, true/false), use $\\boxed{{answer}}$ notation

## Output Format (STRICT):
Use this exact format for parsing:

---BEGIN QA---
Question: <First complete question>
Answer: <First complete answer>
---END QA---

---BEGIN QA---
Question: <Second complete question>
Answer: <Second complete answer>
---END QA---

If no valid Q&A can be extracted, return exactly: [NO QA]


Extract:
{chunk}

Now process the extract and return the result in the specified format.
\end{TextBlock}
\end{tcolorbox}

\begin{tcolorbox}[
colback=blue!5!white,
colframe=blue!30!gray,
  fontupper=\small,
  boxrule=2pt,
  arc=8pt,
  title={\textbf{QA Extraction Prompt for Darwin-Science-Book-QA(Medicine})},
  fonttitle=\normalsize\bfseries,
  enhanced jigsaw,
  breakable
]
\begin{TextBlock}
Below is a document extract from medical sciences domain (clinical medicine, pharmacology, physiology, pathology, public health, etc.).

# Extraction Task
Extract complete, independently solvable medical Q&A pairs following these strict guidelines:

## CRITICAL RULES (Must Follow):
1. **Answer-First Principle**: Only extract a question if its complete answer exists in the document
2. **Zero References Requirement**: 
   - Questions: No references to "the document", "according to the text", "Figure 1", "Table 3", "above", "below", "the study above", etc.
   - Answers: No references to "as in document", "as shown in text", "see Table", "Figure 1", etc.
   - Replace ALL references with actual content from the document
3. **Complete Independence**: 
   - Each question must be self-contained with ALL necessary information
   - If a question is incomplete, incorporate missing context to make it fully solvable
   - Multiple questions must be mutually independent (no "as in first Question", "using previous result")
4. **Answer Must Exist**: All answers must be directly found or clearly derivable from the given text - do not extract if uncertain
5. **Value-Based Selection**: Focus on extracting the MOST IMPORTANT and EDUCATIONAL medical content
6. **No Data-Dependent Conclusions**: Do NOT extract questions whose answers rely heavily on complex data tables, diagnostic images, or clinical charts that cannot be adequately described in text

## For Questions/Problems:
- **Extract explicit question-answer pairs** if they exist (ensure zero references and independence requirements)
- **Prioritize substantive clinical and medical content** over trivial details or data-dependent conclusions
- **For Disease-related content**: Create questions about definition, classification, pathophysiology, clinical manifestations, diagnosis, prevention, complications, or risk factors
- **For Drug/Treatment-related content**: Create questions about drug composition, indications, mechanisms of action, dosage, side effects, or other treatment modalities 
- **For Physiological mechanisms**: Create questions about normal body functions, pathological processes, biochemical mechanisms, or metabolic pathways 
- **For Concepts and comparisons**: Create questions about medical definitions, comparisons between diseases/treatments/approaches, or differential diagnosis
- **Critical**: Each question must be complete and self-contained, as others will not have access to the original text.
- **Do NOT extract**:
  - Questions depending on diagnostic images, complex clinical charts, or detailed lab data tables
  - Questions requiring extensive clinical case details not fully provided
  - Questions whose answers are uncertain, speculative without clear indication, or incomplete
  - Medical advice for specific individual cases (focus on general medical knowledge)
- If no valid Q&A content can be extracted following above rules, return exactly: [NO QA]

## For Answers:
- **Process**: First locate the complete answer in the document, then organize it clearly
- If the answer has explanation, reorganize it into a clear, well-structured format
- If the answer lacks explanation, add necessary medical reasoning or context
- For final answers that need exact matching (multiple-choice, calculations, fill-in-the-blank, true/false), use $\\boxed{{}}$ notation

## Output Format (STRICT):
Use this exact format for parsing:

---BEGIN QA---
Question: <First complete question>
Answer: <First complete answer>
---END QA---

---BEGIN QA---
Question: <Second complete question>
Answer: <Second complete answer>
---END QA---

If no valid Q&A can be extracted, return exactly: [NO QA]

---
Extract:
{chunk}

Now process the extract and return the result in the specified format.
\end{TextBlock}
\end{tcolorbox}

\begin{tcolorbox}[
colback=blue!5!white,
colframe=blue!30!gray,
  fontupper=\small,
  boxrule=2pt,
  arc=8pt,
  title={\textbf{QA Extraction Prompt for Darwin-Science-Book-QA(Physics)}},
  fonttitle=\normalsize\bfseries,
  enhanced jigsaw,
  breakable
]
\begin{TextBlock}
Below is a book document extract.

# Extraction Task
Extract complete, independently solvable physics content following these guidelines:

## CRITICAL RULES (Must Follow):
1. **Answer-First Principle**: Only extract a question if its complete answer exists in the document
2. **Zero References Requirement**: 
   - Questions: No references to "the document", "the text", "Figure 1", "Theorem 1.1", "above", "below", etc.
   - Answers: No references to "as in document", "as shown in text", "see the document","Theorem 1.1" etc.
   - Replace ALL references with actual content from the document
3. **Complete Independence**: 
   - Each question must be self-contained with ALL necessary information
   - If a question is incomplete, incorporate missing context to make it fully solvable
   - Multiple questions must be mutually independent (no "as in first Question ", "using previous result")
4. **Answer Must Exist**: All answers must be directly found or clearly derivable from the given text - do not extract if uncertain

## For Questions/Problems:
- Extract any explicit physics questions ONLY if their answers are present in the document
- If you see only solution steps without a question, DO NOT create a question
- **For Laws/Principles with proofs/derivations**: Create questions asking for the derivation or proof when available
- **For Standalone Laws/Principles** (without derivation): Create questions about their content, applications, or physical meaning ONLY if the answer is in the text
- **For Formulas**: Create questions flexibly (e.g., if text derives E=mc², could ask "Derive the mass-energy relation" or "What is the relationship between mass and energy?" or apply it in a scenario) ONLY if the answer content is in the text
- **For Concepts/Definitions**: Ask about them flexibly, but ensure the complete answer can be found in the document
- **For Phenomena Explanations**: Create questions naturally based on the content
- Extract multiple questions separately if they exist, ensuring each is completely independent without relying on or referencing other questions
- If no valid physics Q&A content can be extracted following above rules, return `[NO QA]`

## For Answers:
- **Process: First locate the answer in the document, then enhance the reasoning steps**
- Include the provided solution, derivation, or explanation when available
- For derivations/proofs, the answer should contain the complete step-by-step process
- If the answer already has explanation, reorganize it into a clear and well-structured format for better readability and understanding
- If the answer lacks explanation, add necessary intermediate reasoning (physical reasoning, mathematical steps, unit analysis, physical interpretation)
- For final answers that need exact matching (multiple-choice, calculations, fill-in-the-blank, true/false), use $\\boxed{{}}$ notation

## Format:
Use this exact format for parsing:

---BEGIN QA---
Question: <First complete question>
Answer: <First complete answer>
---END QA---

---BEGIN QA---
Question: <Second complete question>
Answer: <Second complete answer>
---END QA---


The extract is as follows:
{chunk}

Now process the extract and return the result.
\end{TextBlock}
\end{tcolorbox}

\begin{tcolorbox}[
colback=blue!5!white,
colframe=blue!30!gray,
  fontupper=\small,
  boxrule=2pt,
  arc=8pt,
  title={\textbf{QA Extraction Prompt for Darwin-Science-Book-QA(stem-others)}},
  fonttitle=\normalsize\bfseries,
  enhanced jigsaw,
  breakable
]
\begin{TextBlock}
Below is a document extract from STEM fields (engineering, applied sciences, technology, interdisciplinary sciences, etc.).

# Extraction Task
Extract complete, independently solvable STEM Q&A pairs following these strict guidelines:

## CRITICAL RULES (Must Follow):
1. **Answer-First Principle**: Only extract a question if its complete answer exists in the document
2. **Zero References Requirement**: 
   - Questions: No references to "the document", "the text", "Figure 1", "Table 2", "Equation 3", "above", "below", etc.
   - Answers: No references to "as in document", "as shown in text", "see Figure", "Equation 1", etc.
   - Replace ALL references with actual content from the document
3. **Complete Independence**: 
   - Each question must be self-contained with ALL necessary information
   - If a question is incomplete, incorporate missing context to make it fully solvable
   - Multiple questions must be mutually independent (no "as in first Question", "using previous result")
4. **Answer Must Exist**: All answers must be directly found or clearly derivable from the given text - do not extract if uncertain
5. **Value-Based Selection**: Focus on extracting the MOST IMPORTANT and EDUCATIONAL content
7. **No Figure/Table-Dependent Content**: Do NOT extract questions whose answers rely heavily on complex diagrams, charts, or data tables that cannot be adequately described in text

## For Questions/Problems:
- **Extract explicit question-answer pairs** if they exist (ensure zero references and independence requirements)
- **Prioritize substantive technical and scientific content** over trivial details or data-dependent conclusions
- **Create questions about**:
  - Principles and theories: fundamental concepts, laws, theoretical frameworks, underlying principles
  - Formulas and equations: mathematical expressions, relationships, derivations (when describable in text)
  - Mechanisms and processes: how systems work, operational principles, step-by-step processes
  - Concepts and definitions: technical terminology, key concepts, classifications
  - Properties and characteristics: material properties, system characteristics, performance parameters
  - Comparisons: differences between systems/methods/materials/approaches
- **Critical**: Each question must be self-contained with all necessary technical context, definitions, conditions, and parameters
- **Do NOT extract**:
  - Questions depending on complex diagrams, charts, or detailed data tables
  - Questions requiring extensive background knowledge not provided in the text
  - Trivial details without educational or technical significance
  - Questions whose answers are uncertain or incomplete
- If no valid Q&A content can be extracted following above rules, return exactly: [NO QA]

## For Answers:

- **Process**: First locate the complete answer in the document, then organize it clearly
- If the answer has explanation, reorganize it into a clear, well-structured format
- If the answer lacks explanation, add necessary technical reasoning or context
- For final answers that need exact matching (multiple-choice, calculations, selections, specific values), use \texttt{\$\textbackslash boxed\{answer\}\$} notation

## Output Format (STRICT):
Use this exact format for parsing:

---BEGIN QA---
Question: <First complete question>
Answer: <First complete answer>
---END QA---

---BEGIN QA---
Question: <Second complete question>
Answer: <Second complete answer>
---END QA---

If no valid Q&A can be extracted, return exactly: [NO QA]

---
Extract:
{chunk}

Now process the extract and return the result in the specified format.

\end{TextBlock}
\end{tcolorbox}

\end{document}